\documentclass[runningheads]{llncs}

\usepackage{eccv}

\usepackage{eccvabbrv}

\usepackage{graphicx}
\usepackage{booktabs}

\usepackage[accsupp]{axessibility}  
\usepackage{hyperref}
\hypersetup{hidelinks}
\usepackage{orcidlink}

\usepackage{enumitem}
\usepackage{multirow}
\usepackage[table]{xcolor}  
\usepackage{makecell}
\usepackage{float}  

\newcommand\blfootnote[1]{%
  \begingroup
  \renewcommand\thefootnote{}%
  \footnotetext{#1}%
  \endgroup
}

\usepackage{microtype}  

\makeatletter
\@ifpackageloaded{eccvabbrv}{}{
  \newcommand{\eg}{\emph{e.g.}\xspace}
  \newcommand{\ie}{\emph{i.e.}\xspace}
  \newcommand{\etal}{\emph{et al.}\xspace}

  \newcommand{\cf}{\emph{cf.}\xspace}
}
\makeatother

\usepackage[capitalize,noabbrev]{cleveref}

\AtBeginDocument{%
  \crefname{figure}{Fig.}{Figs.}%
  \Crefname{figure}{Fig.}{Figs.}%
  \crefname{table}{Table}{Tables}%
  \Crefname{table}{Table}{Tables}%
  \crefname{section}{Section}{Sections}%
  \Crefname{section}{Section}{Sections}%
  \crefformat{equation}{Eq.~(#2#1#3)}%
  \Crefformat{equation}{Eq.~(#2#1#3)}%
  \crefrangeformat{equation}{Eqs.~(#3#1#4) to~(#5#2#6)}%
  \crefmultiformat{equation}{Eqs.~(#2#1#3)}{ and~(#2#1#3)}{, (#2#1#3)}{ and~(#2#1#3)}%
  \Crefmultiformat{equation}{Eqs.~(#2#1#3)}{ and~(#2#1#3)}{, (#2#1#3)}{ and~(#2#1#3)}%
}
\definecolor{darkred}{rgb}{0.6,0,0}
\definecolor{revisionpurple}{rgb}{0.5,0,0.5}
\definecolor{weakpink}{rgb}{0.8,0.2,0.4}

\newcommand{\best}[1]{\textbf{\textcolor{teal}{#1}}}
\newcommand{\second}[1]{\underline{#1}}

\begin{document}

\title{Inductive Visual Logic for Few-Shot Out-Of-Distribution Adaptation in VLMs} 

\titlerunning{Inductive Visual Logic}

\author{Hung-Jen Chen\inst{1}\orcidlink{0009-0003-9345-8263} \and
Yu-Heng Ho$^{\star}$\inst{1} \and
Ting-Yao Huang$^{\star}$\inst{1} \and
Po-Hsiang Hsu\inst{1} \and
Li-Yu Chen\inst{1} \and
Chun-Yi Lee\inst{2}\orcidlink{0000-0002-4680-4800} \and
Min Sun\inst{1}\orcidlink{0000-0001-9598-8178}}

\authorrunning{Chen et al.}

\institute{National Tsing Hua University, Hsinchu, Taiwan\\
\email{andyqmongo@gapp.nthu.edu.tw, sunmin@ee.nthu.edu.tw}\\
\and
National Taiwan University, Taipei, Taiwan\\
\email{cylee@csie.ntu.edu.tw}}

\maketitle
\blfootnote{$^{\star}$~Equal contribution.}

\begin{abstract}
Generative vision-language models (VLMs) such as Qwen-VL and LLaVA achieve strong zero-shot performance on tasks overlapping with their pretraining distribution, yet fail on specialized domains where the required discriminative features were never learned, a regime we term \emph{distant} out-of-distribution (OOD). Standard adaptation methods cannot overcome this representational absence because they operate within the encoder's existing feature space. However, VLMs retain a robust \emph{descriptive} capacity even when discrimination collapses: a model that cannot classify a medical scan can still articulate its visual patterns. Exploiting this asymmetry, we introduce \textbf{I}nductive \textbf{V}isual \textbf{L}ogic (IVL), a training-free framework that constructs classification knowledge from the model's surviving descriptive ability. IVL extracts visual traits from few-shot support images through dual-mode prompting, combining semantic descriptions with primitive visual observations, and organizes them into per-class trait dictionaries. At inference, hierarchical filtering identifies spatially grounded trait evidence for classification. Across multiple distant-OOD benchmarks, IVL achieves the highest aggregate accuracy under two VLM backbones while producing interpretable, trait-traceable predictions.
\end{abstract}

\section{Introduction}
\label{section:intro}

Generative vision-language models (VLMs) such as Qwen-VL~\cite{bai2025qwen2} and LLaVA~\cite{liu2023visual} achieve remarkable zero-shot performance on tasks whose visual concepts overlap with web-scale pretraining data~\cite{radford2021learning, jia2021scaling}. This capability, however, conceals a fundamental limitation. When deployed to specialized domains whose discriminative visual primitives were never encountered during pretraining, these models face not a distributional shift but a \emph{representational absence}, where the features required for classification simply do not exist within the learned representation space. Medical imaging, industrial inspection, and scientific analysis routinely fall into this regime, as their visual taxonomies bear little resemblance to web-crawled natural images. Indeed, the strong few-shot results on standard benchmarks largely reflect prior exposure to those task distributions during pretraining rather than a general capacity for rapid adaptation~\cite{li2024task}.

\begin{figure}[t]
    \centering
    \includegraphics[width=\textwidth]{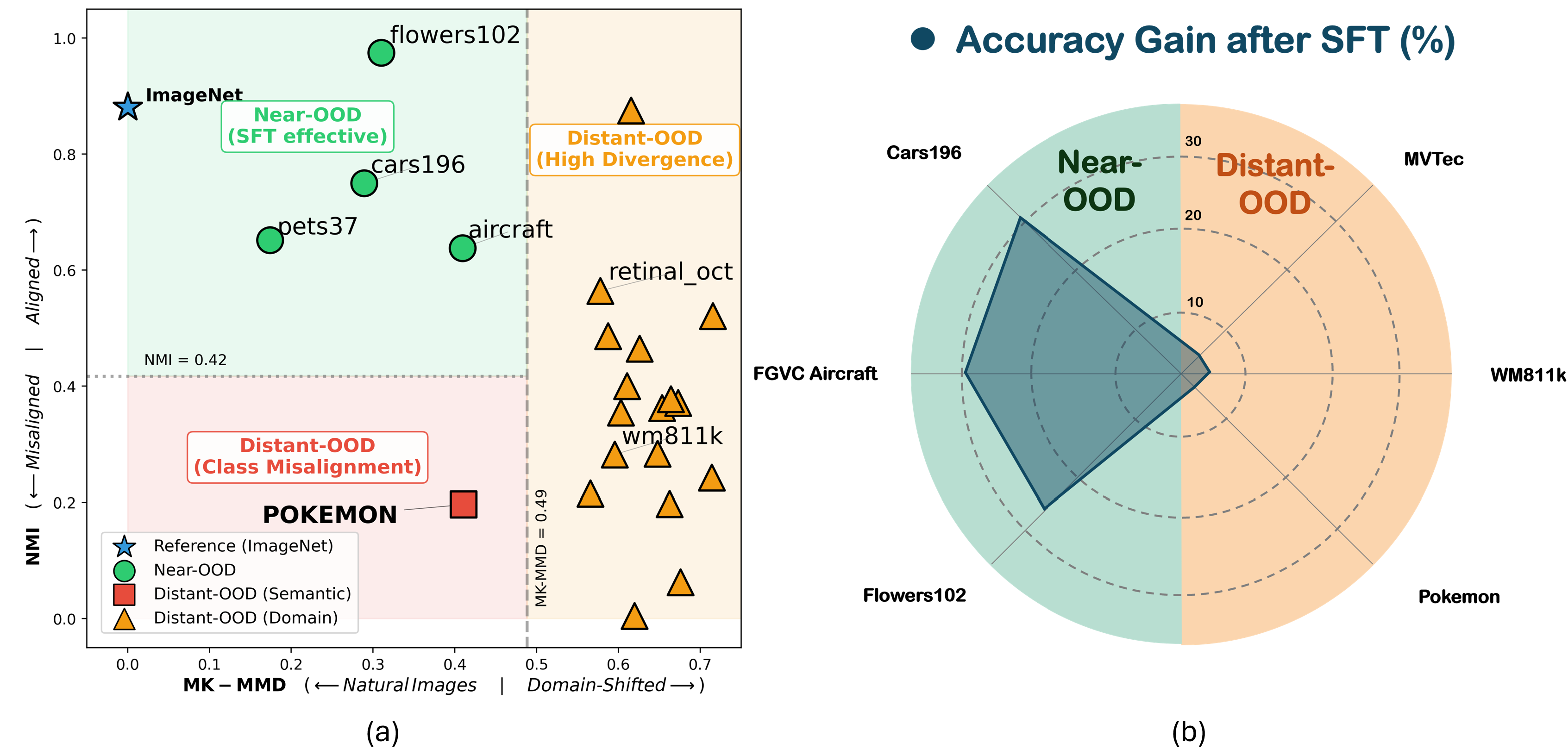}
    \vspace{-1.5em}
    \caption{Distant-OOD vs.\ near-OOD characterization. (a)~Two-dimensional projection based on feature-space divergence (MK-MMD) and latent cluster-label alignment (NMI). Distant-OOD datasets fall outside the low-divergence, high-alignment quadrant where parametric methods succeed. (b)~Radar chart of accuracy gains after SFT across datasets, showing that fine-tuning yields substantial improvements only in near-OOD settings but only marginal gains on specialized domains.}
    \label{fig:teaser-a}
    \vspace{-1.5em}
\end{figure}

This \emph{distant out-of-distribution (OOD)} regime defeats every major adaptation paradigm because all share a common assumption: discriminative visual representations already exist within the model and merely require refinement. Gradient-based methods, including SFT, LoRA~\cite{hu2022lora}, and Visual-RFT~\cite{visualrft}, optimize within the subspace spanned by the encoder's existing features; when the optimal decision boundary requires directions orthogonal to this subspace, gradient descent cannot easily recover the missing structure. This failure extends to RL-based adaptation: Although RL-based post-training generalizes better than SFT in standard settings~\cite{chu2025sft, visualrft}, Visual-RFT offers no advantage over SFT on distant-OOD tasks (\cref{tab:main_results}), suggesting that exploration does not reliably discover features the encoder never learned. As \cref{fig:teaser-a} quantifies, SFT yields negligible gains on distant-OOD domains despite full parameter updates. Prompt-based~\cite{coop, cocoop} and attribute-guided~\cite{tian2024argue, menon2022visual, pratt2023does} approaches remain anchored to the pretrained semantic space; when that space encodes no knowledge of the target domain, generated descriptors correlate spuriously with categories~\cite{roth2023waffling}. In-context learning~\cite{min2022rethinking} reweights existing representations without structural modification, offering no recourse when the features are absent. Distant-OOD adaptation demands the \emph{construction} of new visual knowledge, not refinement of existing representations.

\begin{figure}[t]
    \centering
    \includegraphics[width=\textwidth]{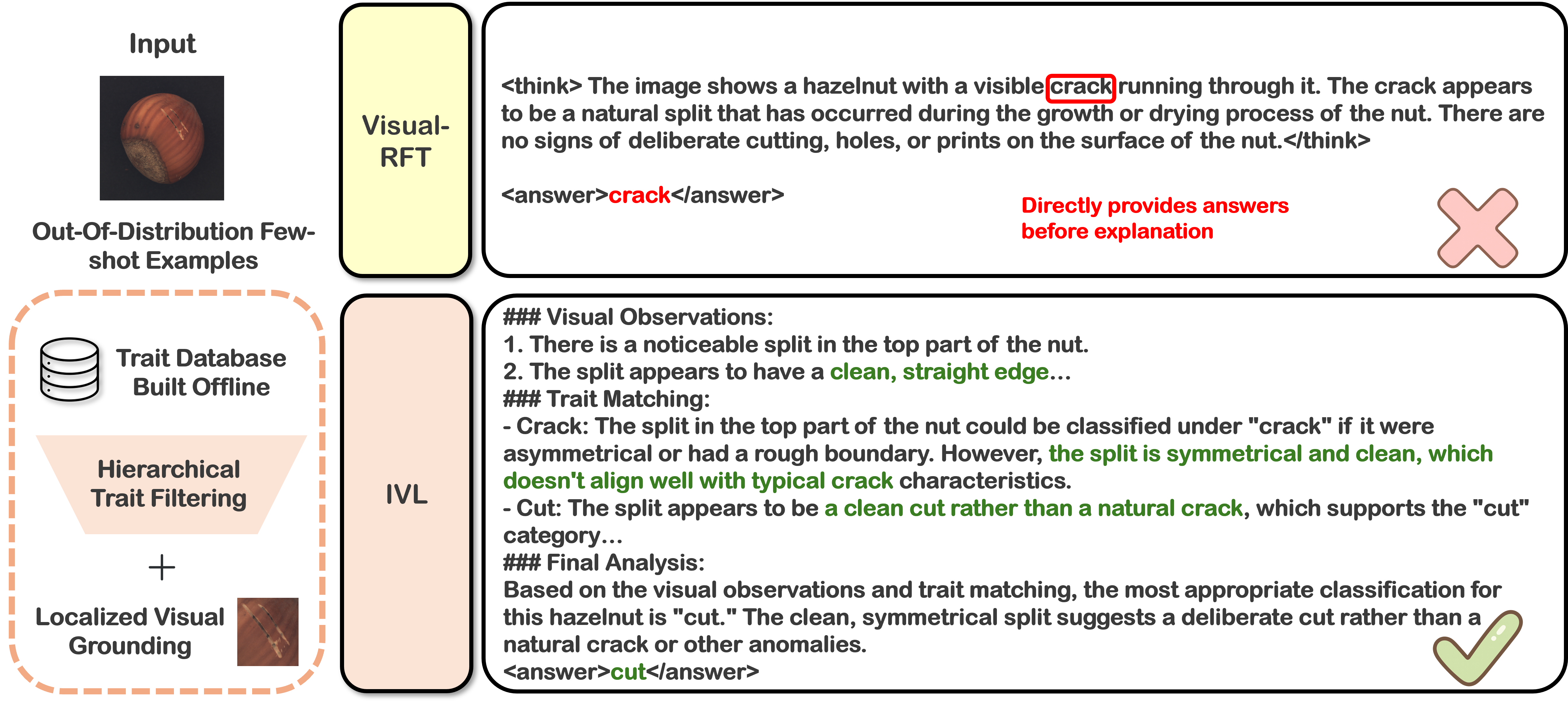}
    \vspace{-2.5em}
    \caption{Reasoning comparison between gradient-based methods and IVL. Visual-RFT produces answers before articulating reasoning, deviating from human cognitive patterns. IVL mirrors human inductive learning by first observing visual traits, comparing them against a structured traits dictionary, refining evidence with prior knowledge, and arriving at an evidence-grounded classification decision.}
    \label{fig:teaser-b}
    \vspace{-1.5em}
\end{figure}

A crucial observation motivates a departure from this paradigm. Although a VLM's discriminative representations collapse in distant-OOD domains, its \emph{generative descriptive capacity} remains largely intact. A model that cannot classify a semiconductor wafer map can still describe ``a ring-shaped concentration along the perimeter'' and ``scattered clusters near the center''; a model that fails to type Pok\'emon can nonetheless report coloration patterns, body shapes, and surface textures. This asymmetry arises because descriptive generation draws on the language decoder's domain-agnostic visual vocabulary, encompassing colors, textures, shapes, and spatial relations, which is reinforced across the full breadth of pretraining and is not tied to any classification task. Human visual learning exploits precisely this distinction: when encountering unfamiliar categories, humans engage in explicit inductive reasoning, systematically observing examples, identifying salient attributes, and constructing classification rules from these traits~\cite{lake2015human, Tenenbaum2011HowTG}. As \cref{fig:teaser-b} illustrates, existing methods instead force models to produce answers from opaque internal representations, often committing to a classification before articulating any reasoning. Channeling VLMs' surviving descriptive ability into structured trait-level reasoning offers a principled path to bypass the representational bottleneck.

Building on this insight, we introduce \textbf{I}nductive \textbf{V}isual \textbf{L}ogic (IVL), a training-free framework that transforms VLMs' surviving descriptive capacity into explicit trait-based classification. IVL extracts discriminative traits from few-shot support images through dual-mode prompting, combining semantic descriptions with low-level observations that remain reliable even when semantic understanding collapses. Extracted traits are organized into per-class dictionaries, and a hierarchical filtering mechanism at inference identifies spatially grounded trait evidence for classification. Unlike attribute-based methods~\cite{menon2022visual, pratt2023does, tian2024argue} that generate descriptors solely from pretrained semantic knowledge, IVL extracts traits from visual observations of support images and grounds every prediction in localized visual evidence. Across distant-OOD benchmarks spanning medical imaging, industrial inspection, and synthetic reasoning, IVL achieves the highest aggregate accuracy under both Qwen-VL and LLaVA in 1-shot and 8-shot settings while remaining entirely training-free. The principal contributions are:

\begin{itemize}

\item \textbf{Representational absence}, as distinct from distributional shift, is identified as the fundamental barrier to VLM adaptation in specialized domains. A two-dimensional characterization based on feature-space divergence (MK-MMD~\cite{long2015learning}) and latent cluster-label alignment (NMI) provides a principled criterion for identifying distant-OOD regimes.

\item \textbf{Explicit trait-based reasoning}, built upon VLMs' surviving descriptive capacity, proves effective precisely where parametric methods fail, achieving broad improvements across diverse benchmarks and two VLM backbones with interpretable, evidence-grounded predictions.

\item \textbf{Dual-mode extraction}, combining semantic and primitive visual prompting, is essential for robust trait construction. The resulting framework requires no weight updates and enables immediate deployment.

\end{itemize}
\section{Related Work}
\label{section:relatedwork}

\subsection{Generative Vision-Language Models}
\label{sec:related_work:vlm}

Generative VLMs~\cite{yin2024survey} such as LLaVA~\cite{liu2023visual} and Qwen-VL~\cite{bai2025qwen2} have demonstrated remarkable zero-shot~\cite{dai2023instructblip} and chain-of-thought reasoning~\cite{zhang2023multimodal} capabilities through LLM backbones that generate textual responses conditioned on visual inputs. Despite their broad knowledge foundation, these models exhibit significant performance degradation on specialized domains where high-level semantic patterns from natural images fail to transfer~\cite{li2023llava}. Parameter-efficient methods like LoRA~\cite{hu2022lora} show limited effectiveness under large domain gaps, motivating few-shot adaptation strategies tailored for generative VLMs.

\subsection{Few-Shot Domain Adaptation with VLMs}
\label{sec:related_work:fsda_vlm}

Few-shot domain adaptation for VLMs spans several paradigms. Prompt learning methods such as CoOp~\cite{coop} and CoCoOp~\cite{cocoop} learn continuous prompts that adapt to novel domains, while attribute-guided approaches like ArGue~\cite{tian2024argue} align models with primitive visual attributes. Parameter-efficient fine-tuning methods, including LoRA~\cite{hu2022lora}, MMA~\cite{mma}, and PACE~\cite{pace}, insert lightweight adapters with minimal overhead; Tip-Adapter~\cite{zhang2021tip} eliminates backpropagation entirely by constructing adapter weights from a key-value cache. Visual-RFT~\cite{visualrft} applies RL-guided fine-tuning for rapid adaptation. Nevertheless, recent studies~\cite{li2024task} and our experiments reveal that these methods universally fail on distant-OOD tasks. Prompt tuning assumes compositional recombination of existing features~\cite{coop}, PEFT methods rely on adaptation within the pretrained subspace~\cite{hu2022lora,zhang2021tip}, and even Visual-RFT performs worse than SFT on most distant-OOD benchmarks. This universal failure suggests that distant-OOD adaptation demands fundamentally different approaches beyond parameter refinement.

\subsection{Trait-Based and Neuro-Symbolic Approaches}
\label{sec:related_work:trait_base}

Attribute-based recognition has long explored interpretable alternatives to end-to-end learning. Classical work on visual attributes~\cite{5206772, 5206594} demonstrated that modeling explicit visual properties enables zero-shot recognition through attribute composition. Concept bottleneck models~\cite{koh2020concept} enforce interpretability by predicting human-understandable concepts before classification, though they require extensive annotations. In the VLM era, \cite{menon2022visual} queried GPT-3 for visual descriptors to construct zero-shot classifiers, and CuPL~\cite{pratt2023does} showed that VLM-generated descriptions improve CLIP's zero-shot accuracy, though neither addresses few-shot learning from novel domains. ArGue~\cite{tian2024argue} demonstrated that aligning VLMs with explicit attributes via prompt tuning mitigates distribution shifts, but recent work~\cite{ICLR2025_530eb593} reveals that VLMs over-rely on spuriously correlated attributes, leading to poor OOD generalization. More fundamentally, VLMs exhibit systematic compositional failures~\cite{thrush2022winoground, ma2023crepe, yuksekgonul2023when} that compound these limitations: models process visual-linguistic inputs through bag-of-words representations rather than genuine compositional understanding, making the construction of new visual concepts from primitives unreliable. Unlike these approaches, IVL grounds both semantic and low-level traits in direct observation of the support images rather than relying on the model to compose discriminative concepts internally, with the low-level mode targeting the domain-agnostic vocabulary that \cref{prop:asymmetry}(ii) identifies as surviving in the distant-OOD regime.

\section{Methodology}
\label{section:method}

\subsection{Problem Formulation}
\label{sec:method:problem_definition}

Consider a generative VLM $f$ with vision encoder $\phi_v$ and language decoder $\phi_l$, pretrained on a large-scale distribution $\mathcal{P}_{\mathrm{train}}$. Given a target distribution $\mathcal{P}_{\mathrm{target}}$ whose discriminative visual primitives lie outside the support of $\mathcal{P}_{\mathrm{train}}$, and a support set $\mathcal{D} = \{(x_i, y_i)\}_{i=1}^{n \times C}$ containing $n$ labeled examples for each of $C$ classes, the objective is to classify query images $x_q$ into one of the $C$ classes without updating any parameters of $f$.

IVL reformulates classification through explicit trait-based reasoning. A \textit{trait} $t$ is a natural-language description of a visual attribute (\eg, ``red wings,'' ``spotted pattern,'' ``layered horizontal bands''). From $\mathcal{D}$, IVL constructs a trait database $\mathcal{T} = \bigcup_{c=1}^{C} \mathcal{T}_c$, where each class $c$ is represented by discriminative traits $\mathcal{T}_c = \{t_1^c, t_2^c, \ldots, t_{m_c}^c\}$ extracted from its support images. Classification of $x_q$ then reduces to identifying which traits in $\mathcal{T}$ are visually grounded in $x_q$ and mapping matched traits to a class prediction.

\subsection{Theoretical Foundations}
\label{sec:method:theory}

The design of IVL rests on two empirically motivated observations about VLM behavior in distant-OOD regimes. Gradient-based adaptation is fundamentally limited when the required discriminative features are absent from the learned representation space, and the model's capacity to generate visual descriptions survives even when its discriminative capacity collapses. These observations are formalized below as propositions; their formal justifications appear in the Supp.~\S\ref{appendix:theory}, and their connection to the measurable diagnostic introduced in \cref{sec:method:representational_absence} is made explicit throughout.


\begin{proposition}[Descriptive-Discriminative Asymmetry]
\label{prop:asymmetry}
Let $f = (\phi_v, \phi_l)$ be a VLM with vision encoder $\phi_v$ and language decoder $\phi_l$, pretrained on distribution $\mathcal{P}_{\mathrm{train}}$. Define the \emph{discriminative capacity} $D(f, \mathcal{P}_{\mathrm{target}})$ as the mutual information $I\bigl(\phi_v(x);\, y\bigr)$ between the encoder's representation and the task label $y$ under target distribution $\mathcal{P}_{\mathrm{target}}$, and define the \emph{descriptive capacity} $G(f, \mathcal{P}_{\mathrm{target}})$ as the expected recall of visually grounded primitive attributes (\eg, colors, textures, shapes, spatial relations) in descriptions generated by $\phi_l$ for images drawn from $\mathcal{P}_{\mathrm{target}}$. If $\phi_l$ was pretrained on a corpus whose visual vocabulary spans general primitive attributes, then:
\begin{enumerate}[label=(\roman*),nosep,leftmargin=*]
    \item $D(f, \mathcal{P}_{\mathrm{target}})$ degrades as the feature-space divergence $\delta\bigl(\mathcal{P}_{\mathrm{train}}, \mathcal{P}_{\mathrm{target}}\bigr)$ increases, approaching chance-level performance (assumptions in Supp.~\S\ref{appendix:theory});
    
    \item $G(f, \mathcal{P}_{\mathrm{target}}) \geq G_0 > 0$ for all $\mathcal{P}_{\mathrm{target}}$, where $G_0$ depends only on the breadth of the vision-language backbone's visual vocabulary and is independent of $\delta$.
\end{enumerate}
\end{proposition}


\begin{proposition}[Gradient Futility under Representational Absence]
\label{prop:gradient_futility}

Let $\phi_v : \mathcal{X} \to \mathbb{R}^d$ be a frozen vision encoder and let $\Phi = \mathrm{span}\bigl\{\phi_v(x) : x \in \mathcal{X}_{\mathrm{train}}\bigr\}$ denote the subspace spanned by training-distribution representations. For a target distribution $\mathcal{P}_{\mathrm{target}}$, let $w^\dagger \in \mathbb{R}^d$ be the unconstrained population minimizer of a squared-loss linear surrogate, and decompose it as $w^\dagger = w^\dagger_\parallel + w^\dagger_\perp$, where $w^\dagger_\parallel \in \Phi$ and $w^\dagger_\perp \in \Phi^\perp$. Under the assumptions in Supp.~\S\ref{appendix:theory}, any linear classifier $\hat{w}$ obtained by optimization over $\Phi$ satisfies

\begin{equation}
\label{eq:gradient_bound}
    \mathcal{R}_{\mathrm{sq}}(\hat{w}) \;\geq\; \mathcal{R}_{\mathrm{sq}}(w^\dagger) \;+\; \Omega\!\left(\frac{\|w^\dagger_\perp\|^2}{\|w^\dagger\|^2}\right),
\end{equation}
where $\mathcal{R}_{\mathrm{sq}}(\cdot)$ denotes squared-loss risk. The excess risk is irreducible with respect to the number of gradient steps, as $w^\dagger_\perp$ lies permanently outside the optimization domain $\Phi$.

\end{proposition}


\begin{remark}[Implications for IVL and SFT]
\label{rem:ivl-sft}
Propositions~\ref{prop:asymmetry} and~\ref{prop:gradient_futility} together provide a theoretical grounding for two complementary observations in practice. 
\paragraph{SFT addresses the symptom, not the cause.}
Supervised fine-tuning repairs the degraded discriminative capacity $D(f, \mathcal{P}_{\mathrm{target}})$ by re-learning task-specific decision boundaries on labeled target-domain data. However, this approach is costly and ignores a crucial asymmetry: the discriminative capacity $D$ is bottlenecked by the \emph{vision encoder} $\phi_v$, whose frozen representations become increasingly uninformative as feature-space divergence $\delta(\mathcal{P}_{\mathrm{train}}, \mathcal{P}_{\mathrm{target}})$ grows, not by the language decoder $\phi_l$, which retains descriptive capacity $G \geq G_0$ regardless of domain shift. 
Proposition~\ref{prop:gradient_futility} establishes that any classifier confined to the pretrained encoder's representation space incurs irreducible excess risk proportional to the missing subspace; \cref{tab:main_results} confirms that encoder-updating methods (SFT, LoRA, Visual-RFT) exhibit the same failure empirically, expending their supervision budget on structure the encoder never encoded.

\paragraph{IVL exploits the preserved descriptive capacity.}
Rather than repairing the degraded vision encoder $\phi_v$, IVL routes reasoning through the guaranteed lower bound on $G$. Concretely, IVL: (i)~uses the vision-language decoder $\phi_l$ to ground images in primitive attribute descriptions, leveraging $G \geq G_0$; (ii)~organizes these descriptions into structured per-class trait dictionaries; and (iii)~classifies novel examples by retrieving and spatially grounding matched traits, bypassing the encoder's degraded discriminative associations. This yields a few-shot pathway that is robust to feature-space divergence $\delta(\mathcal{P}_{\mathrm{train}}, \mathcal{P}_{\mathrm{target}})$ by design, directly addressing the structural limitation identified in Proposition~\ref{prop:gradient_futility}. Where SFT requires labels to reconstruct $D$, IVL substitutes structured reasoning over the preserved descriptive capacity $G$, making the two approaches complementary. The formal derivation, including the information-theoretic argument and the conditions under which $G_0$ is maximized, appears in Supp.~\S\ref{appendix:proof_p1}.
\end{remark}

Taken together, Propositions~\ref{prop:asymmetry} and~\ref{prop:gradient_futility} delineate both the failure mode of existing methods and the opportunity that IVL exploits. Parametric adaptation is structurally limited by the representational gap (\cref{prop:gradient_futility}), yet the VLM retains sufficient descriptive capacity to articulate visual observations about novel domains (\cref{prop:asymmetry}). The following section introduces a two-dimensional diagnostic that determines, before any adaptation attempt, which structural properties of a target dataset govern whether gradient-based adaptation succeeds or fails, using only frozen encoder representations and ground-truth class labels.

\subsection{Characterizing Representational Absence}
\label{sec:method:domain_taxonomy}
\label{sec:method:representational_absence} 


We define the \emph{distant-OOD regime} for a frozen encoder $\phi_v$ as the setting in which the encoder's representation space fails to support the target classification task along one or both of two axes: (i)~the target images occupy regions of the representation space poorly covered by the pretraining distribution, and (ii)~the encoder's internal grouping of target images bears little correspondence to class boundaries. This characterization depends only on the frozen encoder and the target data, not on any downstream adaptation procedure. To make it measurable, we introduce two complementary metrics computed entirely from frozen vision encoder embeddings and ground-truth class labels.


\noindent \textbf{Feature-Space Divergence.} For a target dataset and a reference natural-image distribution (ImageNet-1K), embeddings are extracted from the frozen vision encoder $\phi_v$ and the Multi-Kernel Maximum Mean Discrepancy (MK-MMD)~\cite{long2015learning} is computed between the two sets of vision embeddings. MK-MMD directly measures the feature-space divergence $\delta(\mathcal{P}_{\mathrm{train}}, \mathcal{P}_{\mathrm{target}})$ from \cref{prop:asymmetry}(i). Both distributions comprise vision encoder outputs only; no text or class-label information enters the computation. A high MK-MMD indicates that the encoder maps target images to regions of $\mathbb{R}^d$ that are poorly populated by pretraining data, reducing the discriminative utility of the resulting representations.


\noindent \textbf{Latent Cluster--Label Alignment.} The frozen vision encoder embeddings of the target dataset are clustered via $k$-means ($k{=}C$, number of target classes, $n_{\mathrm{init}}{=}10$, Euclidean distance), and the Normalized Mutual Information (NMI) between the resulting cluster assignments and ground-truth class labels is computed. NMI $\in [0, 1]$ quantifies whether the encoder's internal grouping of target images already reflects task-relevant class boundaries, approximating the discriminative capacity $D(f, \mathcal{P}_{\mathrm{target}})$ from \cref{prop:asymmetry}. A dataset falls in the distant-OOD regime when it exhibits either high MK-MMD or low NMI; each axis captures a structurally distinct failure mode, as illustrated by Pok\'emon, which exhibits low MK-MMD (visually familiar cartoon images) yet low NMI (type labels have no grounded correspondence in the encoder's latent structure). Cross-model robustness and clustering-method sensitivity analyses appear in Supp.~\S\ref{appendix:select_distant_ood}.


MK-MMD and NMI identify when discriminative capacity $D(f)$ has collapsed. Descriptive capacity $G(f) \geq G_0 > 0$ is guaranteed by \cref{prop:asymmetry}(ii) independently of domain shift, and the trait extraction results in \cref{section:experiments} empirically corroborate this. IVL helps only when the regime is also \emph{trait-separable}, that is, when the primitives the decoder can articulate differ across classes (\cref{section:experiments:failure_modes}).


Projecting datasets onto these two axes yields the landscape in \cref{fig:teaser-a}. Standard FGVC benchmarks occupy the low-divergence, high-alignment quadrant where parametric methods succeed; specialized domains fall into the high-divergence or low-alignment quadrants. Regime boundaries are set by a two-stage variant of Otsu's method~\cite{4310076} that thresholds MK-MMD first and then NMI on the low-divergence subset (Supp.~\S\ref{appendix:select_distant_ood}), yielding $\tau_{\mathrm{MMD}} = 0.49$ and $\tau_{\mathrm{NMI}} = 0.42$, derived purely from frozen encoder geometry. Dataset groupings remain stable as $\tau_{\mathrm{NMI}}$ varies across $[0.30, 0.60]$ (Supp.~\S\ref{appendix:select_distant_ood}). Near-OOD datasets show consistently positive 1-shot SFT gains (mean $\Delta = {+}18.6\%$), while distant-OOD datasets show near-zero mean with high variance ($\Delta$ from $-27.5\%$ to $+36.9\%$), consistent with \cref{prop:gradient_futility}.

\subsection{Framework Overview}
\label{sec:method:overview}
\begin{figure}[t!]
    \centering
    \includegraphics[width=\textwidth]{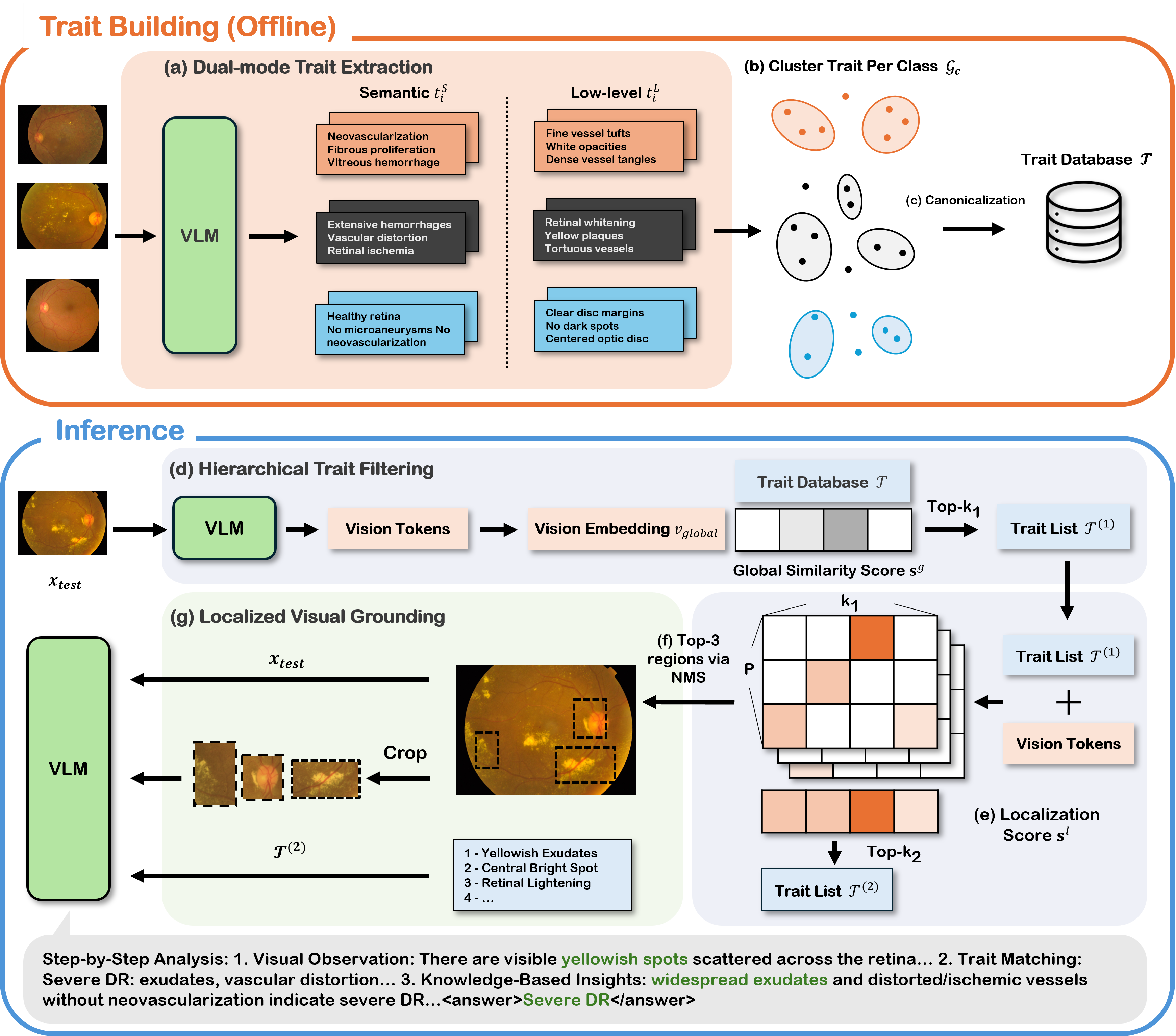}
    \vspace{-1.5em}
    \caption{
        \textbf{Offline Trait Building}: (a) The system extracts semantic ($t_i^S$) and low-level ($t_i^L$) traits from support images (b) clusters them per class and (c) assigns canonical names, and  stores them in database $\mathcal{T}$. \textbf{Inference}: Given test image $x_{\text{test}}$, hierarchical filtering is performed: (d) compute $\mathbf{s}^g = \hat{\mathbf{E}}_{\mathcal{T}}^{\!\top}\, \hat{\mathbf{v}}_{\mathrm{global}}$ to select top-$k_1$ traits $\mathcal{T}^{(1)}$, (e) compute image patch-trait attention scores for localized grounding while selecting top-$k_2$ traits $\mathcal{T}^{(2)}$ and (f) extracting top-3 regions via NMS, and (g) prompt the VLM with the original image, cropped regions, and the refined trait set $\mathcal{T}^{(2)}$ for final classification.
    }
\vspace{-1.5em}
    \label{fig:overview}
\end{figure}

Guided by the theoretical analysis above, IVL instantiates the descriptive-discriminative asymmetry (\cref{prop:asymmetry}) in a two-stage pipeline. \Cref{fig:overview} provides an architectural overview. The first stage, \textbf{Offline Trait Building} (\cref{sec:method:trait_building}), prompts the language decoder in two complementary modes to extract fine-grained visual traits from support images and consolidates them into a compact per-class trait database~$\mathcal{T}$. The second stage, \textbf{Inference} (\cref{section:method:inference}), classifies a query image by identifying which traits in $\mathcal{T}$ are visually grounded in the query through hierarchical filtering, bypassing the discriminative bottleneck identified in \cref{prop:gradient_futility}. The entire pipeline requires no parameter updates and produces interpretable predictions traceable to specific traits and their spatial grounding.

\subsection{Trait Building Stage}
\label{sec:method:trait_building}

To operationalize the surviving descriptive capacity established in \cref{prop:asymmetry}, IVL constructs an explicit trait vocabulary from the support set through three phases illustrated in \cref{fig:overview}~(a--c): dual-mode trait extraction, semantic clustering with canonical naming, and quality filtering.

\subsubsection{Dual-Mode Trait Extraction}
\label{section:method:trait_building:dual_trait_extraction}

The dual-mode decomposition follows directly from the capacity structure in \cref{prop:asymmetry}. A VLM's descriptive capacity $G(f)$ encompasses two qualitatively different resources. The first is residual domain knowledge, which allows the model to produce semantically rich descriptions when it retains partial familiarity with the target domain. The second is the domain-agnostic visual vocabulary guaranteed by \cref{prop:asymmetry}(ii), encompassing primitive descriptions of colors, textures, and shapes that remain available regardless of domain shift. IVL targets each resource with a dedicated extraction mode, ensuring that trait quality degrades gracefully rather than collapsing entirely as the target domain moves further from the pretraining distribution.

\textbf{Semantic traits} $t^S_i$ are elicited through a prompt $p^S$ that requests knowledge-based descriptions, leveraging whatever domain understanding the model retains. \textbf{Low-level traits} $t^L_i$ are elicited through a prompt $p^L$ that constrains responses to primitive visual features and explicitly prohibits category labels, targeting the $G_0 > 0$ floor from \cref{prop:asymmetry}(ii). Rather than producing the generic label ``retinal tissue,'' a low-level prompt yields descriptions such as ``white wrinkled regions in center'' or ``dark void between horizontal bands.'' This decomposition distinguishes IVL from prior attribute-based methods~\cite{menon2022visual, pratt2023does, tian2024argue, roth2023waffling}, which generate descriptors entirely from pretrained semantic knowledge without examining the actual support images. When that semantic knowledge is absent, such methods produce spurious or generic descriptors. IVL instead extracts traits from visual observations of the support set, grounding every descriptor in image content rather than in the model's prior beliefs.

Both prompts follow a unified, domain-agnostic template; the only dataset-specific tokens are the domain and class names. The extraction performs $K$ independent runs per image (typically $K{=}5$), each generating $6$--$8$ traits per mode following the self-consistency principle~\cite{wang2022self}. Semantic trait coverage degrades with feature-space divergence, whereas low-level coverage remains stable because primitive visual vocabulary is domain-agnostic (\cf \cref{prop:asymmetry}). The combined set $\mathcal{T}^S \cup \mathcal{T}^L$ therefore achieves coverage strictly higher than either mode alone whenever the two sets are not fully redundant, which is the typical case for distant-OOD domains. Full prompt templates appear in Supp.~\S\ref{appendix:reproducibility}.


\vspace{-1.0em}

\subsubsection{Clustering, canonicalization, and refinement.}
Raw traits are lexically diverse and redundant. For each class $c$, extracted traits are embedded using Sentence Transformers~\cite{reimers2019sentence} and clustered via HDBSCAN~\cite{mcinnes2017hdbscan}, with isolated traits retained as rare discriminative cues. Each cluster is assigned a canonical name through VLM compositional reasoning (\eg, ``red legs'' and ``crimson arms'' become ``red limbs''). Sentence Transformers are preferred over VLM text encoders because the latter over-compress textual semantic structure (Supp.~\S\ref{appendix:ablation}). Canonicalized semantic traits undergo relevance filtering through VLM common-sense reasoning; low-level traits bypass this step, as the model cannot reliably assess primitive features in unfamiliar domains (\cf \cref{prop:asymmetry}). Final trait embeddings are computed using the VLM's text encoder, yielding $\mathbf{E}_{\mathcal{T}} = [\mathbf{e}_1, \ldots, \mathbf{e}_{|\mathcal{T}|}] \in \mathbb{R}^{d \times |\mathcal{T}|}$.

\subsection{Inference Stage}
\label{section:method:inference}
Given $\mathcal{T}$ and its embedding matrix $\mathbf{E}_{\mathcal{T}}$, inference classifies a query image by identifying visually grounded traits through hierarchical filtering.

\subsubsection{Global Trait Retrieval}
\label{section:method:inference:global_trait}
As illustrated in \cref{fig:overview}~(d), given test image $x_{\mathrm{test}}$, the VLM vision encoder produces $P$ patch tokens $\{v_1, \ldots, v_P\}$, each $v_i \in \mathbb{R}^d$. A global embedding $\mathbf{v}_{\mathrm{global}} = \frac{1}{P}\sum_{i=1}^{P} v_i$ is obtained via mean pooling, and global similarity scores $\mathbf{s}^g = \hat{\mathbf{E}}_{\mathcal{T}}^{\!\top}\, \hat{\mathbf{v}}_{\mathrm{global}} \in \mathbb{R}^{|\mathcal{T}|}$ are computed via L2-normalized cosine similarity. The top-$k_1$ traits form $\mathcal{T}^{(1)} \subset \mathcal{T}$.

\begin{figure}[!tp]
    \centering
    \includegraphics[width=8.5cm]{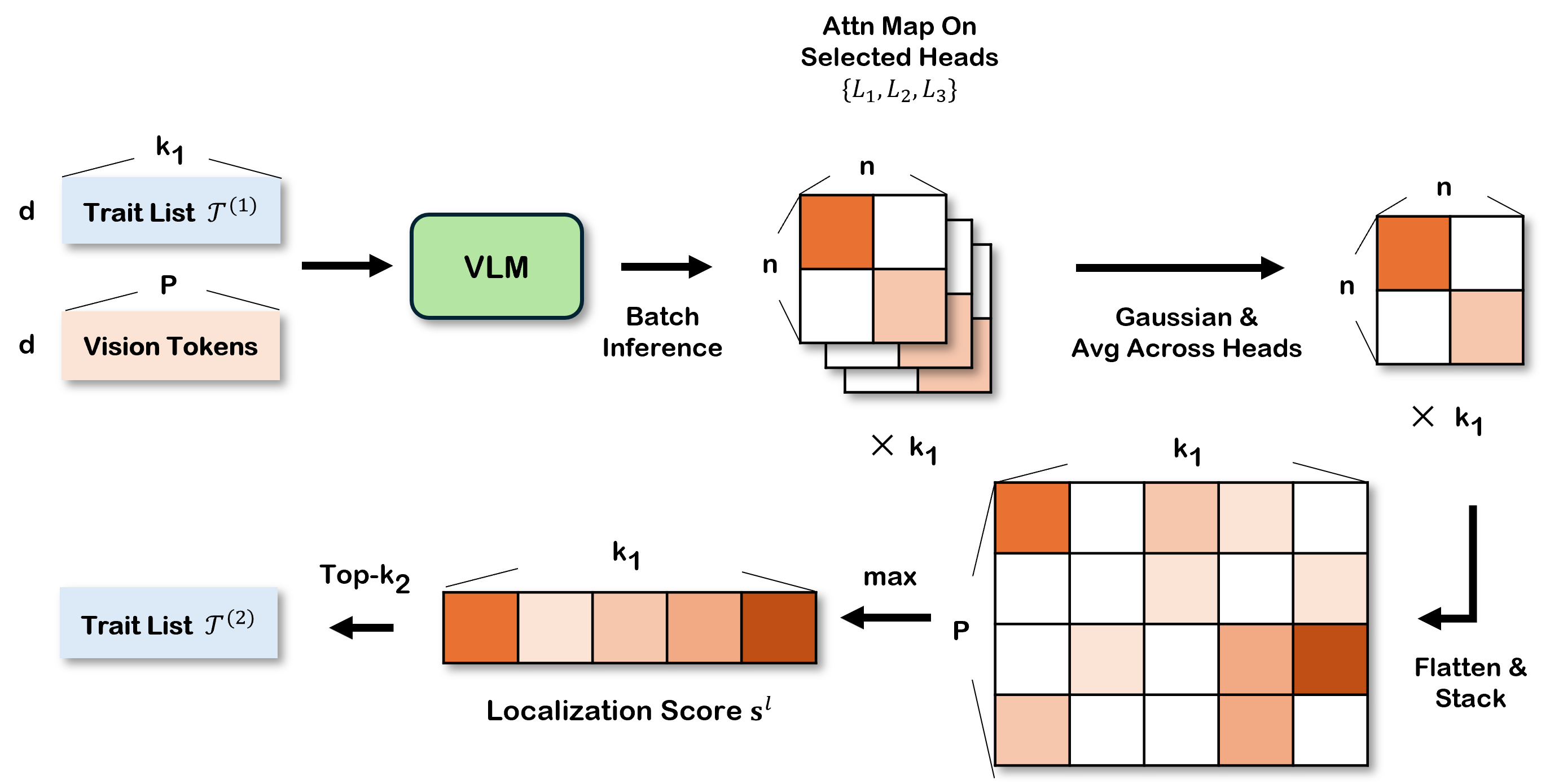}
    \vspace{-1.0em}
    \caption{
    Localized Trait Refinement. Given the test image and the globally filtered trait list $\mathcal{T}^{(1)}$, the VLM produces cross-modal attention maps from the grounding heads $\{L_1, L_2, L_3\}$. After averaging and spatial smoothing, each trait's attention map is flattened into a $P$-dimensional patch-level vector. The localization score $s^l_i$ is computed as the maximum attention over patches, and the top-$k_2$ traits are selected to form $\mathcal{T}^{(2)}$.}

    \label{fig:localization}
\end{figure}

\subsubsection{Localized Trait Refinement}
\label{section:method:inference:local_trait}
Global retrieval cannot verify spatial grounding. Following Kang~\etal~\cite{kang2025your}, IVL extracts cross-modal attention maps from three grounding heads $\{L_1, L_2, L_3\}$ (\cref{fig:localization}). For each trait $t_i \in \mathcal{T}^{(1)}$, maps are averaged and flattened into the patch-trait attention matrix $A \in \mathbb{R}^{P \times k_1}$. A localization score $s^l_i = \max_{p}\, A_{p,i}$ captures whether the trait has concentrated evidence in at least one region. The top-$k_2$ traits form $\mathcal{T}^{(2)} \subset \mathcal{T}^{(1)}$.

\subsubsection{Visual region extraction and classification.}
Attention maps from $\mathcal{T}^{(2)}$ are reshaped into 2D heatmaps, smoothed with a Gaussian filter, and bounding boxes are extracted around the top-3 local maxima (\cref{fig:overview}~(f)). The VLM receives three complementary inputs for classification: the original image $x_{\mathrm{test}}$, the refined 
trait list $\mathcal{T}^{(2)}$ as textual context, and the three cropped regions for focused inspection (\cref{fig:overview}~(g)). By design, classification routes through the decoder's preserved descriptive capacity $G(f)$ while bypassing the encoder's degraded discriminative features (\cref{prop:asymmetry,prop:gradient_futility}). The entire pipeline is training-free and produces interpretable predictions traceable to traits and spatial grounding.

\section{Experimental Results}
\label{section:experiments}

\subsection{Experimental Setup}
\label{section:experiments:setup}

IVL is designed for the distant-OOD regime, where pretrained representations lack discriminative structure. Benchmarks are selected through the two-metric protocol of \cref{sec:method:representational_absence}: datasets exhibiting high MK-MMD or low NMI relative to Otsu-derived thresholds are retained, yielding 18 classification tasks: Retinal OCT~\cite{obuli_sai_naren_2021}, WM811k~\cite{wm811k_dataset}, MVTec~AD~\cite{bergmann2021mvtec} (15 categories evaluated independently), and Pok\'{e}mon. The Pok\'{e}mon benchmark exhibits low visual divergence yet low semantic alignment (elemental type labels have no grounded correspondence in the VLM's label space), isolating semantic collapse from visual unfamiliarity. Full per-dataset metrics appear in Supp.~\S\ref{appendix:select_distant_ood}. MVTec~AD serves as a distant-OOD stress test for general-purpose generative VLMs, not as a comparison with specialized anomaly detection pipelines that leverage task-specific architectures unavailable to general-purpose models.

Baselines span the major adaptation paradigms: MajorVote ($k{=}5$) for test-time scaling, ICL for few-shot demonstration, CuPL~\cite{pratt2023does} and Menon~\etal~\cite{menon2022visual} for attribute-based zero-shot classification, Tip-Adapter~\cite{zhang2021tip} and CoOp~\cite{coop} for adapter-based and prompt-tuning approaches, Visual-RFT~\cite{visualrft} for RL-based few-shot learning, and SFT/SFT+LoRA for gradient-based fine-tuning. DIFO~\cite{tang2024source} is evaluated in Supp.~\S\ref{appendix:non_generative_baselines}. All methods share the same VLM backbone (Qwen2.5-VL-7B or LLaVA-1.6-Mistral-7B) with a fixed few-shot support set across all methods. Additional details appear in Supp.~\S\ref{appendix:reproducibility}.

\begin{table}[t!]
    \centering
    \caption{Experimental results on distant-OOD benchmarks.
    Zero-shot and description-based methods (CuPL, Menon) use 0-shot.
    \best{Bold}: best per dataset within each backbone.
    \second{Underline}: second best. All few-shot results use $n{=}1$ unless marked 8-shot.}
    \vspace{-1.0em}
    \label{tab:main_results}
    \setlength{\aboverulesep}{0pt}
    \setlength{\belowrulesep}{0pt}
    \renewcommand{\arraystretch}{1.2}
    \resizebox{\textwidth}{!}{%
    \begin{tabular}{@{}l cccccccccc>{\columncolor{teal!8}}c@{}}
        \toprule
        \rowcolor{gray!20}
        \textbf{Dataset} &
        \textbf{Zero-Shot} &
        \textbf{MajorVote} &
        \textbf{ICL} &
        \textbf{CuPL\cite{pratt2023does}} &
        \textbf{Menon \etal\cite{menon2022visual}} &
        \textbf{V-RFT\cite{visualrft}} &
        \textbf{SFT+LoRA} &
        \textbf{SFT} &
        \textbf{Tip-Adapter\cite{zhang2021tip}} &
        \textbf{CoOp\cite{coop}} &
        \cellcolor{teal!8}\textbf{IVL (Ours)} \\
        \midrule
        \multicolumn{12}{l}{\cellcolor{gray!5}\textit{Qwen2.5-VL-7B}} \\
        Pok\'{e}mon     & 47.47 & 48.23 & 22.00 & 15.28 & 25.77 & \second{48.77} & 45.59 & 45.52 & 9.72  & 20.01 & \best{50.30} \\
        Retinal OCT     & 14.36 & \second{16.39} & 13.75 & 12.50 & 11.46 & 14.00 & 13.69 & 14.77 & 12.50 & 13.39 & \best{20.83} \\
        WM811k          & 13.13 & \second{13.73} & 12.41 & 12.77 & 12.29 & 9.52  & 9.16  & 12.41 & 11.08 & 9.30  & \best{16.51} \\
        \cmidrule{1-12}
        \multicolumn{12}{l}{\textit{~~8-shot}} \\
        Pok\'{e}mon     & -- & -- & 34.24 & -- & -- & 49.09 & \second{50.09} & 49.94 & 6.94 & 29.17 & \best{52.20} \\
        Retinal OCT     & -- & -- & 19.54 & -- & -- & 13.14 & 14.00 & 16.00 & 20.00 & \best{39.92} & \second{23.21} \\
        WM811k          & -- & -- & 14.22 & -- & -- & 8.92 & 9.16 & 15.90 & 11.08 & \second{17.59} & \best{18.43} \\
        \midrule
        \multicolumn{12}{l}{\textit{MVTec AD -- per-category (Qwen2.5-VL, 1-shot)}} \\
        \quad Bottle     & 50.6 & \second{54.43} & 36.7 & 26.6 & \best{57.0} & 51.9 & 53.2 & 48.1 & 31.65 & 39.2 & 46.8 \\
        \quad Cable      & 12.8 & 11.35 & 14.9 & 9.2  & 12.8 & 9.2  & 12.8 & \second{41.1} & 11.35 & 8.5 & \best{45.4} \\
        \quad Capsule    & \second{46.0} & \best{47.62} & 25.4 & 14.3 & 35.7 & 42.1 & 40.5 & 28.6 & 23.02 & 17.5 & 44.4 \\
        \quad Carpet     & 53.2 & \best{61.26} & 20.7 & 16.2 & 40.5 & 55.0 & 58.6 & 55.0 & 21.62 & 14.4 & \second{60.4} \\
        \quad Grid       & 36.1 & 34.72 & 23.6 & 20.8 & 34.7 & \best{52.8} & \best{52.8} & 40.3 & 27.78 & 15.3 & 43.1 \\
        \quad Hazelnut   & 61.9 & 58.10 & 41.9 & 16.2 & 37.1 & 59.1 & 60.0 & \best{77.1} & 30.48 & 21.9 & \second{72.4} \\
        \quad Leather    & 44.9 & 42.37 & 18.6 & 13.6 & 24.6 & 39.0 & 40.7 & \best{60.2} & 28.81 & 12.7 & \second{48.3} \\
        \quad Metal Nut  & \best{47.3} & 37.27 & 26.4 & 21.8 & 28.2 & 37.3 & 36.4 & 20.9 & 17.27 & 37.3 & \second{40.9} \\
        \quad Pill       & 15.1 & 16.35 & 13.8 & 11.3 & 5.0  & 13.8 & 12.6 & \second{20.8} & 11.95 & 11.9 & \best{22.0} \\
        \quad Screw      & 26.0 & \second{29.22} & 20.1 & 14.3 & 26.0 & 23.4 & 23.4 & 18.8 & 27.27 & 20.8 & \best{33.1} \\
        \quad Tile       & 58.6 & \second{62.16} & 27.0 & 13.5 & 50.5 & 46.9 & 47.8 & 49.6 & 25.23 & 55.0 & \best{65.8} \\
        \quad Toothbrush & 55.0 & 60.00 & 52.5 & \best{72.5} & \best{72.5} & 57.5 & 55.0 & 27.5 & \best{72.5} & 45.0 & 65.0 \\
        \quad Transistor & 18.9 & 11.58 & 10.5 & 9.5  & 9.5  & 13.7 & 13.7 & \second{55.8} & \best{62.11} & 42.1 & 11.6 \\
        \quad Wood       & \second{60.3} & 57.53 & 46.6 & 30.1 & 45.2 & 42.5 & 41.1 & 58.9 & 24.66 & 24.7 & \best{65.8} \\
        \quad Zipper     & 19.6 & 25.17 & 11.2 & 11.2 & 13.3 & \best{28.0} & \best{28.0} & 17.5 & 13.99 & 16.8 & 26.6 \\
        \midrule
        \quad \textbf{MVTec Mean} & 40.4 & 40.61 & 26.0 & 20.1 & 32.8 & 38.1 & 38.4 & \second{41.3} & 28.65 & 25.54 & \best{46.1} \\
        \midrule
        \multicolumn{12}{l}{\cellcolor{gray!5}\textit{LLaVA-1.6-Mistral-7B}} \\
        Pok\'{e}mon     & 28.00 & 33.60 & 5.37  & 15.97 & 8.29  & 37.18 & 34.76 & \second{38.15} & 31.94 & 6.94 & \best{44.60} \\
        Retinal OCT     & 10.57 & 10.96 & 11.46 & 11.68 & 12.61 & 11.18 & 31.38 & \best{34.30} & 24.36 & 12.39 & \second{33.70} \\
        WM811k          & 10.30 & 6.39 & 10.72 & 14.58 & 11.20 & 12.41 & 11.92 & 12.10 & \second{15.90} & 11.45 & \best{17.21} \\
        \cmidrule{1-12}
        \multicolumn{12}{l}{\textit{~~8-shot}} \\
        Pok\'{e}mon     & -- & -- & 4.39 & -- & -- & 41.23 & \second{41.47} & 33.42 & 34.03 & 39.92 & \best{48.41} \\
        Retinal OCT     & -- & -- & 17.50 & -- & -- & 7.23 & 26.29 & \best{35.20} & 12.50 & 12.89 & \second{31.33} \\
        WM811k          & -- & -- & 12.17 & -- & -- & 4.00 & 8.92 & 12.81 & 10.96 & \best{24.11} & \second{19.13} \\
        \midrule
        \multicolumn{12}{l}{\textit{MVTec AD -- per-category (LLaVA-1.6, 1-shot)}} \\
        \quad Bottle     & \second{47.0} & 39.24 & 29.1 & 26.6 & 45.6 & 32.4 & 33.4 & 42.6 & \best{60.76} & 17.72 & 34.2 \\
        \quad Cable      & 5.4  & \best{39.72} & 12.1 & 6.4  & 7.8  & 8.6  & 15.2 & 14.2 & \second{25.53} & 8.51 & 16.4 \\
        \quad Capsule    & 21.4 & 18.25 & 23.8 & 17.5 & 17.5 & \second{44.2} & 38.1 & \best{46.5} & 19.84 & 15.08 & 32.2 \\
        \quad Carpet     & 14.4 & \best{41.44} & 29.7 & 16.2 & 16.2 & 19.0 & 13.4 & 16.4 & 14.41 & 14.41 & \second{34.2} \\
        \quad Grid       & 10.9 & \best{34.72} & \second{33.3} & 23.6 & 29.2 & 9.5  & 28.9 & 27.4 & 15.28 & 18.06 & 25.0 \\
        \quad Hazelnut   & \second{40.0} & 34.29 & 35.2 & 15.2 & 41.9 & 38.1 & 33.7 & 27.1 & 26.67 & 16.19 & \best{47.8} \\
        \quad Leather    & 36.1 & 40.68 & 26.3 & 15.3 & 0.9  & 62.3 & \second{68.3} & \best{73.1} & 25.42 & 11.86 & 58.1 \\
        \quad Metal Nut  & \second{31.8} & 30.91 & 25.5 & 20.0 & 19.1 & 20.0 & 21.4 & 38.4 & \best{54.55} & 19.09 & 31.7 \\
        \quad Pill       & 13.8 & 13.84 & 13.8 & 14.5 & 5.0  & 42.1 & \best{58.3} & \second{54.2} & 21.38 & 16.35 & 48.7 \\
        \quad Screw      & 21.7 & 26.62 & 21.4 & 15.6 & 14.3 & 24.9 & \second{34.1} & \best{38.8} & 18.83 & 15.58 & 28.7 \\
        \quad Tile       & 33.7 & 43.24 & 26.1 & 13.5 & 42.3 & 48.4 & 22.7 & \second{71.3} & 29.73 & 15.32 & \best{76.7} \\
        \quad Toothbrush & 35.0 & 62.50 & 60.0 & 37.5 & 30.0 & 45.0 & 20.0 & 40.0 & \best{72.5} & \second{67.5} & \best{72.5} \\
        \quad Transistor & \second{39.3} & 14.74 & 17.9 & 9.5  & 31.6 & 26.1 & 31.7 & 30.4 & 14.74 & 9.47 & \best{43.2} \\
        \quad Wood       & 8.3  & 46.58 & 32.9 & 27.4 & 28.8 & \best{64.2} & 12.2 & 16.8 & 26.03 & 17.81 & \second{54.8} \\
        \quad Zipper     & 38.4 & 21.68 & 11.2 & 11.2 & 11.2 & \second{41.2} & 32.7 & 31.1 & 10.49 & 13.29 & \best{48.9} \\
        \midrule
        \quad \textbf{MVTec Mean} & 26.48 & 33.90 & 26.6 & 18.0 & 22.8 & 35.1 & 31.0 & \second{37.9} & 29.08 & 18.42 & \best{43.54} \\
        \bottomrule
    \end{tabular}%
    }\vspace{-1.5em}
\end{table}
\subsection{Quantitative Results}
\label{section:experiments:quantitative}

The most striking pattern in \cref{tab:main_results} is the failure of gradient-based adaptation on distant domains. On Retinal OCT under 1-shot Qwen2.5-VL, SFT achieves only $14.77\%$ ($+0.41$\,pp over zero-shot) despite full parameter updates, and Visual-RFT degrades to $14.00\%$; even at 8-shot, neither exceeds $16.00\%$. On WM811k, both SFT+LoRA and Visual-RFT fall below the zero-shot baseline, confirming the gradient futility phenomenon formalized in \cref{prop:gradient_futility}. Test-time scaling (MajorVote) and in-context learning exhibit analogous instability, with MajorVote degrading on WM811k under LLaVA ($6.39\%$ vs.\ $10.30\%$ zero-shot) despite marginal gains elsewhere. IVL avoids these failure modes by building visual knowledge from the VLM's surviving descriptive capacity, achieving the best result on Retinal OCT ($20.83\%$), WM811k ($16.51\%$), and Pok\'{e}mon ($50.30\%$).

Attribute-based methods that generate descriptors from pretrained semantic knowledge without observing target-domain images inherit the model's blind spots, as reflected by CuPL's $20.1\%$ and Menon~\etal's $32.8\%$ on MVTec~AD compared to IVL's $46.1\%$. Across all 18 tasks, IVL places in the top two for 12 under Qwen2.5-VL and 10 under LLaVA, the widest coverage of any single method under both backbones. These improvements hold under both backbones: under LLaVA-1.6, IVL achieves the highest accuracy on Pok\'{e}mon, WM811k, and MVTec~AD mean, and matches the best-performing method on Retinal OCT ($33.70\%$ vs.\ $34.30\%$) without any parameter updates.

Computational analysis (Supp.~\S\ref{sec:computational_analysis}) shows that IVL's trait building requires only a single 32 GB VRAM GPU and 67\% fewer GPU-seconds than SFT, which demands a multi-GPU cluster. Per-image inference incurs a moderate overhead, dominated by the VLM processing enriched trait-grounded prompts rather than by the retrieval pipeline itself.

\subsection{Failure Modes}
\label{section:experiments:failure_modes}
IVL's errors are interpretable from its trait rationale and fall into three modes, each arising at a different level of the pipeline.

\noindent\textbf{Reference-free vocabulary collapse (mechanism level).} Because IVL inspects each query in isolation rather than against a normal reference, benign within-tolerance variation is read as a defect and the trait dictionary collapses toward one dominant concept. The signature is a single trait cited as decisive across many queries, as on MVTec~AD Transistor, where most categories are all predicted \textit{bent\_lead} and accuracy falls to $11.6\%$. Scoring candidate traits against support references rather than in isolation would address this (Supp.~\S\ref{appendix:failure_mode}).

\noindent\textbf{Dictionary coverage failure (retrieval level).} When hierarchical filtering drops the ground-truth class from the candidate dictionary, no downstream reasoning can recover it, and the signature is a trace whose cited traits never include the true class. This accounts for $48.4\%$ of Pok\'emon errors and motivates open-vocabulary dictionary construction (Supp.~\S\ref{sec:supp:trait_alignment}).

\noindent\textbf{Trait-separability boundary (applicability level).} IVL can only succeed when the classes differ in traits the model can actually name. When they do not, there is nothing for descriptive reasoning to match on, and accuracy stays near chance. FractalDB \cite{kataoka2020pre} is the clearest case, since its labels come from the generative parameters that produced each image rather than from any visible attribute. This is not a defect IVL can fix but a genuine limit of trait-based reasoning, marking where parametric or hybrid methods are still needed (Supp.~\S\ref{appendix:failure_mode}).

\subsection{Ablation Studies}
\label{section:experiments:ablation}

\subsubsection{Dual-mode prompt design.}
Since the dual-mode prompt strategy is the central design choice distinguishing IVL from prior attribute-based methods, \cref{tab:prompt_ablation} provides a systematic per-category evaluation on MVTec~AD. Three patterns emerge. First, low-level prompting outperforms semantic prompting in 10 of 15 categories, confirming that on distant-OOD domains the VLM's primitive visual vocabulary is more reliable than its domain-specific knowledge, as predicted by \cref{prop:asymmetry}. Second, dual-mode extraction surpasses the better single mode on 12 of 15 categories, demonstrating that the two prompt strategies contribute non-redundant discriminative information. Third, the three exceptions are interpretable: Transistor suffers from vocabulary collapse (Supp.~\S\ref{appendix:failure_mode}), while Toothbrush and Zipper are visually simple categories where low-level descriptors alone suffice. This dual-mode design is the key distinction from CuPL~\cite{pratt2023does} and Menon~\etal~\cite{menon2022visual}, which rely exclusively on semantic prompting without visual grounding and underperform IVL's dual-mode results on all 15 categories.
\begin{table}[tp!]
    \centering
    \caption{Prompt design ablation on MVTec~AD (Qwen2.5-VL, 1-shot, 3-run mean per mode). Semantic: knowledge-driven prompting. Low-level: observation-driven prompting. $\Delta$: dual-mode gain over best single mode.}
    \vspace{-0.8em}
    \label{tab:prompt_ablation}
    \setlength{\aboverulesep}{0pt}
    \setlength{\belowrulesep}{0pt}
    \renewcommand{\arraystretch}{1.1}
    \resizebox{\textwidth}{!}{%
    \begin{tabular}{@{}l ccccccccccccccc c@{}}
        \toprule
        \rowcolor{gray!20}
        & \textbf{Bottle} & \textbf{Cable} & \textbf{Capsule} & \textbf{Carpet} & \textbf{Grid} & \textbf{Hazelnut} & \textbf{Leather} & \textbf{Metal Nut} & \textbf{Pill} & \textbf{Screw} & \textbf{Tile} & \textbf{Toothbrush} & \textbf{Transistor} & \textbf{Wood} & \textbf{Zipper} & \textbf{Mean} \\
        \midrule
        Semantic  & 39.2 & 14.7 & \second{35.7} & 36.9 & 34.3 & \second{52.7} & 28.0 & 26.7 & 5.9  & \second{32.3} & \second{57.7} & 51.7 & \best{15.4} & 40.2 & 17.0 & 32.6 \\
        Low-level & \second{40.5} & \second{30.3} & \second{35.7} & \second{50.8} & \second{36.6} & 46.7 & \second{39.5} & \second{31.2} & \second{10.5} & 30.3 & 52.6 & \best{72.5} & \second{14.4} & \second{47.9} & \best{29.3} & \second{37.9} \\
        \rowcolor{teal!8}
        Dual      & \best{46.8} & \best{45.4} & \best{44.4} & \best{60.4} & \best{43.1} & \best{72.4} & \best{48.3} & \best{40.9} & \best{22.0} & \best{33.1} & \best{65.8} & \second{65.0} & 11.6 & \best{65.8} & \second{26.6} & \best{46.1} \\
        \midrule
        $\Delta$  & \small{+6.3} & \small{+15.1} & \small{+8.7} & \small{+9.6} & \small{+6.5} & \small{+19.7} & \small{+8.8} & \small{+9.7} & \small{+11.5} & \small{+0.8} & \small{+8.1} & \small{\textcolor{red}{$-7.5$}} & \small{\textcolor{red}{$-3.8$}} & \small{+17.9} & \small{\textcolor{red}{$-2.7$}} & \small{+7.2} \\
        \bottomrule
    \end{tabular}%
    }
\end{table}

\begin{table}[t!]
    \centering
    \caption{Component ablation on the Pok\'{e}mon benchmark (144-sample balanced subset, Qwen2.5-VL, 1-shot). $\Delta$: change relative to full pipeline.}
    \vspace{-0.8em}
    \label{tab:ablation_main}
    \setlength{\tabcolsep}{4pt}
    \renewcommand{\arraystretch}{1.0}
    \scriptsize
    \begin{tabular}{@{}llcc@{}}
        \toprule
        \textbf{Component} & \textbf{Configuration} & \textbf{Acc.\ (\%)} & \textbf{$\Delta$} \\
        \midrule
        Extraction (\cref{sec:method:trait_building}) & Semantic only   & 32.6 & \textcolor{red}{$-6.2$} \\
                                                      & Low-level only  & 27.8 & \textcolor{red}{$-11.1$} \\
        \midrule
        Filtering (\cref{section:method:inference}) & Global only      & 39.6 & \textcolor{teal}{$+0.7$} \\
                                                     & Local only       & 37.5 & \textcolor{red}{$-1.4$} \\
                                                     & No filtering     & 36.8 & \textcolor{red}{$-2.1$} \\
        \midrule
        Quality (\cref{sec:method:trait_building})   & w/o salience     & 36.3 & \textcolor{red}{$-2.6$} \\
        \midrule
        Grounding (\cref{section:method:inference})  & w/o localized    & 34.7 & \textcolor{red}{$-4.2$} \\
        \midrule
        \rowcolor{teal!8}
        \multicolumn{2}{l}{\textbf{Full pipeline}}   & \textbf{38.9} & --- \\
        \bottomrule
    \end{tabular}\vspace{-1.5em}
    
\end{table}

\subsubsection{Component ablation.}

\Cref{tab:ablation_main} isolates the remaining components on the Pok\'{e}mon benchmark. The relative ranking of extraction modes reverses compared to MVTec~AD: semantic-only ($32.6\%$) outperforms low-level-only ($27.8\%$), because the VLM retains partial semantic knowledge of cartoon imagery but lacks grounded type-label associations. This reversal confirms that dual-mode extraction is necessary because the dominant failure mode varies across domains. Removing visual grounding degrades accuracy by $-4.2$\,pp, confirming that the VLM requires direct observation of localized regions where discriminative traits manifest. Disabling salience filtering costs $-2.6$\,pp, and removing all hierarchical filtering degrades accuracy by $-2.1$\,pp; the marginal global-only gain ($+0.7$\,pp) reflects Pok\'{e}mon's whole-body structure, where local refinement is less critical than on spatially complex datasets. Detailed experiments appear in Supp.~\S\ref{appendix:ablation}.

\renewcommand{\arraystretch}{1}

\section{Conclusion}
\label{section:conclusion}

This paper introduces IVL, a training-free framework that addresses VLM failures on distant-OOD tasks through trait-based reasoning rather than parameter adaptation. The central insight is that these failures stem from structurally absent visual primitives, not from optimization limitations; gradient-based adaptation cannot construct discriminative features that the encoder never learned. IVL constructs such features explicitly through dual-mode trait extraction and hierarchical filtering, grounding each classification decision in spatially localized visual evidence. Experiments across distant-OOD benchmarks confirm that this approach succeeds where parametric methods fail, establishing trait-based reasoning as a principled and immediately deployable path for VLM adaptation in specialized domains. Extending trait-based reasoning toward open-set recognition, where the target class may lie outside the support set, and compositional generalization is a promising direction for future work.

\section*{Acknowledgement}
\label{section:acknowledgement}
The authors gratefully acknowledge the support from the National Science and Technology Council (NSTC) in Taiwan under grant numbers NSTC 113-2221-E-007-105-MY3, NSTC 114-2634-F-002-004, NSTC 114-2221-E-002-069-MY3, NSTC 113-2221-E-002-212-MY3, NSTC 115-2634-F-002-012, and NSTC 115-2218-E-A49-010, as well as the support from the Academia Sinica Scholar Award (ASSA) under grant number AS-ASSA-115-02, NTU Artificial Intelligence Center of Research Excellence, and Taiwan Centers of Excellence in Artificial Intelligence. This research was also supported by the NVIDIA Academic Grant Program. The authors would also like to express their appreciation for the hardware grant donation of the GPUs from NVIDIA Corporation and NVIDIA AI Technology Center (NVAITC) used in this work. Furthermore, the authors extend their gratitude to the National Center for High-Performance Computing (NCHC) for providing computational and storage resources. The authors also thank the NVIDIA Taipei-1 supercomputer for providing essential computing resources.

\bibliographystyle{splncs04}
\bibliography{references}
\clearpage
\appendix
\renewcommand{\thesection}{\Alph{section}}
\renewcommand{\thesubsection}{\thesection.\arabic{subsection}}
\renewcommand{\theHsection}{appendix.\Alph{section}}
\section*{Supplementary Material}
\addcontentsline{toc}{section}{Supplementary Material}
\section*{Supplementary Overview}
This supplementary document provides theoretical derivations, empirical analyses, and implementation details supporting the main paper. It is organized as follows:

\begin{itemize}[nosep,leftmargin=*]
\item \hyperref[appendix:theory]{\textbf{\S A}}: Theoretical derivations for Propositions~1--2 and the dual-mode complementarity analysis.
\item \hyperref[appendix:select_distant_ood]{\textbf{\S B}}: Detailed per-dataset MK-MMD/NMI measurements, cross-model robustness, and threshold sensitivity.
\item \hyperref[sec:supp:descriptive_capacity]{\textbf{\S C}}: Trait extraction statistics across the OOD spectrum.
\item \hyperref[sec:supp:trait_alignment]{\textbf{\S D}}: Trait-class alignment analysis confirming that IVL's decisions are grounded in trait evidence.
\item \hyperref[appendix:ablation]{\textbf{\S E}}: Ablation studies for each pipeline component.
\item \hyperref[appendix:reproducibility:dataset]{\textbf{\S F}}: Dataset descriptions and human evaluation protocol.
\item \hyperref[appendix:failure_mode]{\textbf{\S G}}: Failure mode analysis.
\item \hyperref[sec:computational_analysis]{\textbf{\S H}}: Computational cost breakdown comparing IVL, SFT, and ICL.
\item \hyperref[appendix:non_generative_baselines]{\textbf{\S I}}: Non-generative VLMs baseline comparisons.
\item \hyperref[appendix:reproducibility]{\textbf{\S J}}: Reproducibility details (prompts, baseline configurations, hyperparameters).
\item \hyperref[appendix:qualitative]{\textbf{\S K}}: Qualitative results on IVL vs Visual-RFT.

\end{itemize}
\newpage
\section*{Notation Reference}
\label{appendix:supp_notation}
\vspace{-2.5em}
\begin{table}[H]
\centering
\caption{Summary of notation.}
\vspace{-1em}
\label{tab:notation}
\renewcommand{\arraystretch}{1.1}
\resizebox{\linewidth}{!}{%
\begin{tabular}{@{}l c l@{}}
\toprule
\textbf{Category} & \textbf{Symbol} & \textbf{Description} \\
\midrule
\textit{Problem Setting}
& $f$ & Generative VLM (composed of $\phi_v$ and $\phi_l$) \\
& $\phi_v$ & Vision encoder \\
& $\phi_l$ & Language decoder \\
& $\mathcal{P}_{\mathrm{train}}$ & Pretraining distribution \\
& $\mathcal{P}_{\mathrm{target}}$ & Target distribution \\
& $\mathcal{D} = \{(x_i, y_i)\}_{i=1}^{n \times C}$ & Few-shot support set \\
& $C$ & Number of target classes \\
& $n$ & Number of support examples per class \\
& $x_i,\; y_i$ & Image instance and its class label \\
& $x_q \;/\; x_{\mathrm{test}}$ & Query / test image \\
\midrule
\textit{Trait-Based Reasoning}
& $t$ & A single trait (natural-language visual attribute) \\
& $t^S_j,\; t^L_j$ & Semantic and low-level traits \\
& $p^S,\; p^L$ & Prompts for semantic and low-level extraction \\
& $\mathcal{T} = \bigcup_{c=1}^{C} \mathcal{T}_c$ & Full trait database (union over all classes) \\
& $\mathcal{T}_c$ & Per-class trait set for class $c$ \\
& $\mathcal{G}_c = \{G_1^c, G_2^c, \ldots\}$ & Trait clusters for class $c$ (from HDBSCAN) \\
& $K$ & Number of independent extraction runs per image \\
& $\mathbf{e}_t \in \mathbb{R}^d$ & Embedding of trait $t$ (from VLM text encoder) \\
& $\mathbf{E}_{\mathcal{T}} \in \mathbb{R}^{d \times |\mathcal{T}|}$ & Trait embedding matrix \\
& $d$ & Embedding dimension \\
\midrule
\textit{Hierarchical Inference}
& $\{v_1, \ldots, v_P\}$ & Patch-level visual embeddings ($v_i \in \mathbb{R}^d$) \\
& $\mathbf{v}_{\mathrm{global}}$ & Global image embedding (mean-pooled patches) \\
& $P$ & Number of patch tokens \\
& $\mathbf{s}^g \in \mathbb{R}^{|\mathcal{T}|}$ & Global cosine-similarity scores \\
& $k_1$ & Number of traits retained after global filtering \\
& $\mathcal{T}^{(1)} \subset \mathcal{T}$ & Globally filtered trait set \\
& $\{L_1, L_2, L_3\}$ & Cross-modal grounding attention heads \\
& $A \in \mathbb{R}^{P \times k_1}$ & Patch-trait attention matrix \\
& $s^l_i$ & Localization score for trait $t_i$ ($\max_p A_{p,i}$) \\
& $k_2$ & Number of traits retained after local refinement \\
& $\mathcal{T}^{(2)} \subset \mathcal{T}^{(1)}$ & Locally refined trait set \\
\midrule
\textit{Theoretical Analysis}
& $D(f, \mathcal{P}_{\mathrm{target}})$ & Discriminative capacity: $I(\phi_v(x);\, y)$ under $\mathcal{P}_{\mathrm{target}}$ \\
& $G(f, \mathcal{P}_{\mathrm{target}})$ & Descriptive capacity: expected recall of grounded attributes \\
& $G_0$ & Domain-shift-independent lower bound on $G$ \\
& $\delta(\mathcal{P}_{\mathrm{train}}, \mathcal{P}_{\mathrm{target}})$ & Feature-space divergence (MK-MMD) \\
& $\Phi$ & Subspace spanned by training-distribution representations \\
& $w^\dagger;\; w^\dagger_\parallel,\; w^\dagger_\perp$ 
& Squared-loss population minimizer and its $\Phi/\Phi^\perp$ components \\
& $\mathcal{R}_{\mathrm{sq}}(\cdot)$ & Squared-loss risk  \\
\midrule
\textit{Evaluation Metrics}
& MK-MMD & Multi-Kernel Maximum Mean Discrepancy \\
& NMI & Normalized Mutual Information (cluster--label alignment) \\
& $\tau_{\mathrm{MMD}},\; \tau_{\mathrm{NMI}}$ & Otsu-derived regime boundary thresholds \\
\bottomrule
\end{tabular}
}
\end{table}

\section{Theoretical Derivations}
\label{appendix:theory}

This section provides the formal arguments supporting Propositions~1 and~2 and Remark~1 stated in the main text. The derivations formalize intuitions about VLM behavior in distant-OOD regimes under clearly stated assumptions. These are intended as principled analyses rather than novel mathematical theorems; the assumptions are designed to be realistic for modern VLMs such as Qwen2.5-VL and LLaVA.

\subsection{Proof of Proposition~1: Descriptive-Discriminative Asymmetry}
\label{appendix:proof_p1}

\paragraph{Setup and definitions.}
Consider a VLM $f = (\phi_v, \phi_l)$ with vision encoder $\phi_v : \mathcal{X} \to \mathbb{R}^d$ and language decoder $\phi_l$, pretrained on distribution $\mathcal{P}_{\mathrm{train}}$. The target distribution $\mathcal{P}_{\mathrm{target}}$ defines a classification task with label space $\mathcal{Y} = \{1, \ldots, C\}$.

\begin{definition}[Discriminative Capacity]
\label{def:discriminative}
The discriminative capacity of $f$ on $\mathcal{P}_{\mathrm{target}}$ is defined as the mutual information between the encoder's representation and the task label:
\begin{equation}
    D(f, \mathcal{P}_{\mathrm{target}}) \;=\; I\bigl(\phi_v(x);\, y\bigr), \quad (x, y) \sim \mathcal{P}_{\mathrm{target}}.
\end{equation}
\end{definition}

\begin{definition}[Descriptive Capacity]
\label{def:descriptive}
Let $\mathcal{A} = \{a_1, a_2, \ldots, a_M\}$ be a reference set of $M$ primitive visual attributes spanning general categories (colors, textures, shapes, spatial relations). For an image $x$, let $\mathcal{A}(x) \subseteq \mathcal{A}$ denote the set of attributes that are visually present, and let $\hat{\mathcal{A}}(x) = \mathrm{Extract}(\phi_l, x)$ denote the set of attributes mentioned in the VLM's generated description. The descriptive capacity is defined as
\begin{equation}
    G(f, \mathcal{P}_{\mathrm{target}}) \;=\; \mathbb{E}_{x \sim \mathcal{P}_{\mathrm{target}}} \left[\frac{|\hat{\mathcal{A}}(x) \cap \mathcal{A}(x)|}{|\mathcal{A}(x)|}\right],
\end{equation}
\ie, the expected recall of visually grounded primitive attributes in the generated descriptions.
\end{definition}

\paragraph{Assumptions.}
\begin{enumerate}[label=\textbf{(A\arabic*)},leftmargin=*]
    \item \label{assum:encoder} \textbf{Encoder specialization.} The vision encoder $\phi_v$ is trained to maximize \\ $I(\phi_v(x); y_{\mathrm{train}})$ on pretraining labels $y_{\mathrm{train}}$. As a consequence, $\phi_v$ develops feature detectors that are informative for the pretraining class taxonomy but not necessarily for arbitrary target taxonomies.
    \item \label{assum:divergence} \textbf{Divergence-informativeness link.} The mutual information $I(\phi_v(x); y)$ under $\mathcal{P}_{\mathrm{target}}$ decreases as the feature-space divergence $\delta(\mathcal{P}_{\mathrm{train}}, \mathcal{P}_{\mathrm{target}})$ increases (e.g., as measured by MK-MMD on encoder representations). This holds because higher divergence implies that target images are mapped to less structured regions of the encoder's representation space, where class-conditional distributions overlap.
    \item \label{assum:vocabulary} \textbf{Broad visual vocabulary.} The language backbone $\phi_l$ was pretrained on a corpus containing descriptions of primitive visual attributes across diverse visual contexts. The vocabulary coverage $\alpha = |\mathcal{A}_{\phi_l}| / |\mathcal{A}|$, where $\mathcal{A}_{\phi_l} \subseteq \mathcal{A}$ denotes the attributes whose corresponding descriptive tokens are well-represented in the language model's training data, satisfies $\alpha \geq \alpha_0$ for some constant $\alpha_0 > 0$.
    \item\label{assum:subspace} \textbf{Representational incompleteness} \emph{(used in \S A.2).}
    Let
    \[
        \Phi \;:=\; \mathrm{span}\bigl\{\phi_v(x) : x \in \mathcal{X}_{\mathrm{train}}\bigr\} \;\subseteq\; \mathbb{R}^d
    \]
    be the linear subspace spanned by the encoder's representations under the pretraining distribution; $\Phi$ is intended as the effective pretraining-supported discriminative subspace.
    For any target task with unconstrained squared-loss minimizer $w^\dagger \in \mathbb{R}^d$, the projection residual
    satisfies $\|w^\dagger_\perp\| > 0$, where $w^\dagger_\perp = w^\dagger - \Pi_\Phi w^\dagger$.
    Furthermore, there exists $\sigma > 0$ such that
    \[
    \mathbb{E}_{x \sim \mathcal{P}_{\mathrm{target}}}
    [\langle w^\dagger_\perp / \|w^\dagger_\perp\|,\, \phi_v(x)\rangle^2] \geq \sigma^2.
    \]
    Finally, letting $\Sigma = \mathbb{E}_{x \sim \mathcal{P}_{\mathrm{target}}}[\phi_v(x)\phi_v(x)^\top]$, we assume $\Pi_\Phi \Sigma w^\dagger_\perp = 0$, i.e., representations along the missing directions $\Phi^\perp$ are uncorrelated with those along $\Phi$ under $\mathcal{P}_{\mathrm{target}}$. This condition is natural in the distant-OOD regime, where the target domain activates directions in $\Phi^\perp$ that were never structured by pretraining.
\end{enumerate}

\paragraph{Derivation of claim (i): Discriminative degradation.}
Under \ref{assum:encoder} and \ref{assum:divergence}, the data processing inequality yields
\begin{equation}
    I(\phi_v(x); y) \;\leq\; I(x; y) \;=\; H(y) - H(y \mid x),
\end{equation}
where the upper bound $H(y) - H(y|x)$ depends on the intrinsic class separability. For the encoder's representation specifically, the mutual information can be decomposed as:
\begin{equation}
    I(\phi_v(x); y) = H(\phi_v(x)) - H(\phi_v(x) \mid y).
\end{equation}
When $\delta(\mathcal{P}_{\mathrm{train}}, \mathcal{P}_{\mathrm{target}})$ is large, images from distinct target classes are mapped to overlapping regions of the encoder's feature space (since the encoder was not trained to distinguish these classes). This increases $H(\phi_v(x) \mid y)$ while $H(\phi_v(x))$ remains bounded, causing $I(\phi_v(x); y) \to 0$. In the extreme case where the encoder maps all target-domain images to a single cluster (total representational collapse), $H(\phi_v(x)) \to 0$ and the mutual information vanishes.

More precisely, let the class-conditional encoder distributions under $\mathcal{P}_{\mathrm{target}}$ be $q_c = \phi_{v\#}(\mathcal{P}_{\mathrm{target}}(\cdot \mid y = c))$ for $c = 1, \ldots, C$. When the divergence $\delta$ is large, the overlap between $q_c$ and $q_{c'}$ increases for $c \neq c'$, and the Bayes error rate $\varepsilon^* = 1 - \mathbb{E}_x[\max_c P(y = c \mid \phi_v(x))]$ approaches $1 - 1/C$ (chance level). By Fano's inequality:
\begin{equation}
\label{eq:fano}
    I(\phi_v(x); y) \;\leq\; H(y) - H_b(\varepsilon^*) - \varepsilon^* \log(C - 1),
\end{equation}
where $H_b(\cdot)$ is the binary entropy. As $\varepsilon^* \to 1 - 1/C$, the right-hand side approaches zero, establishing the degradation of $D(f, \mathcal{P}_{\mathrm{target}})$.

\paragraph{Derivation of claim (ii): Descriptive survival.}
Under \ref{assum:vocabulary}, the language decoder $\phi_l$ can articulate at least the fraction $\alpha_0$ of primitive attributes in $\mathcal{A}$ for any input, because the visual-to-language mapping for these attributes was reinforced across the full breadth of pretraining data. The key observation is that primitive attribute recognition (\eg, identifying that a region is ``bright,'' ``circular,'' or ``textured'') depends on low-level visual features that are encoded even in poorly structured regions of the representation space. Unlike class-discriminative features, which require specific feature-label associations, primitive attributes are compositional and domain-agnostic: the concept of ``redness'' does not depend on whether the red object is a car, a blood vessel, or a Pok\'emon.

Formally, for any target image $x$ with ground-truth attributes $\mathcal{A}(x)$, the expected recall satisfies:
\begin{equation}
    G(f, \mathcal{P}_{\mathrm{target}}) = \mathbb{E}_x\!\left[\frac{|\hat{\mathcal{A}}(x) \cap \mathcal{A}(x)|}{|\mathcal{A}(x)|}\right] \;\geq\; \mathbb{E}_x\!\left[\frac{|\hat{\mathcal{A}}_{\phi_l}(x) \cap \mathcal{A}(x)|}{|\mathcal{A}(x)|}\right] \;\geq\; \alpha_0 \cdot r_0,
\end{equation}
where $\hat{\mathcal{A}}_{\phi_l}(x)$ restricts to attributes within the language backbone's vocabulary coverage, and $r_0 > 0$ is the minimum recall achievable on the covered attributes. The product $G_0 = \alpha_0 \cdot r_0 > 0$ depends only on the breadth of the language backbone's visual vocabulary and the quality of the visual-to-language grounding, both of which are properties of the pretraining process and are independent of $\delta(\mathcal{P}_{\mathrm{train}}, \mathcal{P}_{\mathrm{target}})$. \qed

\paragraph{Discussion of assumptions.}
Assumption~\ref{assum:divergence} is the strongest condition: in practice, the relationship between feature-space divergence and mutual information is monotonic only on average, and individual datasets may exhibit non-monotonic behavior due to accidental alignment of target features with pretraining features. The Pok\'emon benchmark illustrates a boundary case: low MK-MMD (visual familiarity) but low NMI (semantic misalignment), where the encoder preserves visual similarity structure without encoding type-based class boundaries. Assumption~\ref{assum:vocabulary} is well-supported empirically: modern VLMs trained on web-scale image-text pairs encounter a broad range of visual attributes during pretraining, and their descriptive capability on novel domains has been documented in multiple studies~\cite{yin2024survey}. The bound $G_0$ is conservative; in practice, VLMs generate substantially richer descriptions than the minimum guarantee suggests.

\subsection{Proof of Proposition~2: Gradient Futility under Representational Absence}

We first establish that $w^\dagger_\parallel$ is the squared-loss risk minimizer within $\Phi$, then derive the
excess risk lower bound under \textbf{squared-loss risk}, with a note on the 0-1 risk bound
following as a remark.
For analytical tractability, the derivation below uses a squared-loss surrogate on a binary linear subproblem (e.g., one-vs-rest), with $y \in \{-1, +1\}$. Throughout, $w^\dagger$ denotes the unconstrained population minimizer of $\mathcal{R}_{\mathrm{sq}}(w) = \mathbb{E}_{(x,y)\sim\mathcal{P}}[(1 - y\langle w, \phi_v(x)\rangle)^2]$, decomposed as $w^\dagger = w^\dagger_\parallel + w^\dagger_\perp$ with $w^\dagger_\parallel \in \Phi$ and $w^\dagger_\perp \in \Phi^\perp$. This formalizes the same geometric obstruction as Proposition~2: constraining adaptation to $\Phi$ induces irreducible excess risk proportional to the missing mass in $\Phi^\perp$. The same intuition extends to multiclass settings via one-vs-rest or pairwise binary subproblems.

\paragraph{Step 1: Margin decomposition.}
For any $w \in \Phi$ and a point $(x, y)$ with representation $z = \phi_v(x)$, decompose
$w^\dagger$ as $w^\dagger = w^\dagger_\parallel + w^\dagger_\perp$ where $w^\dagger_\parallel \in \Phi$ and
$w^\dagger_\perp \in \Phi^\perp$. The margin under $w$ satisfies:
\[
    m_w(z) = y\langle w, z\rangle,
    \qquad
    m^\dagger(z) = y\langle w^\dagger_\parallel, z\rangle + y\langle w^\dagger_\perp, z\rangle.
\]
Since $w \in \Phi$ and $w^\dagger_\perp \in \Phi^\perp$, no $w \in \Phi$ can recover the term
$y\langle w^\dagger_\perp, z\rangle$.

\paragraph{Step 2: $w^\dagger_\parallel$ is optimal within $\Phi$.}
Let $\hat{w} = \arg\min_{w \in \Phi} \mathcal{R}_{\mathrm{sq}}(w)$ be the true in-subspace squared-loss minimizer, and let $w^\dagger_\parallel = \Pi_\Phi w^\dagger$ denote the orthogonal projection of $w^\dagger$ onto $\Phi$. For any $w \in \Phi$, expand the squared loss:

\[
    \mathcal{R}_{\mathrm{sq}}(w)
    = \mathbb{E}\bigl[(1 - y\langle w, z\rangle)^2\bigr].
\]
We decompose the residual:
\begin{equation}
\label{eq:decomposed_risk}
    1 - y\langle w, z\rangle
    = \underbrace{(1 - y\langle w^\dagger, z\rangle)}_{\text{irreducible}}
    + \; y\langle w^\dagger_\parallel - w,\, z\rangle
    + \; y\langle w^\dagger_\perp, z\rangle,
\end{equation}

Taking expectations and using the uncorrelation condition of Assumption~\ref{assum:subspace} ($\Pi_\Phi \Sigma w^\dagger_\perp = 0$), the cross-term between $y\langle w^\dagger_\parallel - w,\, z\rangle$ and $y\langle w^\dagger_\perp, z\rangle$ vanishes for all $w \in \Phi$, so the in-$\Phi$ minimizer is $\hat{w} = w^\dagger_\parallel$.


\paragraph{Step 3: Squared-loss excess risk lower bound.}

Since $w^\dagger$ is the unconstrained minimizer of $\mathcal{R}_{\mathrm{sq}}$, its first-order optimality condition gives $\mathbb{E}[(1 - y\langle w^\dagger, z\rangle)\,yz] = 0$, so the cross-term between the irreducible residual $(1 - y\langle w^\dagger, z\rangle)$ and $y\langle w^\dagger_\perp, z\rangle$ in~\eqref{eq:decomposed_risk} also vanishes. Therefore:
\[
    \mathcal{R}_{\mathrm{sq}}(\hat{w}) - \mathcal{R}_{\mathrm{sq}}(w^\dagger)
    = \mathbb{E}\bigl[(y\langle w^\dagger_\perp, z\rangle)^2\bigr]
    = \mathbb{E}\bigl[\langle w^\dagger_\perp, z\rangle^2\bigr].
\]
By Assumption~\ref{assum:subspace}, representations $z = \phi_v(x)$ satisfy
$\mathbb{E}[\langle w^\dagger_\perp/\|w^\dagger_\perp\|,\, z\rangle^2] \geq \sigma^2$, so:
\[
    \mathbb{E}\bigl[\langle w^\dagger_\perp, z\rangle^2\bigr]
    = \|w^\dagger_\perp\|^2 \cdot \mathbb{E}\!\left[\left\langle \frac{w^\dagger_\perp}{\|w^\dagger_\perp\|},\,
    z\right\rangle^{\!2}\right]
    \;\geq\; \sigma^2 \|w^\dagger_\perp\|^2.
\]
Normalizing by $\|w^\dagger\|^2$ and setting $c_{\sigma} = \sigma^2$:
\begin{equation}
\label{eq:excess_risk}
    \inf_{w \in \Phi}\mathcal{R}_{\mathrm{sq}}(w) - \mathcal{R}_{\mathrm{sq}}(w^\dagger)
    \;\geq\; c_{\sigma} \cdot \frac{\|w^\dagger_\perp\|^2}{\|w^\dagger\|^2}.
\end{equation}
Since $\mathcal{R}_{\mathrm{sq}}(\hat{w}) \geq \inf_{w\in\Phi}\mathcal{R}_{\mathrm{sq}}(w)$
by definition, we conclude:
\begin{equation}
\label{eq:excess_risk_omega}
    \mathcal{R}_{\mathrm{sq}}(\hat{w})
    \;\geq\; \mathcal{R}_{\mathrm{sq}}(w^\dagger)
    + \Omega\!\left(\frac{\|w^\dagger_\perp\|^2}{\|w^\dagger\|^2}\right),
\end{equation}
where $\Omega(\cdot)$ is used informally to indicate that the excess risk grows with
representational absence $\|w^\dagger_\perp\|/\|w^\dagger\|$; the concrete lower bound is the
constant-factor statement in~\eqref{eq:excess_risk}. \qed


\par\smallskip\noindent\textit{Remark (0--1 Risk Bound).} An analogous lower bound proportional to $\|w^\dagger_\perp\|/\|w^\dagger\|$ (linear rather than squared) can be obtained under standard margin and tail conditions on the data distribution; the squared-loss result established above is therefore the conservative claim.
\par\smallskip

\paragraph{Extension to fine-tuning (LoRA/SFT).}
Proposition~2 isolates the governing geometry in the frozen-encoder limit. Full SFT and LoRA relax this limit by perturbing $\phi_v$, and we show next that the same representational obstruction persists under that perturbation, with the experimental results of the main paper confirming the predicted failure.

When the encoder is partially updated through LoRA or full SFT, the effective subspace $\Phi$ is perturbed to $\Phi' = \Phi + \Delta\Phi$. With few-shot data ($n$ examples per class), the perturbation $\Delta\Phi$ is poorly constrained: the $n \times C$ training points span at most an $(nC)$-dimensional subspace of $\mathbb{R}^d$, which for typical few-shot settings ($n \leq 8$, $C \leq 20$) is vastly smaller than $d \geq 1024$ for modern VLMs. In the few-shot regime $nC \ll d$, the update signal for discovering useful out-of-subspace directions is severely limited, making adaptation unstable and prone to spurious correlations within the limited training set~\cite{zhang2020revisiting, geirhos2020shortcut}. This is consistent with the empirical observation in the main text that 1-shot SFT gains across the 15 MVTec categories range from $-27.5\%$ to $+36.9\%$: the few-shot data is insufficient to identify useful directions in $\Phi^\perp$, and the optimizer latches onto spurious correlations that vary unpredictably across categories.

\paragraph{Connection to sample complexity.}
For the in-subspace component $w^\dagger_\parallel$, standard PAC-Bayes bounds give sample complexity $\mathcal{O}(r / \epsilon^2)$ to achieve excess risk $\epsilon$ within $\Phi$, where $r = \dim(\Phi)$ is the effective dimension. The bound in~\eqref{eq:excess_risk} shows this convergence has a hard floor: in distant-OOD regimes where $\|w^\dagger_\perp\|^2 / \|w^\dagger\|^2$ is large, even perfect estimation of $w^\dagger_\parallel$ leaves irreducible excess risk of order $\|w^\dagger_\perp\|^2/\|w^\dagger\|^2$. Additional samples from $\mathcal{P}_{\mathrm{target}}$ cannot reduce this floor within the frozen-encoder (or weakly perturbed few-shot) regime, since it is determined entirely by the geometry of $\Phi$ relative to $w^\dagger$, not by estimation error within $\Phi$.

\subsection{Dual-Mode Complementarity Analysis}
\label{appendix:dual_mode}

This section provides the extended analysis supporting the dual-mode trait extraction design introduced in Sec.~3.5 of the main paper.

\paragraph{Setup.}
Let $\mathcal{T}^S$ and $\mathcal{T}^L$ denote the semantic and low-level trait sets extracted for a given dataset. Define the trait coverage function with respect to the ground-truth class distinctions $\mathcal{Y} = \{1, \ldots, C\}$ as follows.

\begin{definition}[Trait Coverage]
\label{def:coverage}
For a trait set $\mathcal{T}$ and a pair of classes $(c, c')$, define the pairwise discriminability as
\begin{equation}
    \mathrm{Disc}(\mathcal{T}, c, c') = \bigl|\{t \in \mathcal{T} : t \in \mathcal{T}_c \;\Delta\; \mathcal{T}_{c'}\}\bigr|,
\end{equation}
where $\Delta$ denotes symmetric difference, \ie, $\mathrm{Disc}$ counts traits that are present in one class but not the other. The coverage is the fraction of class pairs that are discriminable:
\begin{equation}
    \mathrm{Cov}(\mathcal{T}) = \frac{1}{\binom{C}{2}} \sum_{c < c'} \mathbf{1}\bigl[\mathrm{Disc}(\mathcal{T}, c, c') > 0\bigr].
\end{equation}
\end{definition}

\paragraph{Claim.}
Under the capacity asymmetry established in Proposition~1:
\begin{enumerate}[label=(\roman*),nosep,leftmargin=*]
    \item Semantic trait coverage $\mathrm{Cov}(\mathcal{T}^S)$ degrades with increasing feature-space divergence $\delta$, because the VLM's domain knowledge becomes increasingly unreliable and semantic prompting is more likely to trigger description collapse (producing generic or incorrect labels).
    \item Low-level trait coverage $\mathrm{Cov}(\mathcal{T}^L)$ remains approximately stable across divergence levels, because low-level prompting produces purely observational descriptions that depend on the domain-agnostic visual vocabulary.
    \item The combined coverage satisfies $\mathrm{Cov}(\mathcal{T}^S \cup \mathcal{T}^L) \geq \max\bigl(\mathrm{Cov}(\mathcal{T}^S),\, \mathrm{Cov}(\mathcal{T}^L)\bigr)$, with strict inequality whenever $\mathcal{T}^S$ and $\mathcal{T}^L$ contribute non-redundant discriminating traits for at least one class pair.
\end{enumerate}

\paragraph{Argument.}
Claim (iii) follows directly from the definition of coverage: if \\ $\mathrm{Disc}(\mathcal{T}^S, c, c') = 0$ but $\mathrm{Disc}(\mathcal{T}^L, c, c') > 0$ for some class pair $(c, c')$, then the union $\mathcal{T}^S \cup \mathcal{T}^L$ discriminates this pair while $\mathcal{T}^S$ alone does not, increasing the coverage.

Claims (i) and (ii) follow from the structure of Proposition~1. Semantic prompting asks the VLM to leverage its domain knowledge; when this knowledge is absent (high $\delta$), the generated traits are generic and fail to differentiate between target classes, reducing $\mathrm{Disc}(\mathcal{T}^S, c, c')$ toward zero across many pairs. Low-level prompting explicitly bypasses domain knowledge and requests primitive visual observations (colors, textures, shapes, spatial patterns), which are grounded in the surviving descriptive capacity $G(f, \mathcal{P}_{\mathrm{target}}) \geq G_0$. These observations remain informative because distinct classes typically differ in at least some low-level visual aspects, even when the VLM has no understanding of what these differences signify.

\paragraph{Empirical support.}
The dual-mode ablation in the main paper (Table~2) and the component level ablation in Supp.~\cref{tab:ablation:trait_prompt_variant} provide direct evidence: on distant-OOD datasets, removing either trait mode degrades accuracy relative to the full dual-mode system. The combined system consistently outperforms both individual modes, confirming that the two trait types contribute non-redundant discriminative information. On the Pok\'emon benchmark, neither single mode recovers full dual-mode accuracy, consistent with claim~(iii) that the union discriminates class pairs that no single mode separates alone.

\paragraph{Practical implication.}
The complementarity analysis justifies the dual-mode design as more than an engineering heuristic: it is a necessary consequence of the asymmetric capacity structure. A system relying solely on semantic traits would inherit the fragility of the VLM's domain knowledge, while a system relying solely on low-level traits would sacrifice the rich categorical information available in near-OOD domains. The dual-mode design provides graceful degradation: as semantic capacity diminishes across the OOD spectrum, low-level traits absorb an increasing share of the discriminative burden, ensuring robust coverage regardless of the divergence level.


\section{Characterizing Representational Absence: Detailed Analysis}
\label{appendix:select_distant_ood}

As established in Sec.~3.3 of the main paper, the distant-OOD regime is characterized by two structural properties of the frozen encoder's representations: high feature-space divergence (MK-MMD) from the pretraining distribution and/or low latent cluster--label alignment (NMI) with respect to the target task. This section provides the complete per-dataset measurements, validates the characterization against downstream adaptation outcomes, and examines its robustness across VLM architectures.


\subsection{Per-Dataset MK-MMD and NMI Values}

\Cref{tab:mkmmd_nmi_full} and \Cref{fig:ood_taxonomy_both} report MK-MMD and NMI for all 23 evaluated datasets, computed from the frozen Qwen2.5-VL vision encoder. ImageNet serves as the reference distribution for MK-MMD and is excluded from regime assignment; the 22 non-reference datasets are each assigned to near-OOD or distant-OOD. A dataset is assigned to the distant-OOD regime if it exhibits MK-MMD $> \tau_{\mathrm{MMD}}$ or NMI $< \tau_{\mathrm{NMI}}$. The 1-shot SFT adaptation gain $\Delta_{\mathrm{SFT}} = \mathrm{Acc}_{\text{1-shot SFT}} - \mathrm{Acc}_{\text{zero-shot}}$ is reported as independent empirical validation, not as a definitional criterion.

Thresholds are determined by a two-stage variant of Otsu's method that reflects the causal structure of the two axes. MK-MMD is thresholded first: it measures whether target images occupy regions of the representation space covered by the pretraining distribution, a question independent of class semantics. NMI is thresholded second, computed only on the near-side subset identified in Stage~1, where it is interpretable as semantic class alignment rather than a domain-shift artefact. Computing a global NMI threshold is problematic because domain-shifted datasets can exhibit arbitrarily low NMI due to feature collapse, conflating two structurally distinct failure modes; applying NMI-first is similarly ill-posed as low NMI may reflect either semantic misalignment or representation collapse. Two-stage Otsu with MK-MMD first separates these failure modes cleanly, yielding $\tau_{\mathrm{MMD}} = 0.49$ and $\tau_{\mathrm{NMI}} = 0.42$; the latter sits within a natural gap between Pok\'emon ($\mathrm{NMI}=0.196$) and the four FGVC benchmarks ($\mathrm{NMI}\geq 0.638$), with \cref{tab:threshold_sensitivity} confirming stability as $\tau_{\mathrm{NMI}}$ varies across $[0.30, 0.60]$.

\begin{table}[h!]
    \centering
    \caption{Representational absence metrics and adaptation outcomes across all evaluated datasets (Qwen2.5-VL-7B). Regime assignment is determined by MK-MMD and NMI thresholds; $\Delta_{\mathrm{SFT}}$ is reported for validation. Datasets are grouped by regime and sorted by MK-MMD within each group.}
    \label{tab:mkmmd_nmi_full}
    \small
    \begin{tabular}{l c c c c c}
    \toprule
    \textbf{Dataset}  & \textbf{MK-MMD} & \textbf{NMI} & \textbf{0-shot (\%)} & \textbf{$\Delta_{\mathrm{SFT}}$ (\%)} & \textbf{Regime} \\
    \midrule
    \multicolumn{6}{l}{\textit{Reference}} \\
    ImageNet         & ---   & 0.881 & --- & ---      & (reference) \\
    \midrule
    \multicolumn{6}{l}{\textit{Near-OOD (low MK-MMD, high NMI)}} \\
    Pets-37          & 0.174 & 0.651 & 80.9 & +6.3\textsuperscript{$\dagger$}  & Near-OOD \\
    Cars-196         & 0.289 & 0.750 & 56.2 & +27.3  & Near-OOD \\
    Flowers-102      & 0.310 & 0.975 & 72.4 & +23.9  & Near-OOD \\
    Aircraft         & 0.409 & 0.638 & 35.9 & +16.9  & Near-OOD \\
    \midrule
    \multicolumn{6}{l}{\textit{Distant-OOD: Semantic Gap (low MK-MMD, low NMI)}} \\
    Pok\'emon        & 0.411 & 0.196 & 47.47 & $-$1.95   & Distant-OOD \\
    \midrule
    \multicolumn{6}{l}{\textit{Distant-OOD: Domain-Shifted (high MK-MMD)}} \\
    MVTec Grid       & 0.566 & 0.215 & 36.1  & +4.2     & Distant-OOD \\
    Retinal OCT      & 0.578 & 0.564 & 14.36 & $+$0.41 & Distant-OOD \\
    MVTec Wood       & 0.587 & 0.486 & 60.3  & $-$1.4   & Distant-OOD \\
    WM-811K          & 0.596 & 0.282 & 13.13 & $-$0.72  & Distant-OOD \\
    MVTec Metal Nut  & 0.603 & 0.355 & 47.3  & $-$26.4  & Distant-OOD \\
    MVTec Hazelnut   & 0.610 & 0.400 & 61.9  & +15.2    & Distant-OOD \\
    MVTec Tile       & 0.615 & 0.875 & 58.6  & $-$9.0   & Distant-OOD \\
    MVTec Toothbrush & 0.620 & 0.005 & 55.0  & $-$27.5  & Distant-OOD \\
    MVTec Cable      & 0.626 & 0.465 & 12.8  & +28.3    & Distant-OOD \\
    MVTec Transistor & 0.648 & 0.283 & 18.9  & +36.9    & Distant-OOD \\
    MVTec Bottle     & 0.653 & 0.362 & 50.6  & $-$2.5   & Distant-OOD \\
    MVTec Capsule    & 0.663 & 0.197 & 46.0  & $-$17.4  & Distant-OOD \\
    MVTec Pill       & 0.664 & 0.378 & 15.1  & +5.7     & Distant-OOD \\
    MVTec Leather    & 0.673 & 0.371 & 44.9  & +15.3    & Distant-OOD \\
    MVTec Screw      & 0.676 & 0.063 & 26.0  & $-$7.2   & Distant-OOD \\
    MVTec Zipper     & 0.714 & 0.243 & 19.6  & $-$2.1   & Distant-OOD \\
    MVTec Carpet     & 0.715 & 0.521 & 53.2  & +1.8     & Distant-OOD \\
    \midrule
    \multicolumn{6}{l}{\textit{MVTec AD aggregate}} \\
    MVTec Mean       & ---   & ---   & 40.4  & +0.9   & Distant-OOD \\
    \bottomrule
    \multicolumn{6}{@{}p{0.95\columnwidth}@{}}{\footnotesize\textsuperscript{$\dagger$}The modest $\Delta_{\mathrm{SFT}}$ for Pets-37 reflects a ceiling effect: zero-shot accuracy is already $80.9\%$, leaving limited room for improvement. Its near-OOD status is confirmed by low MK-MMD ($0.174$) and high NMI ($0.651$).} \\
    \end{tabular}
\end{table}

Several observations merit discussion. First, the separation between regimes is stark at 1-shot. All near-OOD datasets exhibit positive $\Delta_{\mathrm{SFT}}$ (the smallest being Pets-37 at $+6.3\%$, attributable to a ceiling effect), while distant-OOD datasets exhibit near-zero mean gain with high variance ($\Delta_{\mathrm{SFT}}$ ranging from $-27.5\%$ to $+36.9\%$ across MVTec categories), indicating that gradient-based adaptation becomes \emph{unreliable} rather than uniformly ineffective. This volatility is consistent with Proposition~2 of the main paper: when the frozen encoder lacks task-relevant structure, single-example gradient updates are dominated by projection noise. Second, Retinal OCT is assigned to the distant-OOD regime by its high MK-MMD ($0.578$) despite exhibiting moderate NMI ($0.564$), demonstrating that MK-MMD detects the visual domain gap that NMI alone would miss. The downstream result $\Delta_{\mathrm{SFT}} = +0.41\%$ is consistent with the near-zero-mean pattern observed across distant-OOD datasets, providing independent empirical support for this assignment. Third, the MVTec categories span a wide range of NMI values ($0.005$--$0.875$) while uniformly exhibiting high MK-MMD ($> 0.56$), confirming that domain-shifted datasets are reliably identified by feature-space divergence regardless of cluster-label alignment. These three observations together confirm that the regime distinction is structurally meaningful and not an artifact of threshold choice.

Near-OOD datasets are included in \Cref{tab:mkmmd_nmi_full} as calibration references rather than primary evaluation targets of IVL. The theoretical motivation follows directly from Propositions~1 and~2. In the near-OOD regime, low MK-MMD and high NMI indicate that the frozen encoder preserves substantial discriminative capacity $D(f, \mathcal{P}_{\mathrm{target}})$, consistent with negligible representational absence and the gradient futility floor of Proposition~2 being near zero. Gradient-based adaptation can therefore effectively recover useful task structure within $\Phi$ without the structural barrier that motivates IVL. IVL is designed to operate when this condition fails, routing classification through the preserved descriptive capacity $G(f, \mathcal{P}_{\mathrm{target}}) \geq G_0$ to compensate for degraded discriminative capacity. When $D$ remains intact, this routing is not expected to provide a structural advantage over direct parametric adaptation. Evaluating IVL on near-OOD datasets would therefore conflate two structurally distinct regimes, as formalised in Remark~1 of the main paper.

\subsection{Cross-Model Robustness}

\begin{figure}[h!]
    \centering
    \includegraphics[width=\linewidth]{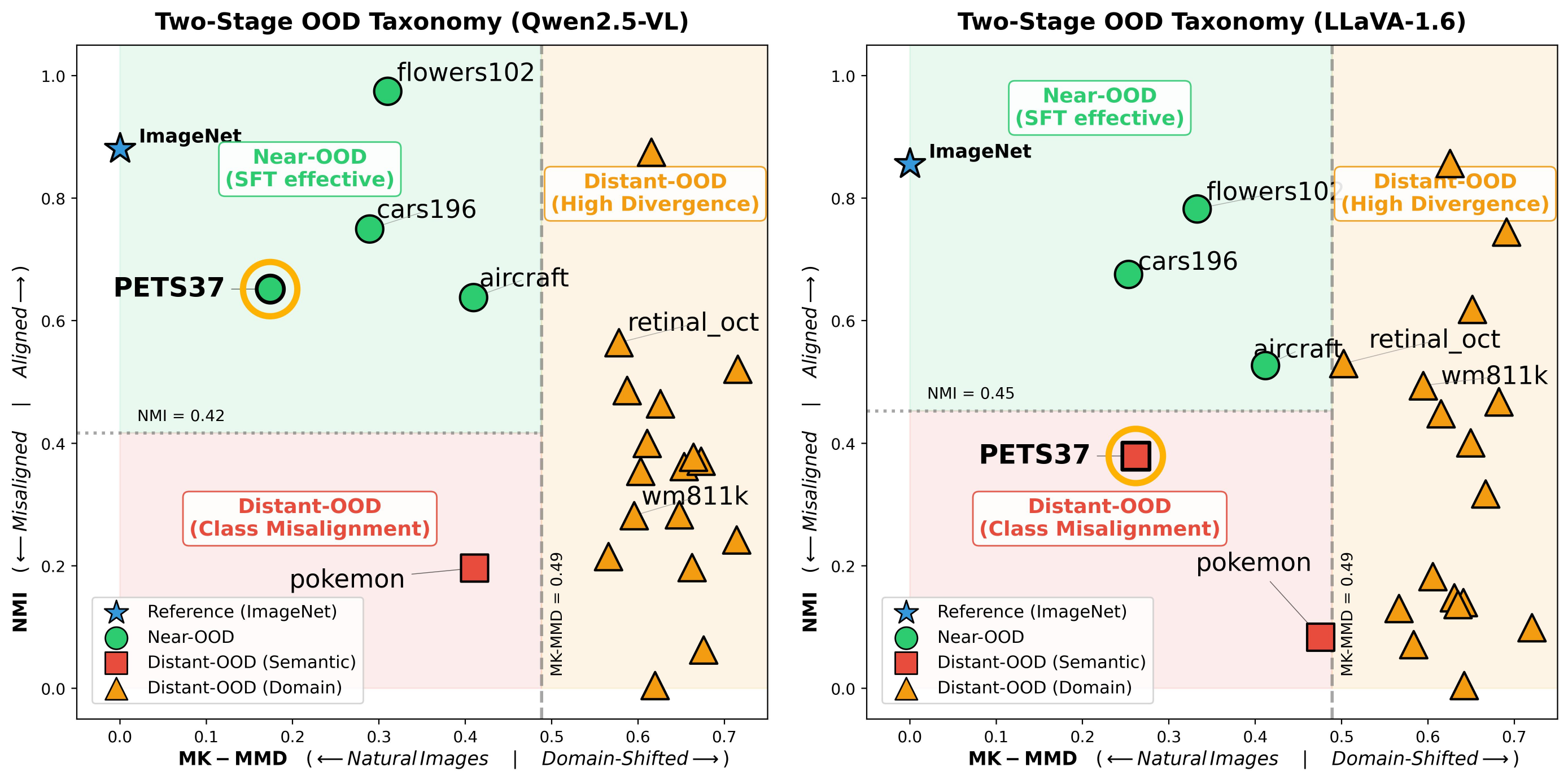}
    \caption{Two-stage OOD taxonomy under Qwen2.5-VL (\emph{left}) and LLaVA-1.6 (\emph{right}). Thresholds are determined independently by two-stage Otsu applied to each encoder's own metric distribution: $\tau_{\mathrm{MMD}} = 0.49$ for both; $\tau_{\mathrm{NMI}} = 0.42$ (Qwen2.5-VL) and $\tau_{\mathrm{NMI}} = 0.45$ (LLaVA-1.6).}
    \label{fig:ood_taxonomy_both}
\end{figure}

To verify that the regime structure is not an artifact of a specific encoder architecture, we replicate the analysis using the LLaVA-1.6-Mistral-7B vision encoder, applying the same two-stage Otsu procedure independently to LLaVA's own metric distribution. This yields $\tau_{\mathrm{MMD}} = 0.49$ and $\tau_{\mathrm{NMI}} = 0.45$ for LLaVA. As shown in \Cref{fig:ood_taxonomy_both}, both models yield structurally identical regime partitions with a single exception: Pets-37.

\paragraph{The Pets-37 case: a counter-intuitive prediction supported by experiment.}
Pets-37 is a standard fine-grained benchmark that every practitioner would consider near-OOD. Under Qwen2.5-VL, the MK-MMD/NMI framework agrees with this intuition (MK-MMD $= 0.174$, NMI $= 0.651$). Under LLaVA, however, the framework makes a counter-intuitive prediction: Pets-37 is reclassified as distant-OOD due to low NMI ($0.379 < \tau_{\mathrm{NMI}} = 0.45$). This prediction is derived purely from the frozen encoder's geometry, before any adaptation is attempted and independent of any downstream performance metric.

The downstream experiments in \Cref{tab:pets37_llava} confirm that this counter-intuitive assignment is supported. Under LLaVA-1.6, Pets-37 exhibits all three hallmarks of the distant-OOD regime identified in the main paper.

\begin{table}[h!]
    \centering
    \caption{Adaptation results on Pets-37 under LLaVA-1.6-Mistral-7B. The dataset is assigned to the distant-OOD regime by the MK-MMD/NMI framework under two-stage Otsu (NMI $= 0.379 < \tau_{\mathrm{NMI}} = 0.45$).}
    \label{tab:pets37_llava}
    \small
    \begin{tabular}{l cccc}
    \toprule
    & \textbf{Zero-Shot} & \textbf{SFT} & \textbf{CoOp} & \textbf{IVL (Ours)} \\
    \midrule
    1-shot & 18.26 & 28.14 & 2.93 & \textbf{31.34} \\
    8-shot & ---   & 29.13 & 5.63 & \textbf{33.18} \\
    \bottomrule
    \end{tabular}
\end{table}

\begin{enumerate}[nosep,leftmargin=*]
\item \textbf{Low zero-shot performance.} LLaVA-1.6 achieves only $18.26\%$ on Pets-37, compared to $80.9\%$ under Qwen2.5-VL. The vision encoder maps pet breed images to poorly structured regions of the representation space, exactly as the low NMI predicts.
\item \textbf{Modest SFT gains.} One-shot SFT reaches $28.14\%$, a $+9.9$\,pp improvement that falls far short of the $+18.6\%$ mean gain observed for near-OOD datasets (\Cref{tab:mkmmd_nmi_full}). At 8-shot, SFT gains plateau at $29.13\%$, indicating that additional data does not resolve the representational gap, consistent with the gradient futility predicted by Proposition~2.
\item \textbf{Prompt-tuning collapse.} CoOp collapses to $2.93\%$ at 1-shot and $5.63\%$ at 8-shot, well below the $18.26\%$ zero-shot baseline. This is consistent with the prompt-tuning failure mode observed across distant-OOD tasks in Table~1 of the main paper (Pok\'emon $6.94\%$, WM-811K $11.45\%$ under LLaVA), confirming that continuous prompts cannot recombine features that the encoder never learned~\cite{ma2023understanding}.
\end{enumerate}

IVL outperforms all baselines on this task ($31.34\%$ at 1-shot, $33.18\%$ at 8-shot), matching the expected behavior in the distant-OOD regime where trait-based reasoning provides an advantage over gradient-based adaptation. This case provides supporting evidence against the concern that the MK-MMD/NMI framework is a post-hoc construction fitted to preconceived dataset categories: a framework designed to confirm existing intuitions would never reclassify a standard fine-grained benchmark as distant-OOD. The fact that this reclassification is (i)~derived solely from the frozen encoder's geometry, (ii)~made before any adaptation experiment, and (iii)~subsequently validated by the downstream performance pattern confirms that the framework captures genuine structure in the encoder's representation space. The regime shift from near-OOD (Qwen-VL) to distant-OOD (LLaVA) for the same dataset further demonstrates the model-dependent nature of representational absence: out-of-distribution status is a property of the model-dataset pair, not of the dataset alone.

\subsection{Near-OOD Control Cases}
\label{appendix:near_ood_control}

Whereas \Cref{tab:pets37_llava} shows IVL winning once Pets-37 crosses into the distant-OOD regime under LLaVA, the complementary near-OOD case confirms the other side of the boundary stated in Remark~1 of the main paper. \Cref{tab:pets37_imageneta_near} reports two settings on which the frozen encoder already preserves the relevant semantics. Under Qwen2.5-VL, Pets-37 is near-OOD and SFT lifts accuracy from $80.9\%$ to $87.2\%$, while ImageNet-A behaves the same way ($79.4\%$ to $82.0\%$) and is near-OOD on both axes (MK-MMD 0.092, NMI 0.544). In both cases IVL neither helps nor meaningfully hurts ($79.8\%$ and $78.9\%$), which is the expected outcome when discriminative capacity is intact and trait routing adds processing without new information.

\begin{table}[h!]
\centering
\caption{Near-OOD control cases (1-shot). When the frozen encoder preserves the target semantics, SFT improves over zero-shot and IVL provides no structural advantage, consistent with Remark~1 of the main paper. ImageNet-A uses a random 20-class subset.}
\label{tab:pets37_imageneta_near}
\small
\begin{tabular}{@{}l l ccc@{}}
\toprule
\textbf{Dataset} & \textbf{Backbone} & \textbf{ZS} & \textbf{SFT} & \textbf{IVL} \\
\midrule
Pets-37    & Qwen2.5-VL & 80.9 & \textbf{87.2} & 79.8 \\
ImageNet-A & Qwen2.5-VL & 79.4 & \textbf{82.0} & 78.9 \\
\bottomrule
\end{tabular}
\end{table}

\subsection{Threshold Robustness}
\label{appendix:threshold_robustness}

The regime boundaries $\tau_{\mathrm{MMD}} = 0.49$ and $\tau_{\mathrm{NMI}} = 0.42$ are determined by two-stage Otsu applied to the Qwen-VL measurements. A robust boundary should sit within a natural gap in the data, so that small perturbations do not change any dataset's assignment. Because MK-MMD cleanly separates the domain-shifted datasets (all with MK-MMD $> 0.56$), the only contestable boundary is $\tau_{\mathrm{NMI}}$. \Cref{tab:threshold_sensitivity} confirms that sweeping $\tau_{\mathrm{NMI}}$ across $[0.30, 0.60]$, with $\tau_{\mathrm{MMD}}$ fixed at its Otsu value, leaves all 22 assignments unchanged, indicating that the near-OOD and distant-OOD clusters are well-separated with no borderline cases in the intervening range.

\begin{table}[h!]
    \centering
\caption{Threshold robustness (Qwen2.5-VL). With $\tau_{\mathrm{MMD}}$ fixed at its Otsu value $0.49$, the NMI threshold $\tau_{\mathrm{NMI}}$ is swept across a band bracketing its Otsu value $0.42$. All 22 datasets retain identical regime assignments, confirming a natural separation in the data.}
    \label{tab:threshold_sensitivity}
    \begin{tabular}{c c}
    \toprule
    \textbf{$\tau_{\mathrm{NMI}}$} & \textbf{Datasets unchanged} \\
    \midrule
    0.30 & 22/22 \\
    0.36 & 22/22 \\
    0.42 (Otsu) & 22/22 \\
    0.48 & 22/22 \\
    0.54 & 22/22 \\
    0.60 & 22/22 \\
    \bottomrule
    \end{tabular}
\end{table}

Implementation details: $k$-means uses $k{=}C$, $n_{\mathrm{init}}{=}10$, \texttt{random\_state=42}; MK-MMD employs 5 Gaussian kernels with bandwidths $\sigma_k = 2^k \cdot \tilde{d}$ for $k \in \{-2,-1,0,1,2\}$, where $\tilde{d}$ is the median pairwise distance between the two feature sets, following the median heuristic of Long~\etal~\cite{long2015learning}.

\subsection{Reference Distribution Robustness}
\label{appendix:reference_distribution_robustness}
\begin{figure}[tp]
    \centering
    \includegraphics[width=\linewidth]{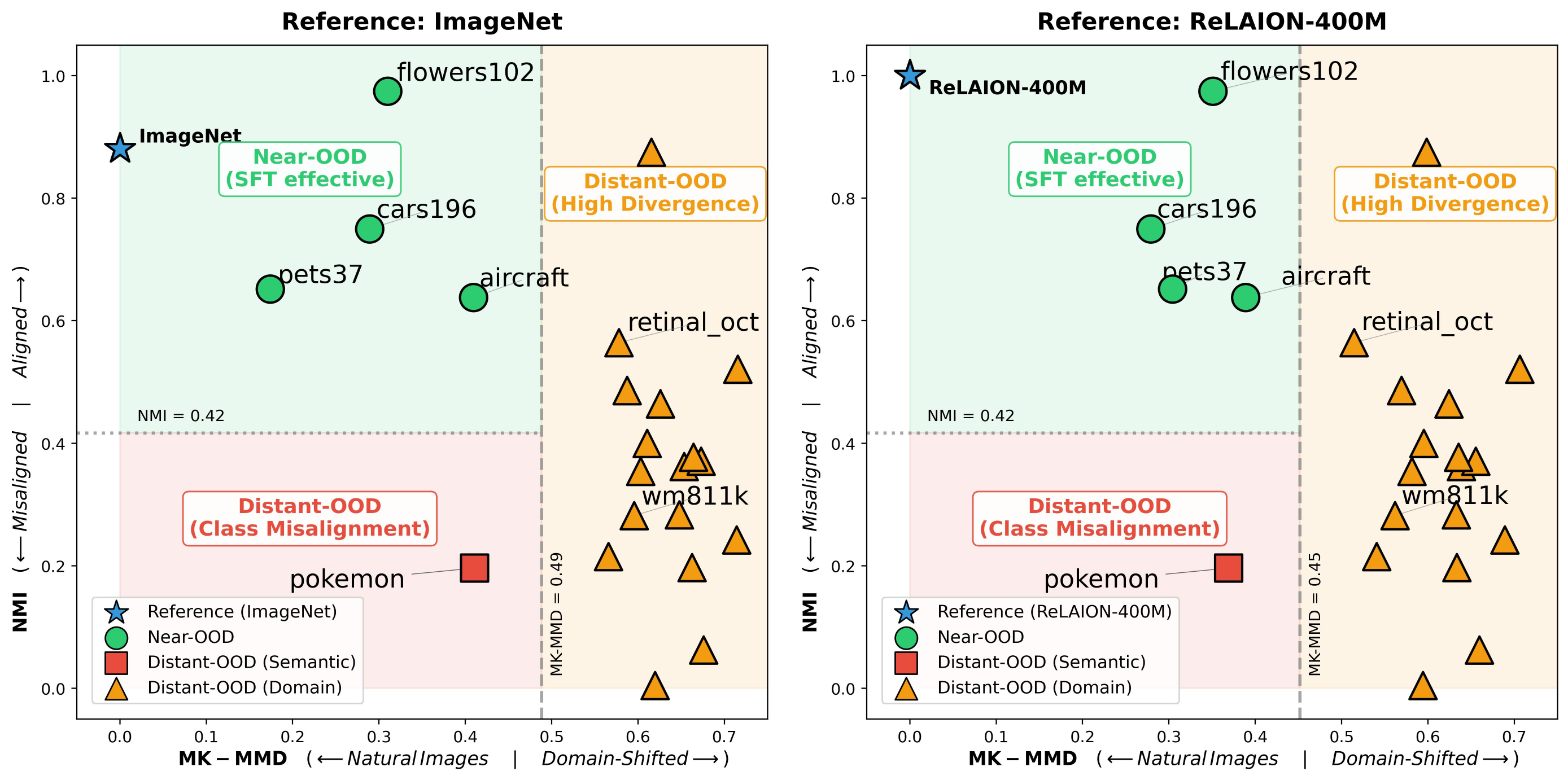}
    \caption{Two-stage OOD taxonomy under Qwen2.5-VL with ImageNet-1K (\emph{left}) and ReLAION-400M (\emph{right}) as the MK-MMD reference. Because NMI never involves the reference, each dataset keeps its vertical position across panels and moves only horizontally. No dataset crosses a regime boundary. Only $\tau_{\mathrm{MMD}}$ shifts ($0.49 \to 0.45$); $\tau_{\mathrm{NMI}} = 0.42$ is identical under both references.}
    \label{fig:ood_taxonomy_reference}
\end{figure}

The diagnostic is robust to the choice of MK-MMD reference distribution. Since NMI is reference-free, only the MK-MMD axis can change, and the wide MK-MMD gap between the near-side and domain-shifted datasets means moderate shifts in the reference move datasets horizontally without altering the Stage-1 partition. We confirm this by recomputing the Qwen2.5-VL diagnostic against ReLAION-400M \cite{schuhmann2021laion} (10k samples). Two-stage Otsu shifts $\tau_{\mathrm{MMD}}$ from $0.49$ to $0.45$ and leaves $\tau_{\mathrm{NMI}}$ exactly at $0.42$, because the unchanged Stage-1 partition feeds Stage-2 the same NMI distribution. As \Cref{fig:ood_taxonomy_reference} shows, every dataset shifts only horizontally and none crosses a regime boundary, so all 22 assignments are identical to \Cref{tab:mkmmd_nmi_full}.

\subsection{Additional Benchmarks}
To broaden medical coverage beyond Retinal OCT, we evaluate APTOS \cite{aptos2019-blindness-detection}, a diabetic-retinopathy grading benchmark, under Qwen2.5-VL at 1-shot. Its high MK-MMD (0.616) and low NMI (0.200) place it well within the distant-OOD regime, so the assignment follows the diagnostic directly. The downstream results reproduce the pattern of the other medical benchmarks. Zero-shot reaches $33.8\%$, one-shot SFT degrades to $15.8\%$, and IVL recovers to $49.0\%$. This SFT-collapse-then-IVL-recovery behavior mirrors Retinal OCT and WM-811K, confirming that the distant-OOD behavior documented for medical imaging is not specific to a single dataset.


\section{Trait Extraction Statistics Across the OOD Spectrum}
\label{sec:supp:descriptive_capacity}

Tables~\ref{tab:descriptive_capacity} and~\ref{tab:descriptive_capacity_mvtec_per_category} report two complementary characterizations of the extracted trait vocabulary across the full OOD spectrum: descriptive volume (Traits/Image) and cross-class informativeness (MeanNormIDF), defined as
\begin{equation}
    \mathrm{MeanNormIDF} = \frac{1}{C}\sum_{c=1}^{C}\frac{1}{|\mathcal{T}_c|}\sum_{t \in \mathcal{T}_c} \frac{\log\bigl(C\,/\,\mathrm{df}(t)\bigr)}{\log C},
    \label{eq:meannormidf}
\end{equation}
where $\mathrm{df}(t)$ is the number of classes containing canonical trait $t$ after synonym resolution and a score of 1.0 indicates a fully class-specific vocabulary. The $\log C$ normalization ensures comparability across datasets with different class 
counts; raw IDF scores would inflate for datasets with more classes, conflating 
vocabulary richness with class count. Across all 22 datasets (7~core $+$ 15~MVTec categories), the VLM extracts 11.2--27.1 traits per image and maintains MeanNormIDF in the range 0.733--0.972, with neither metric exhibiting systematic decline as MK-MMD increases; WM-811K and Retinal~OCT yield the highest trait counts (27.1 and 18.8 respectively) and retain high MeanNormIDF (0.733 and 0.810).

Both metrics are consistent with the lower bound $G \geq G_0 > 0$ established in Proposition~1(ii) of the main paper: descriptive volume is preserved regardless of domain shift, and the trait vocabulary retains class-discriminative structure across the full OOD spectrum. The comparatively lower MeanNormIDF on WM-811K and Retinal OCT reflects genuine inter-class visual ambiguity in these domains, where discriminative primitives recur across multiple pathology classes and thereby increase trait document frequency, rather than a collapse of descriptive quality. Functional evidence that the extracted traits carry genuine discriminative signal is provided by the dual-mode ablation (Table~2 of the main paper): low-level extraction alone achieves non-trivial classification accuracy across all benchmarks, confirming that trait quality is preserved alongside volume.

\begin{table}[tp!]
\centering
\caption{%
Trait extraction statistics across the OOD spectrum (core datasets; MVTec is reported per-category in Table~\ref{tab:descriptive_capacity_mvtec_per_category}).
Traits/Image and MeanNormIDF show no systematic decline with MK-MMD, confirming that descriptive volume and class-informativeness are preserved even when discriminative capacity collapses.%
}
\label{tab:descriptive_capacity}
\small
\resizebox{\linewidth}{!}{%
\begin{tabular}{lcccccc}
\toprule
\textbf{Dataset} & \textbf{Category} & \textbf{MK-MMD} & \textbf{\#Cls} & \textbf{Traits/Image} & \textbf{Avg.\ Len} & \textbf{MeanNormIDF} \\
\midrule
Oxford Pets   & Near-OOD    & 0.174 &  37 & $12.5 \pm 3.5$ & 2.2 & 0.890 \\
Stanford Cars & Near-OOD    & 0.289 & 196 & $16.7 \pm 4.9$ & 2.2 & 0.897 \\
Flowers-102   & Near-OOD    & 0.310 & 102 & $14.6 \pm 3.7$ & 2.3 & 0.919 \\
FGVC Aircraft & Near-OOD    & 0.409 & 100 & $15.6 \pm 4.2$ & 2.3 & 0.904 \\
\midrule
Pok\'{e}mon   & Distant-OOD & 0.411 &  18 & $14.4 \pm 4.3$ & 2.2 & 0.959 \\
Retinal OCT   & Distant-OOD & 0.578 &   8 & $18.8 \pm 5.4$ & 2.5 & 0.810 \\
WM-811K       & Distant-OOD & 0.596 &   9 & $27.1 \pm 5.2$ & 2.2 & 0.733 \\
\bottomrule
\end{tabular}
}
\end{table}

\begin{table}[tp!]
\centering
\caption{
Per-category trait extraction statistics on MVTec~AD (15 independent 1-shot datasets), ordered by MK-MMD.
MeanNormIDF remains consistently high (0.874--0.972) across all categories despite large domain gap (MK-MMD $\geq 0.566$).%
}
\label{tab:descriptive_capacity_mvtec_per_category}
\small
\begin{tabular}{lccccc}
\toprule
\textbf{MVTec Category} & \textbf{MK-MMD} & \textbf{\#Cls} & \textbf{Traits/Image} & \textbf{Avg.\ Len} & \textbf{MeanNormIDF} \\
\midrule
Grid        & 0.566 & 6 & $15.3 \pm 3.5$ & 2.1 & 0.972 \\
Wood        & 0.587 & 6 & $15.8 \pm 3.6$ & 2.4 & 0.912 \\
Metal Nut   & 0.603 & 5 & $17.2 \pm 2.9$ & 2.2 & 0.910 \\
Hazelnut    & 0.610 & 5 & $14.2 \pm 2.9$ & 2.1 & 0.927 \\
Tile        & 0.615 & 6 & $15.0 \pm 3.1$ & 2.2 & 0.932 \\
Toothbrush  & 0.620 & 2 & $15.0 \pm 2.0$ & 2.3 & 0.889 \\
Cable       & 0.626 & 9 & $16.0 \pm 3.7$ & 2.3 & 0.916 \\
Transistor  & 0.648 & 5 & $19.2 \pm 3.8$ & 2.2 & 0.902 \\
Bottle      & 0.653 & 4 & $13.2 \pm 1.3$ & 2.2 & 0.944 \\
Capsule     & 0.663 & 6 & $12.3 \pm 3.4$ & 2.0 & 0.916 \\
Pill        & 0.664 & 8 & $11.2 \pm 2.0$ & 2.2 & 0.819 \\
Leather     & 0.673 & 6 & $15.3 \pm 3.4$ & 2.1 & 0.896 \\
Screw       & 0.676 & 6 & $12.0 \pm 3.4$ & 2.0 & 0.906 \\
Zipper      & 0.714 & 8 & $15.2 \pm 3.6$ & 2.2 & 0.874 \\
Carpet      & 0.715 & 6 & $15.0 \pm 4.4$ & 2.0 & 0.902 \\
\midrule
Mean        &       &   & $14.8 \pm 2.0$ & 2.1 & 0.908 \\
\bottomrule
\end{tabular}
\end{table}

\section{Trait Grounding and Class Alignment}
\label{sec:supp:trait_alignment}

To quantify whether IVL's classification decisions are genuinely grounded in trait evidence, we analyze the alignment between cited traits and ground-truth classes on two structurally different benchmarks, Pok\'emon and MVTec~AD. For each prediction, we parse the VLM's trait-matching reasoning automatically. A strictly rule-based procedure extracts the ``Trait Matching'' section of the structured output via regex, identifies cited class labels using case-insensitive substring matching, and classifies the citation context (e.g., support vs.\ reject) using deterministic keyword lists (see Supp.~\S\ref{appendix:reproducibility}). This parsing operates entirely on the structured output fields produced by IVL's inference prompt, requires no manual annotation, and ensures no secondary LLM bias is introduced. All results use Qwen2.5-VL at 1-shot; MVTec numbers are averaged over 3 independent runs with different support-set samples.

Two complementary metrics capture the interpretability of IVL's decisions:

\begin{itemize}[nosep,leftmargin=*]
\item \textbf{Correct $\to$ trait-driven:} Among correct predictions, the fraction whose cited traits belong to the ground-truth class. This measures whether correct predictions arise from genuine trait evidence rather than coincidence.
\item \textbf{Incorrect $\to$ wrong-class traits:} Among incorrect predictions, the fraction whose cited traits belong to the (wrong) predicted class. This measures whether errors are \emph{structured} (attributable to specific trait confusion between visually similar classes) or random.
\end{itemize}

\noindent Together, these metrics characterize how fully IVL's decisions can be traced to explicit visual evidence, whether right or wrong. \Cref{tab:trait_alignment_summary} reports Pok\'{e}mon and all 15 MVTec categories in a unified view.

\begin{table}[tp!]
\centering
\caption{Unified trait-class alignment results (Qwen2.5-VL, 1-shot). The random baseline $1/C$ is the expected correct~$\to$~trait-driven rate under class-independent trait citation. MVTec numbers are averaged over 3 support-set runs.}
\vspace{-1.0em}
\label{tab:trait_alignment_summary}
\small
\begin{tabular}{lccccr}
\toprule
\textbf{Dataset/Category} & $C$ & \textbf{Random}
  & \textbf{Corr.$\to$trait} & \textbf{Incorr.$\to$wrong}
  & $\times$\textbf{chance} \\
\midrule
\multicolumn{6}{l}{\textit{Pok\'emon}} \\
Pok\'emon (881 images) & 18 & 5.6\% & 81.2\% & 72.3\% & 14.6$\times$ \\
\midrule
\multicolumn{6}{l}{\textit{MVTec AD (per-category)}} \\
Transistor  & 5 & 20.0\% & 90.9\% & 86.3\% & 4.5$\times$ \\
Tile        & 6 & 16.7\% & 76.5\% & 71.1\% & 4.6$\times$ \\
Wood        & 6 & 16.7\% & 76.5\% & 68.2\% & 4.6$\times$ \\
Carpet      & 6 & 16.7\% & 72.1\% & 53.3\% & 4.3$\times$ \\
Zipper      & 8 & 12.5\% & 70.9\% & 79.7\% & 5.7$\times$ \\
Grid        & 6 & 16.7\% & 65.5\% & 52.3\% & 3.9$\times$ \\
Leather     & 6 & 16.7\% & 64.7\% & 47.8\% & 3.9$\times$ \\
Metal Nut   & 5 & 20.0\% & 61.8\% & 59.7\% & 3.1$\times$ \\
Bottle      & 4 & 25.0\% & 53.9\% & 41.8\% & 2.2$\times$ \\
Capsule     & 6 & 16.7\% & 53.7\% & 32.0\% & 3.2$\times$ \\
Pill        & 8 & 12.5\% & 46.0\% & 31.7\% & 3.7$\times$ \\
Hazelnut    & 5 & 20.0\% & 42.3\% & 61.2\% & 2.1$\times$ \\
Cable       & 9 & 11.1\% & 39.1\% & 29.0\% & 3.5$\times$ \\
Screw       & 6 & 16.7\% & 30.9\% & 39.7\% & 1.9$\times$ \\
Toothbrush  & 2 & 50.0\% & 20.0\% & 19.2\% & 0.4$\times$ \\
\bottomrule
\end{tabular}

\end{table}

On Pok\'emon, 81.2\% of correct predictions are explainable through trait evidence (67.8\% directly supported by ground-truth class traits, 13.4\% correct via elimination of competing classes). Among incorrect predictions, 72.3\% systematically matched traits from the predicted (wrong) class, and 48.4\% of errors occurred when the ground-truth class was absent from the filtered trait dictionary, making correct classification impossible regardless of reasoning quality. On MVTec, correct~$\to$~trait-driven alignment varies by category (30.9\%--90.9\%, excluding Toothbrush), which is expected because some defect categories are visually distinctive while others require finer-grained discrimination among subtly different anomalies. Crucially, the relevant comparison is not between benchmarks but between each dataset/category and its random baseline.

\paragraph{Per-category analysis with random baseline.}
Since the number of classes $C$ varies across MVTec categories (from 2 to 9), the random baseline for the correct~$\to$~trait-driven metric (the expected rate if trait citations were independent of class identity) is $1/C$ per category. \Cref{tab:trait_alignment_summary} reports these per-category results directly.

All categories except Toothbrush exceed the random baseline by $1.9$--$5.7\times$, confirming that trait citations are systematically class-relevant rather than arbitrary. Several patterns merit discussion.

The correct~$\to$~trait-driven rate correlates with how visually distinctive each category's defect signatures are. Categories with spatially salient defects (Transistor: 90.9\%, Tile: 76.5\%, Wood: 76.5\%) achieve near-Pok\'emon-level alignment, while categories requiring subtle distinctions among fine-grained defect types (Screw: 30.9\%, Cable: 39.1\%) show weaker, but still well-above-chance, alignment.

The incorrect~$\to$~wrong-class rate reveals the structure of IVL's errors. Categories with high wrong-class rates (Transistor: 86.3\%, Zipper: 79.7\%, Tile: 71.1\%) indicate that errors stem from genuine trait overlap between visually similar defect types: the model found real visual evidence, but it pointed to the wrong class. Categories with lower wrong-class rates (Toothbrush: 19.2\%, Cable: 29.0\%, Pill: 31.7\%) reflect cases where the trait database lacks sufficient discriminative coverage. This distinction carries practical value: high wrong-class errors are addressable by adding contrastive traits that differentiate confusable classes, while low wrong-class errors signal a need for richer trait extraction. Both failure modes are diagnosable from IVL's explicit reasoning trace, an advantage that opaque parametric methods do not provide.

Toothbrush is a known outlier: with only $C{=}2$ classes and the lowest NMI in the entire evaluation (0.005), it represents the extreme vocabulary collapse regime where the encoder provides essentially no task-relevant structure. Its below-chance alignment (0.4$\times$) is consistent with the failure mode analysis in Supplementary~\S\ref{appendix:failure_mode}.

\subsection{Per-Class Trait Signatures}
\label{sec:supp:trait_signatures}
IVL traits are shared primitives rather than class prototypes, so classes are represented by combinations of traits rather than a single centroid.
Per-class signatures (the top-$k$ most frequent canonical traits per class) are therefore more informative than a global t-SNE projection of trait embeddings.

\Cref{tab:pokemon_class_signatures} lists the top-3 canonical traits for a representative subset of Pok\'emon types.
15/18 class signatures are human-recognizable from these traits alone, confirming that the trait dictionary captures semantically meaningful class-level distinctions.

\begin{table}[h!]
\centering
\caption{Representative per-class top-3 trait signatures on the Pok\'emon benchmark.}
\label{tab:pokemon_class_signatures}
\small
\begin{tabular}{ll}
\toprule
\textbf{Type} & \textbf{Top-3 canonical traits} \\
\midrule
Fire   & orange body, red fur, flame-tipped tail \\
Steel  & metallic limbs, metallic surface, gray body \\
Ice    & ice crystal structure, blue accents, ice cube head \\
Bug    & antennae-like appendages, spiky appearance, black legs \\
\bottomrule
\end{tabular}
\end{table}

The same pattern holds for industrial anomaly detection.
\Cref{tab:hazelnut_class_signatures} lists the top-3 canonical traits for all five MVTec-Hazelnut classes: 5/5 signatures are human-recognizable, and for every defect class the anomaly name (\emph{crack}, \emph{cut}, \emph{hole}, \emph{print}) appears verbatim among the top-3 traits, while the nominal class is anchored by \emph{clean surface}.

\begin{table}[h!]
\centering
\caption{Per-class top-3 trait signatures on the MVTec-Hazelnut benchmark (semantic-mode IVL, 5-shot). The defect-name token appears verbatim in the top-3 traits for every anomaly class.}
\label{tab:hazelnut_class_signatures}
\small
\begin{tabular}{ll}
\toprule
\textbf{Class} & \textbf{Top-3 canonical traits} \\
\midrule
crack & darker brown crack, visible separation, cracked shell \\
cut   & sharp cut edges, irregular cut edge, visible separation lines \\
hole  & smooth contrasting texture, nut hole detail, darker brown hole \\
print & text-like white spots, white shell contrast, white spot circle \\
none  & clean surface, smooth ridged surface, consistent shell structure \\
\bottomrule
\end{tabular}
\end{table}

\subsection{Per-Trait Grounding Visualization and Comparison to SFT}
\label{appendix:trait_grounding}
We visualize how individual low-level traits ground spatially in a query and how this grounding differs from a fine-tuned baseline, using a defect-free MVTec Hazelnut sample. The Hazelnut category illustrates how IVL's structured CoT reasoning recovers from cases where attention localization alone is non-discriminative.

On this normal sample, SFT's \textit{none} representation collapses after 1-shot fine-tuning, with the \textit{crack}/\textit{none} activation ratio reaching $108\times$ and \textit{none} falling to the noise floor. The cause is that ``no defect'' is not a groundable visual primitive, so few-shot fine-tuning leaves no representation for \textit{none} to bind to, and SFT's own reasoning trace reveals the failure by describing ``a natural split or crack.'' IVL's localization alone is also non-discriminative here. Its per-class CoT reasoning, however, recognizes the same region as ``the cap of the nut,'' systematically rejects each defect trait, and predicts \textit{none}.

\begin{figure}[h!]
    \centering
    \includegraphics[width=0.92\linewidth]{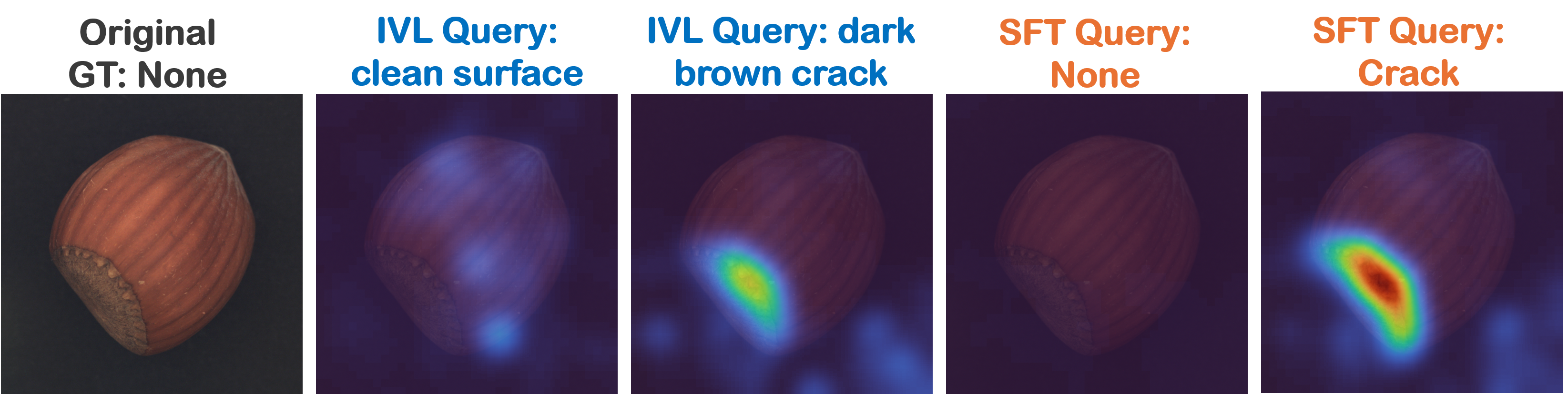}
    \caption{Per-trait grounding on a normal MVTec Hazelnut sample. Left to right: original image; IVL query attention for ``clean surface'' (none trait); IVL query attention for ``brown crack'' (crack trait); SFT attention for none; SFT attention for crack.}
    \label{fig:hazelnut_grounding_contrast}
\end{figure}

\section{Ablation Studies}
\label{appendix:ablation}
In this section, we conduct ablation studies on the Pokemon dataset to assess the contribution of individual components in our proposed pipeline. Due to the high computational cost of full-scale experiments, all ablations are performed on a balanced 144-sample subset of the test set. Accordingly, the accuracy values reported in this section are specific to this subset and are not directly comparable to the results in Table~1 of the main paper, which is evaluated on the full image test set. Overall, the ablations show that most components contribute substantially to performance, while the default configuration is selected for full-pipeline robustness and overall consistency rather than for maximizing accuracy on this reduced subset alone. As shown in the following tables, simplifying or removing individual modules often leads to non-trivial changes in accuracy, indicating that the effectiveness of the method arises from the interaction of multiple design choices rather than from any single component in isolation.

\paragraph{Trait Extraction.} 
We validate the necessity of our dual-mode extraction strategy in \cref{tab:ablation:trait_prompt_variant}. Relying solely on semantic prompting results in a 6.2\% accuracy drop, confirming our observation in Section~3.5 of the main paper that semantic queries in novel domains often trigger \textit{description collapse}, where models produce generic labels rather than discriminative features. Conversely, using only low-level features leads to an 11.1\% decline, as primitive descriptions lack the structural coherence provided by semantic knowledge. The hybrid approach effectively bridges this gap.

\begin{table}[htbp]
    \centering
    \caption{Ablation: Trait Prompt Variants}
    \vspace{-1.0em}
    \label{tab:ablation:trait_prompt_variant}
    \setlength{\tabcolsep}{6pt}
    \renewcommand{\arraystretch}{0.9}
        \begin{tabular}{lcc}
            \toprule
            \textbf{Configuration}  & \textbf{Accuracy (\%)} & \textbf{$\Delta$ Accuracy (\%)} \\
            \midrule
            \textbf{Hybrid (Default)} & 38.9 & \multicolumn{1}{c}{---} \\
            Semantic  & 32.6 & \textcolor{red}{$-6.2$} \\
            Low-Level & 27.8 & \textcolor{red}{$-11.1$} \\
            \bottomrule
        \end{tabular}
        \vspace{-2.0em}
\end{table}

\paragraph{Filtering.}
\Cref{tab:ablation:filtering_analysis} reports the impact of our hierarchical 
filtering stages. The \textit{Global Only} configuration achieves a marginal $+0.7\%$ gain 
on this subset, confirming that global embeddings carry sufficient discriminative signal for 
the object-centric Pok\'emon images. We nonetheless retain the two-stage default because 
the local refinement step produces the patch-level attention maps consumed by the visual 
grounding stage: removing grounding entirely costs $-4.2\%$ (\Cref{tab:ablation:visual_grounding}), 
making the net contribution of local refinement $+3.5$ pp once its downstream effect is 
accounted for. Skipping local refinement would therefore force grounding to rely on 
global attention, risking predictions anchored to spurious background correlations rather 
than localized evidence. Removing both stages (\textit{No Filtering}) costs $-2.1\%$, and 
over-filtering ($k_2{=}50$) costs $-5.6\%$, confirming that the filtering budget itself matters.

\begin{table}[htbp]
    \centering
    \setlength{\tabcolsep}{6pt}
    \renewcommand{\arraystretch}{0.9}
    \caption{Ablation: Filtering Strategy}
    \label{tab:ablation:filtering_analysis}
    \begin{tabular}{lcc}
        \toprule
        \textbf{Configuration} & \textbf{Accuracy (\%)} & \textbf{$\Delta$ Accuracy (\%)} \\
        \midrule
        \textbf{Two-Stage (Default)} & 38.9 & \multicolumn{1}{c}{---} \\
        Global Only & 39.6 & \textcolor{teal}{$+0.7$} \\
        Local Only & 37.5 & \textcolor{red}{$-1.4$} \\
        No Filtering & 36.8 & \textcolor{red}{$-2.1$} \\
        Strict Filtering ($k_2=50$) & 33.3 & \textcolor{red}{$-5.6$} \\
        \bottomrule
    \end{tabular}
\end{table}

\paragraph{Clustering.}
\Cref{tab:ablation:clustering_method} assesses the impact of
clustering strategy. K-means achieves a $+2.1\%$ advantage on this balanced 144-sample
subset, where the fixed cluster count $k{=}C$ aligns well with the uniform class
distribution. We retain HDBSCAN as the default because it does not require a pre-specified
cluster count and groups traits based on local density, preserving rare but discriminative
traits that would otherwise be absorbed into larger K-means centroids. This property is
critical on datasets with unequal trait density across classes, which is the typical case
in real-world few-shot scenarios beyond the controlled ablation subset.

\begin{table}[htbp]
    \centering
    \setlength{\tabcolsep}{6pt}
    \renewcommand{\arraystretch}{0.9}
    \caption{Ablation: Clustering Method}
    \label{tab:ablation:clustering_method}
    \begin{tabular}{lcc}
        \toprule
        \textbf{Configuration}  & \textbf{Accuracy (\%)} & \textbf{$\Delta$ Accuracy (\%)} \\
        \midrule
        \textbf{HDBSCAN (Default)} & 38.9 & \multicolumn{1}{c}{---} \\
        K-means & 41.0 & \textcolor{teal}{$+2.1$} \\
        Similarity-based & 32.6 & \textcolor{red}{$-6.2$} \\
        \bottomrule
    \end{tabular}
\end{table}


\paragraph{Embedding Model.}
We compare embedding architectures in \cref{tab:ablation:embedding_method}. Despite the Qwen VLM encoder having higher dimensionality (3584 vs. 768), the specialized Sentence Transformer achieves superior performance. This empirical result supports our hypothesis that VLM encoders, optimized for vision-text alignment, tend to over-compress the textual semantic space. Sentence Transformers maintain the necessary granularity to distinguish between subtle trait variations, which is critical for accurate trait clustering. This ablation concerns the embedding used for trait clustering only. The trait embeddings used for global retrieval at inference ($\mathbf{E}_{\mathcal{T}}$, Sec.~3.6 of the main paper) are always produced by the VLM text encoder and therefore share the $d$-dimensional space of the pooled vision tokens in the $\mathbf{s}^g$ computation.

\begin{table}[htbp]
    \centering
    \setlength{\tabcolsep}{6pt}
    \renewcommand{\arraystretch}{0.9}

    \caption{Ablation: Clustering Embedding Method}
    \label{tab:ablation:embedding_method}
    \resizebox{\linewidth}{!}{%
\begin{tabular}{llccc}
        \toprule
        \textbf{Configuration}  & \textbf{Dimension} & \textbf{Accuracy (\%)} & \textbf{$\Delta$ Accuracy (\%)} \\
        \midrule
        \textbf{Sentence Transformer (Default)}  & 768 & 38.9 & \multicolumn{1}{c}{---} \\
        Qwen2.5-VL Encoder  & 3584 & 37.5 & \textcolor{red}{$-1.4$} \\
        \bottomrule
    \end{tabular}
}
\end{table}

\paragraph{Salience Filtering.}
We evaluate the importance of logical quality control in \cref{tab:ablation:salience_filtering}. As described in Section~3.5 of the main paper (Clustering, canonicalization, and refinement.), this step employs VLM common-sense reasoning to filter out semantically inconsistent traits (e.g., removing `aquatic' features from terrestrial classes). Removing this filtering step results in a 2.6\% drop in accuracy, demonstrating that without this relevance check, the trait database accumulates noise that degrades the precision of downstream matching.

\begin{table}[htbp]
    \centering
    \setlength{\tabcolsep}{6pt}
    \renewcommand{\arraystretch}{0.9}

    \caption{Ablation: Salience Filtering}
    \label{tab:ablation:salience_filtering}
    \resizebox{\linewidth}{!}{%
\begin{tabular}{lccc}
        \toprule
        \textbf{Configuration} & \textbf{Salience} & \textbf{Accuracy (\%)} & \textbf{$\Delta$ Accuracy (\%)} \\
        \midrule
        \textbf{Salience Filtering (Default)}  & On & 38.9 & \multicolumn{1}{c}{---} \\
        w/o Salience Filtering  & Off & 36.3 & \textcolor{red}{$-2.6$} \\
        \bottomrule
    \end{tabular}
}
\end{table}
\paragraph{Localized Visual Grounding.}
Finally, \cref{tab:ablation:visual_grounding} highlights the critical role of providing explicit visual evidence during classification. We disable the final step of feeding the extracted image crops to the VLM (as detailed in Section~3.6 of the main paper). This omission results in a 4.2\% decrease in accuracy, indicating that simply identifying traits is insufficient; the model relies on the direct observation of these localized regions to validate its predictions in distant-OOD scenarios.

\begin{table}[htbp]
    \centering
    \setlength{\tabcolsep}{6pt}
    \renewcommand{\arraystretch}{0.9}
    \caption{Ablation: Localized Visual Grounding}
    \label{tab:ablation:visual_grounding}
    \resizebox{\linewidth}{!}{%
\begin{tabular}{lccc}
        \toprule
        \textbf{Configuration} & \textbf{Visual Grounding} & \textbf{Accuracy (\%)} & \textbf{$\Delta$ Accuracy (\%)} \\
        \midrule
        \textbf{With Grounding (Default)} & On & 38.9 & \multicolumn{1}{c}{---} \\
        w/o Grounding & Off & 34.7 & \textcolor{red}{$-4.2$} \\
        \bottomrule
    \end{tabular}
}
\end{table}


\section{Dataset}
\label{appendix:reproducibility:dataset}
\subsection{Public Datasets}

Regime assignment for all benchmarks follows the MK-MMD/NMI characterisation
in \Cref{appendix:select_distant_ood}; per-dataset measurements and regime labels are reported in \cref{tab:mkmmd_nmi_full}. The primary evaluation targets of IVL are the distant-OOD benchmarks: Retinal OCT, WM-811k, and MVTec AD. For MVTec AD, we treat each of the 15 object categories as an independent recognition dataset with its own defect label space, consistent with the per-category evaluation protocol used throughout the paper. For classification benchmarks, we construct 1-shot and 8-shot support sets by sampling 1 and 8 images per class respectively. For each MVTec AD category, one image per defect type forms the single one-shot support set.

\subsection{Pokémon Dataset}
\begin{figure}[h]
    \centering
    \includegraphics[width=0.75\linewidth]{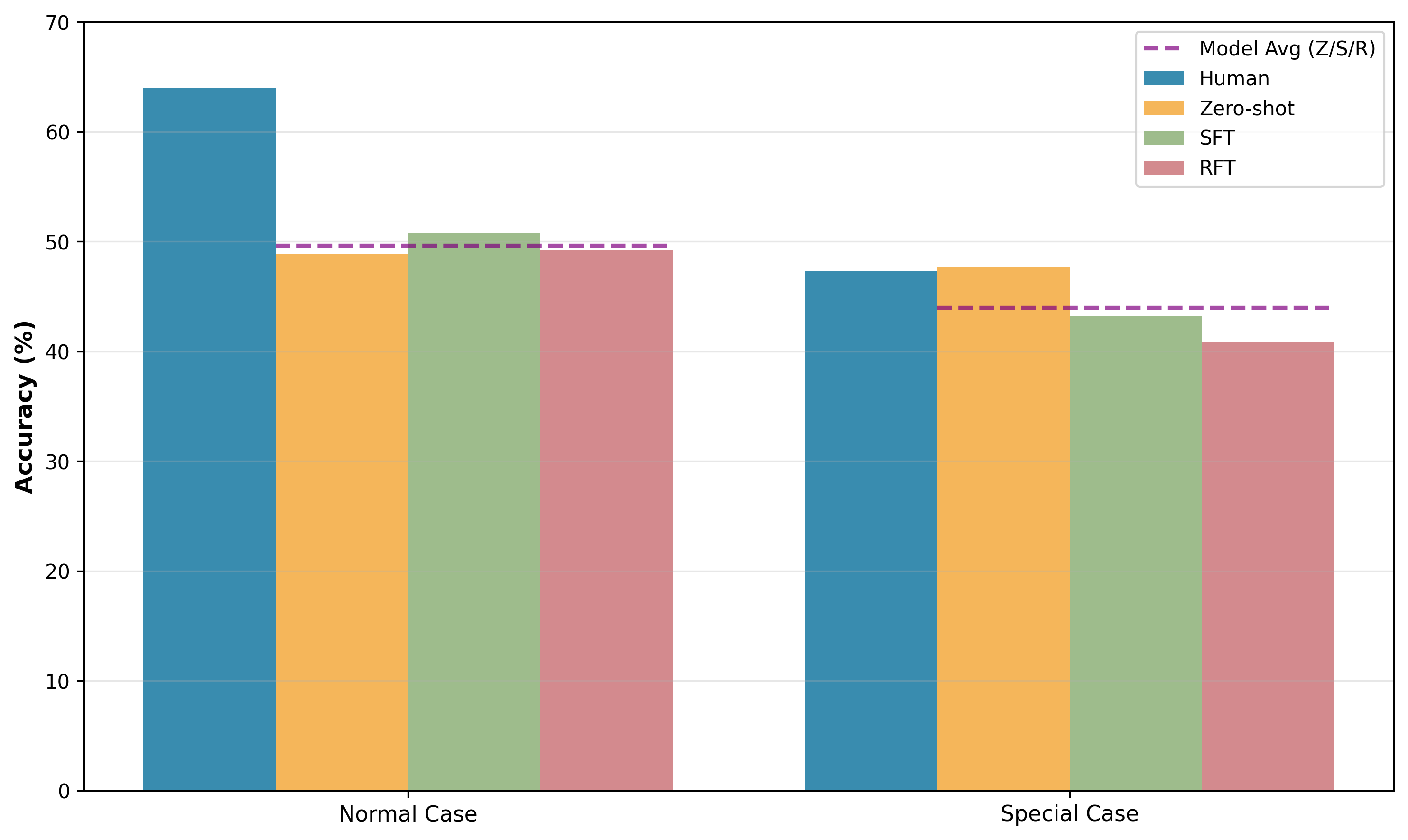}
    \caption{Human and Model Performance on Normal vs. Special Case Pokémon}
    \label{fig:human_and_model}
\end{figure}
In search of a more controlled distant-OOD benchmark, we curated a private evaluation set of 1,025 Pokémon species from the Pokémon Database~\cite{pokemondb, nintendopokemon}, selected for high visual quality and comprehensive coverage. The image files are not redistributed with this work. \footnote{Pok\'emon and the names of Pok\'emon species and types are trademarks of Nintendo, Creatures Inc., and GAME FREAK inc. These names are used solely to describe a research benchmark, and this work is not affiliated with, sponsored by, or endorsed by these companies.} The primary objective of this benchmark is to evaluate the model's ability to classify the elemental type of each Pokémon. Approximately half of the Pokémon in this dataset possess dual-type characteristics. For evaluation , our criteria required an exact match for single-type Pokémon and the correct identification of either one or both types for dual-type Pokémon to be marked as correct. The support set consists exclusively of single-type Pokémon, ensuring that all models learn from unambiguous class-specific supervision without being confounded by secondary type associations during support-set construction.

\paragraph{Dataset splits.}
Of the 1,025 total Pokémon species, 144 are reserved as the support pool (8 per type $\times$ 18 types). For 8-shot experiments, all 144 support images are used; for 1-shot, one image per type is sampled from these 144. The remaining 881 images form the full test set used in all main-paper results and the alignment analysis (Supp.~\S\ref{sec:supp:trait_alignment}). Ablation studies (Supp.~\S\ref{appendix:ablation}) use a balanced 144-sample subset drawn from the 881-image test set to reduce computational cost; all accuracy numbers reported in Table~1 of the main paper and in \Cref{tab:mkmmd_nmi_full} use the full 881-image test set.

To study the models' performance compared to human ability, we established human performance baselines by designing a questionnaire and categorizing participants into three expertise levels, which are analogous to different stages of model training:

\begin{itemize}
    \item \textbf{Beginners}, defined as those familiar only with iconic Pokémon like Pikachu, correspond to the performance of our base models (e.g., 7B Qwen2.5-VL).
\end{itemize}

\begin{itemize}
    \item \textbf{Intermediates}, who possess implicit knowledge of type associations (e.g., color schemes, morphological patterns), represent the human equivalent of few-shot \\ trained models.
\end{itemize}

\begin{itemize}
    \item \textbf{Experts}, identified by their extensive familiarity from completing Pokémon games, are analogous to fully trained models.
\end{itemize}

The dataset was evenly distributed across three questionnaire versions, each containing 341--342 Pokémons. Each version featured a similar distribution of well-known Pokémons species to ensure a balanced evaluation for all participant groups. Over a one-week collection period, we gathered 47 total responses (16 beginners, 16 intermediates, and 15 experts), evaluated with the same criteria as the models. The accuracies for beginners, intermediates and experts are 56.1\%, 61.9\% and 85.2\%, respectively.

Analyzing the accuracies between human and models, these results suggest that humans are able to identify patterns not captured by the models, enabling them to achieve higher accuracy. As shown in \cref{fig:human_and_model}, we broke down the Pokémon dataset into normal case and special case. The special case comprises Ultra Beast Pokémons and Paradox Pokémons, which are intentionally designed to decouple the logical connection between a Pokémon's appearance and its type. The figure reveals that when evaluating the special cases, human performance drops by nearly 20\%, falling to a level comparable to the models. In contrast, the models' performance remains approximately stable across both cases. This suggests that humans learned visual rules applicable to normal cases which failed to generalize to the special cases. The models, conversely, failed to learn these underlying patterns, even after fine-tuning, leading to their consistent, mediocre performance across both scenarios. This phenomenon consolidates our discovery that tasks trivial for humans through brief exposure remain computationally intensive for VLMs despite extensive parameter adaptation.

\section{Failure Mode}
\label{appendix:failure_mode}

\subsection{MVTec AD Transistor Category: Vocabulary Collapse}
\begin{figure}[t!]
    \centering
    \includegraphics[width=0.7\linewidth]{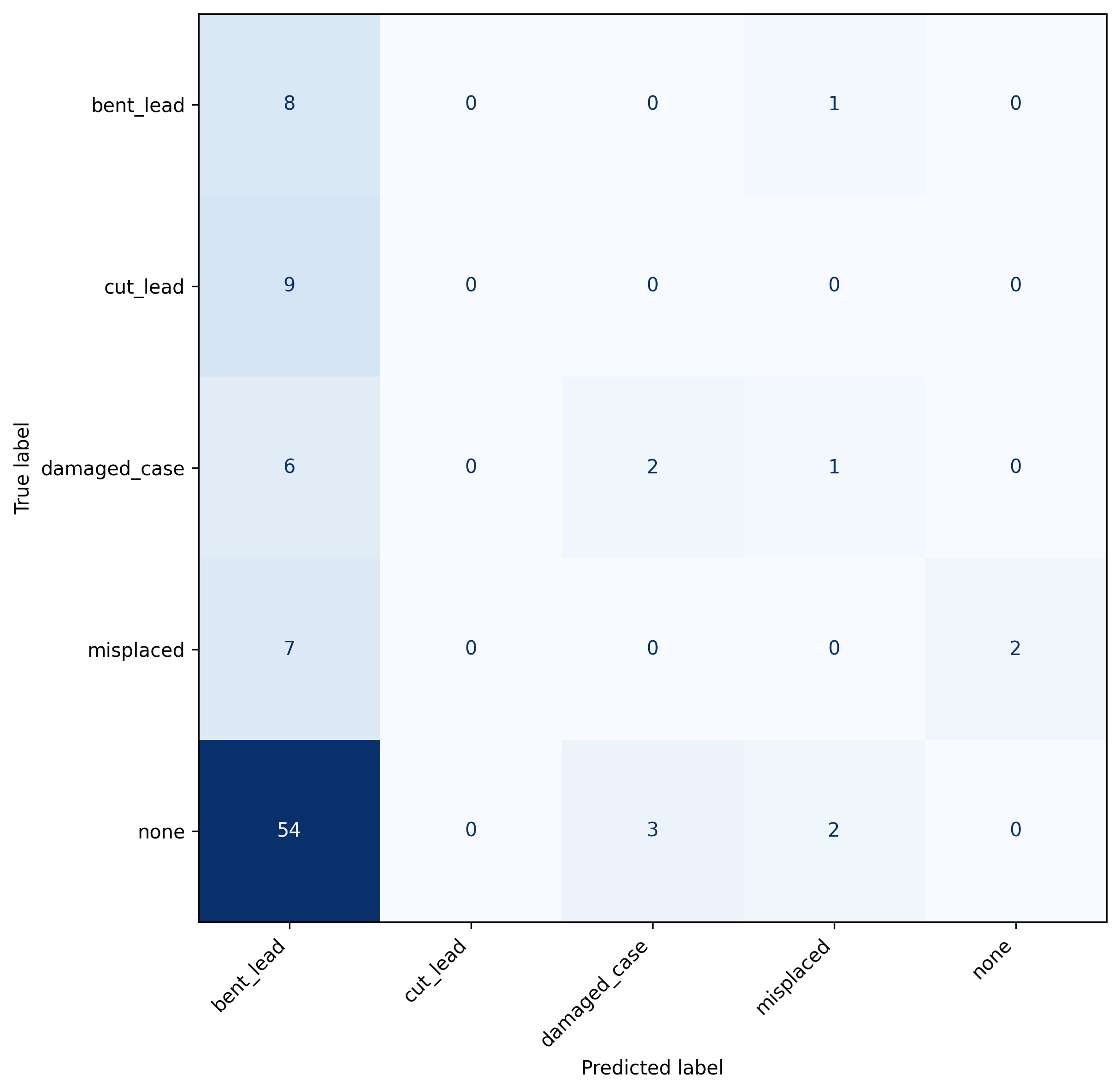}
    \caption{
    {
    Confusion matrix of IVL on the MVTec AD Transistor dataset. 
    Nearly all defect types, including \textit{none}, \textit{cut\_lead}, \textit{damaged\_case}, and \textit{misplaced}, are misclassified as \textit{bent\_lead}. 
    This collapse reflects a systematic failure mode. IVL interprets the presence of curved metal leads as definitive evidence for the \textit{bent\_lead} defect, 
    even though normal samples and several other defect types naturally exhibit curved leads.
    }}
    \label{fig:mvtec_confusion_matrix}
\end{figure}

We identified a systematic failure mode of IVL in the MVTec AD Transistor category. As shown in \Cref{fig:mvtec_confusion_matrix}, IVL predicts \textit{bent\_lead} for samples from multiple ground-truth categories, including \textit{none}, \textit{cut\_lead}, \textit{damaged\_case}, and \textit{misplaced}. This behavior corresponds to the vocabulary collapse failure mode discussed in the main paper, where the IVL trait dictionary reduces to a single dominant defect concept across multiple ground-truth categories.

\begin{figure}[tp!]
    \centering
    \includegraphics[width=0.15\linewidth]{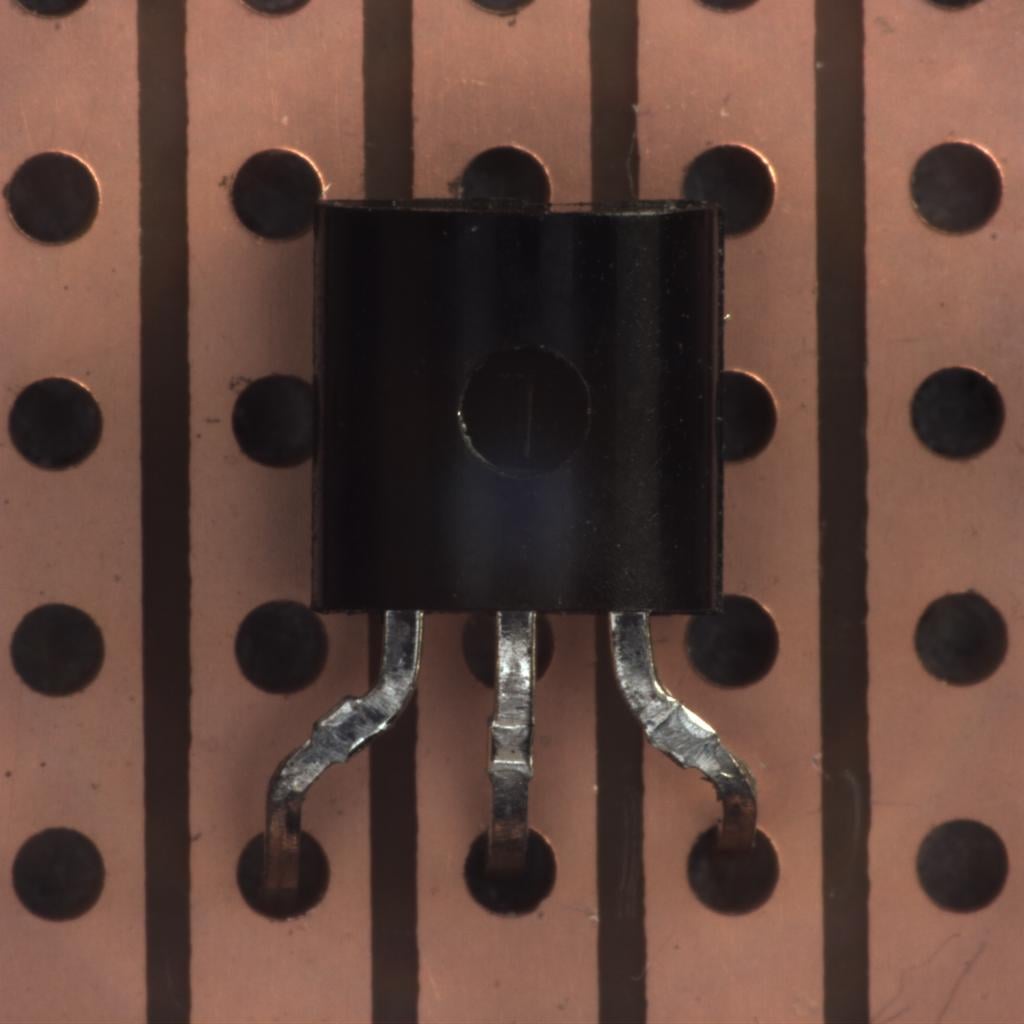}
    \includegraphics[width=0.15\linewidth]{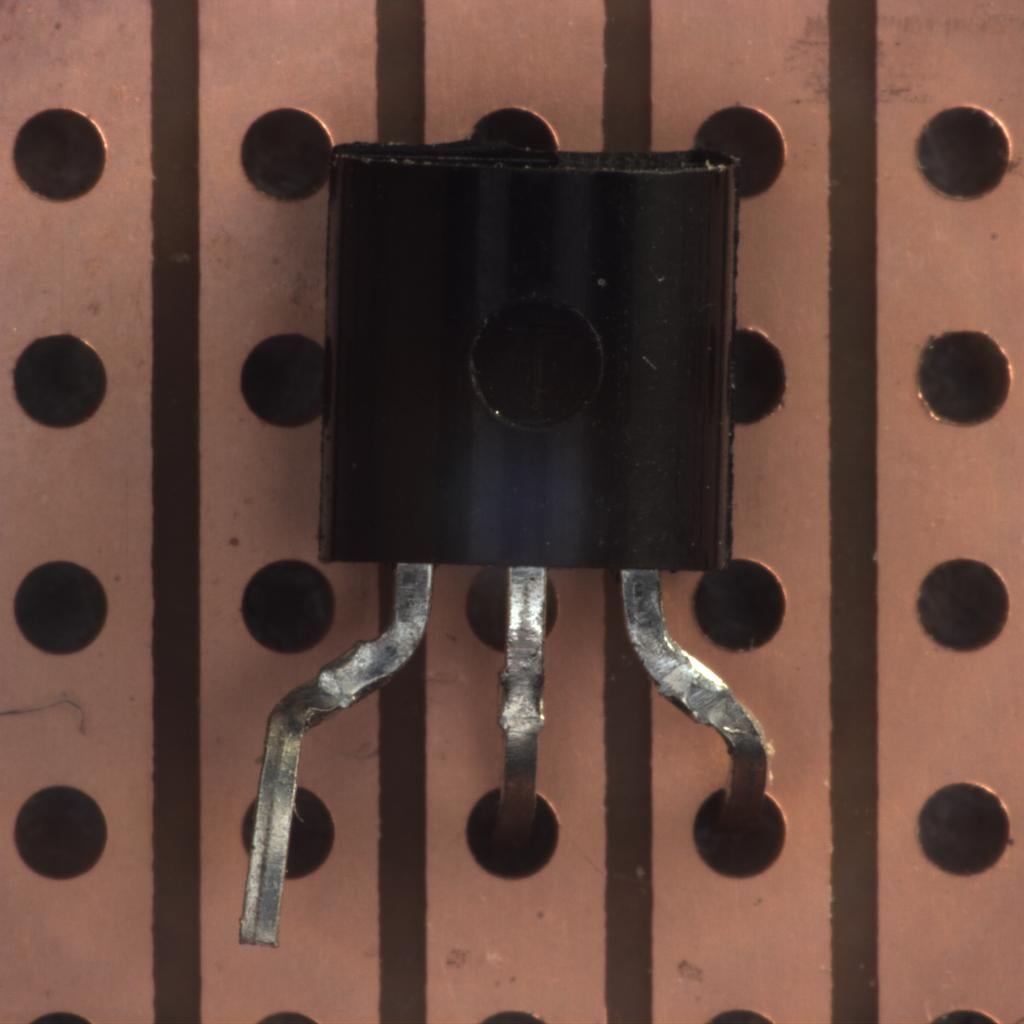}
    \includegraphics[width=0.15\linewidth]{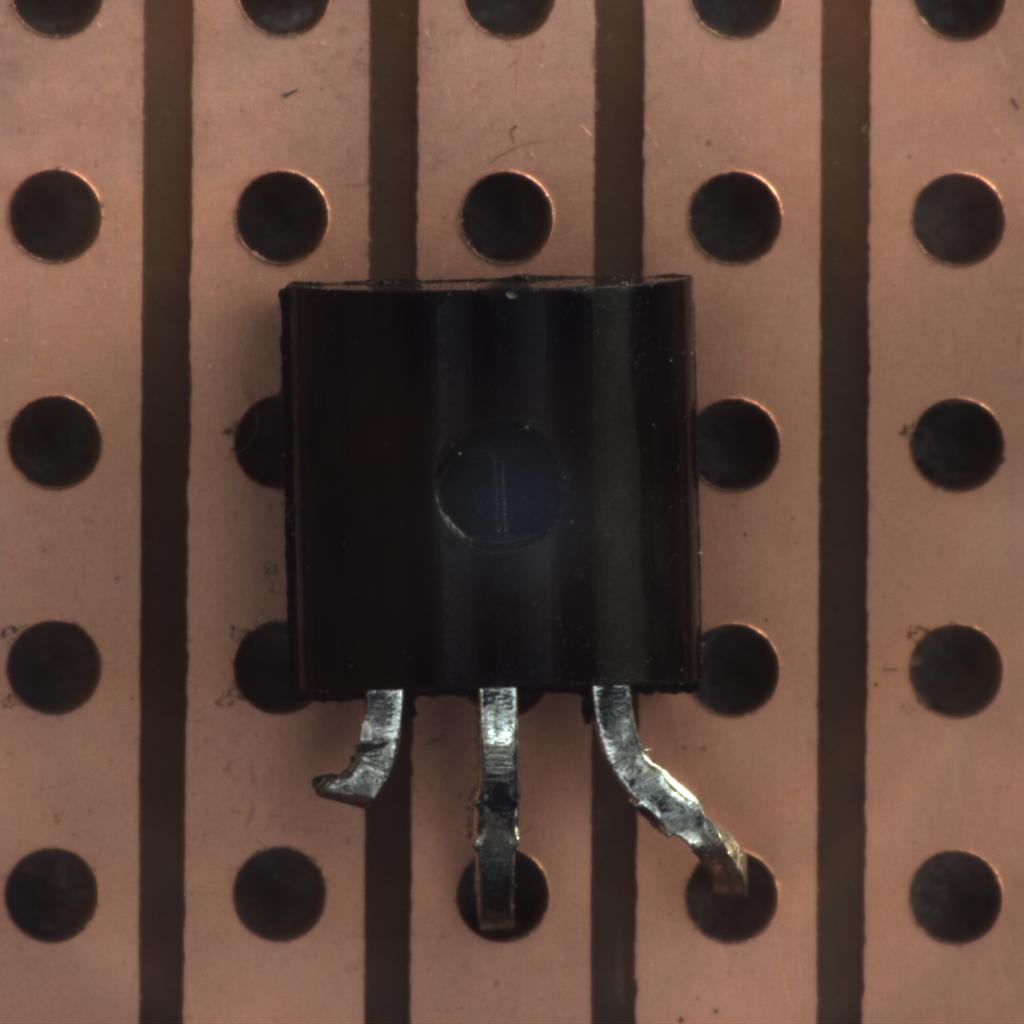}
    \includegraphics[width=0.15\linewidth]{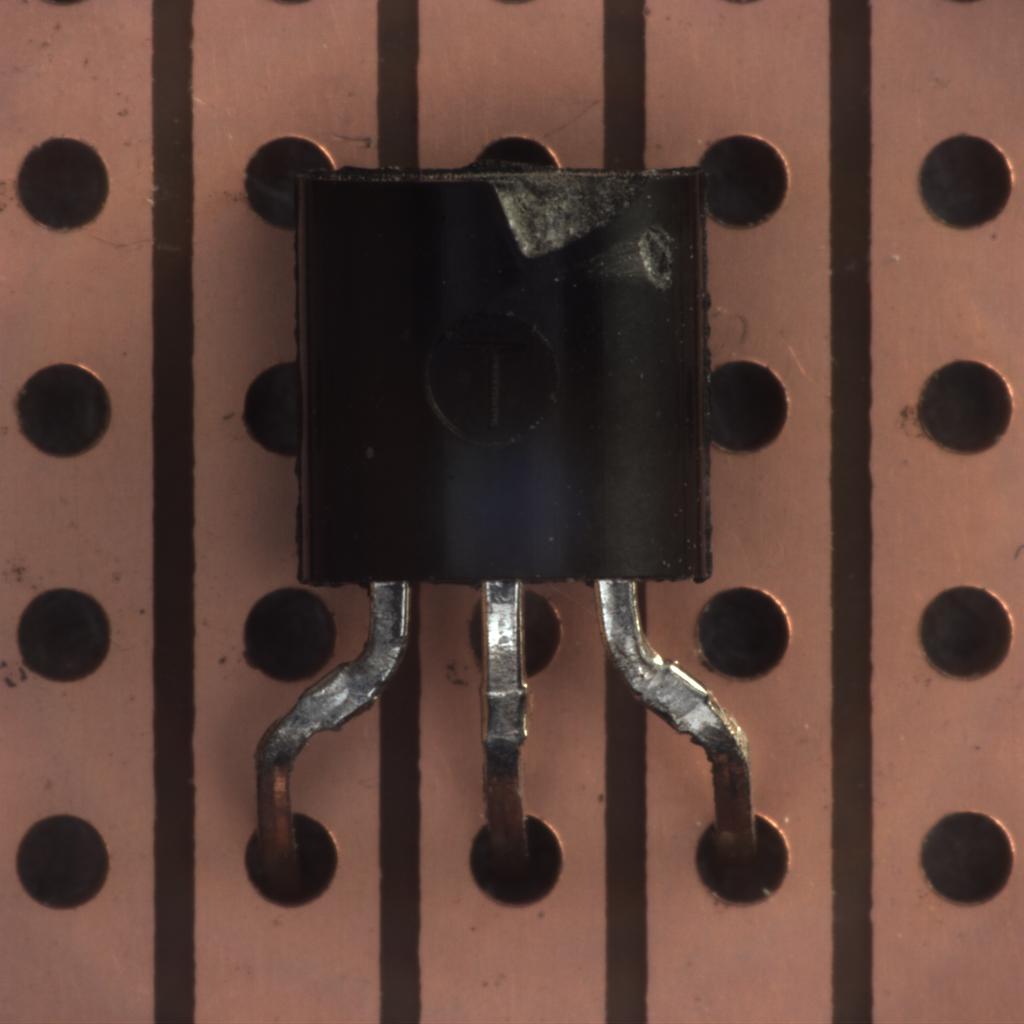}
    \includegraphics[width=0.15\linewidth]{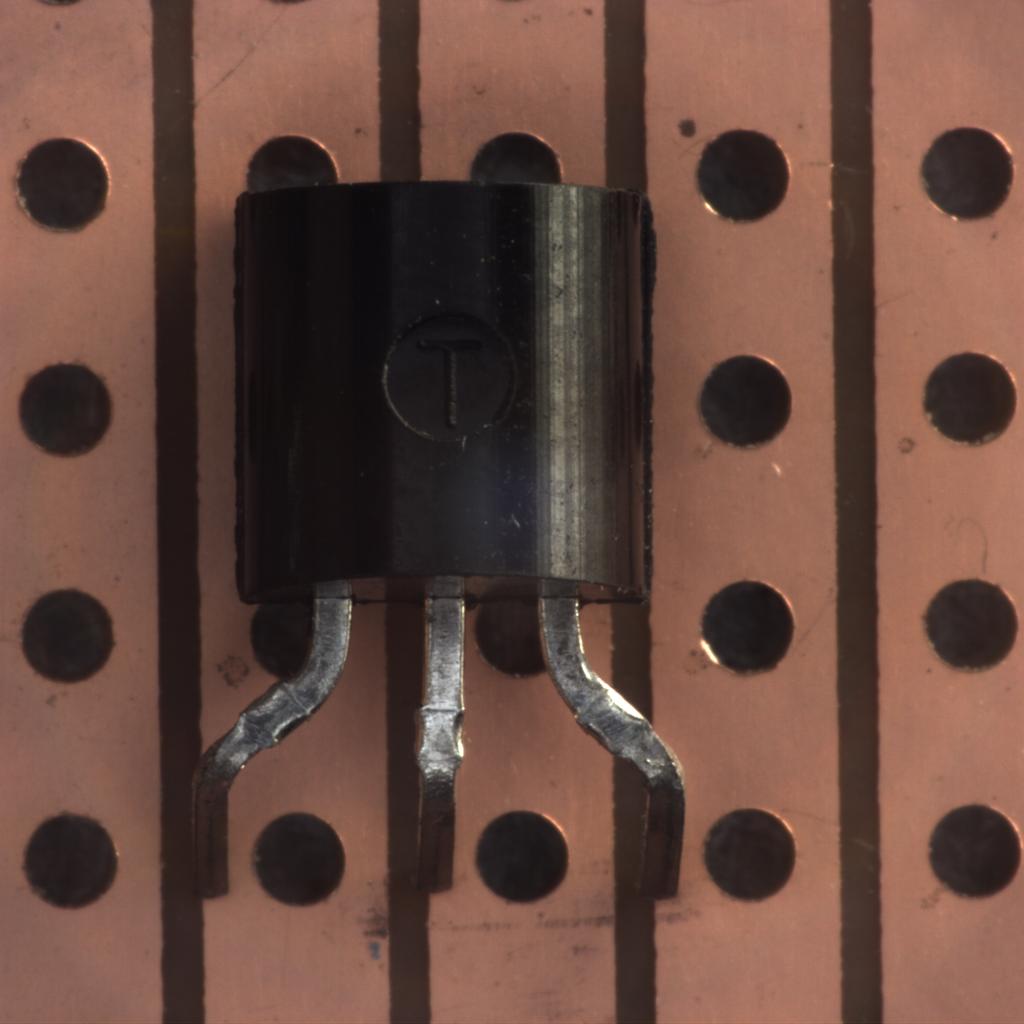}

    \caption{
    Representative samples of all five defect types in the MVTec AD Transistor category: 
    \textit{none}, \textit{bent\_lead}, \textit{cut\_lead}, \textit{damaged\_case}, and \textit{misplaced}. 
    }
    \label{fig:transistor_all_types}
\end{figure}

\begin{figure}[tp!]
    \centering
    \includegraphics[width=0.8\linewidth]{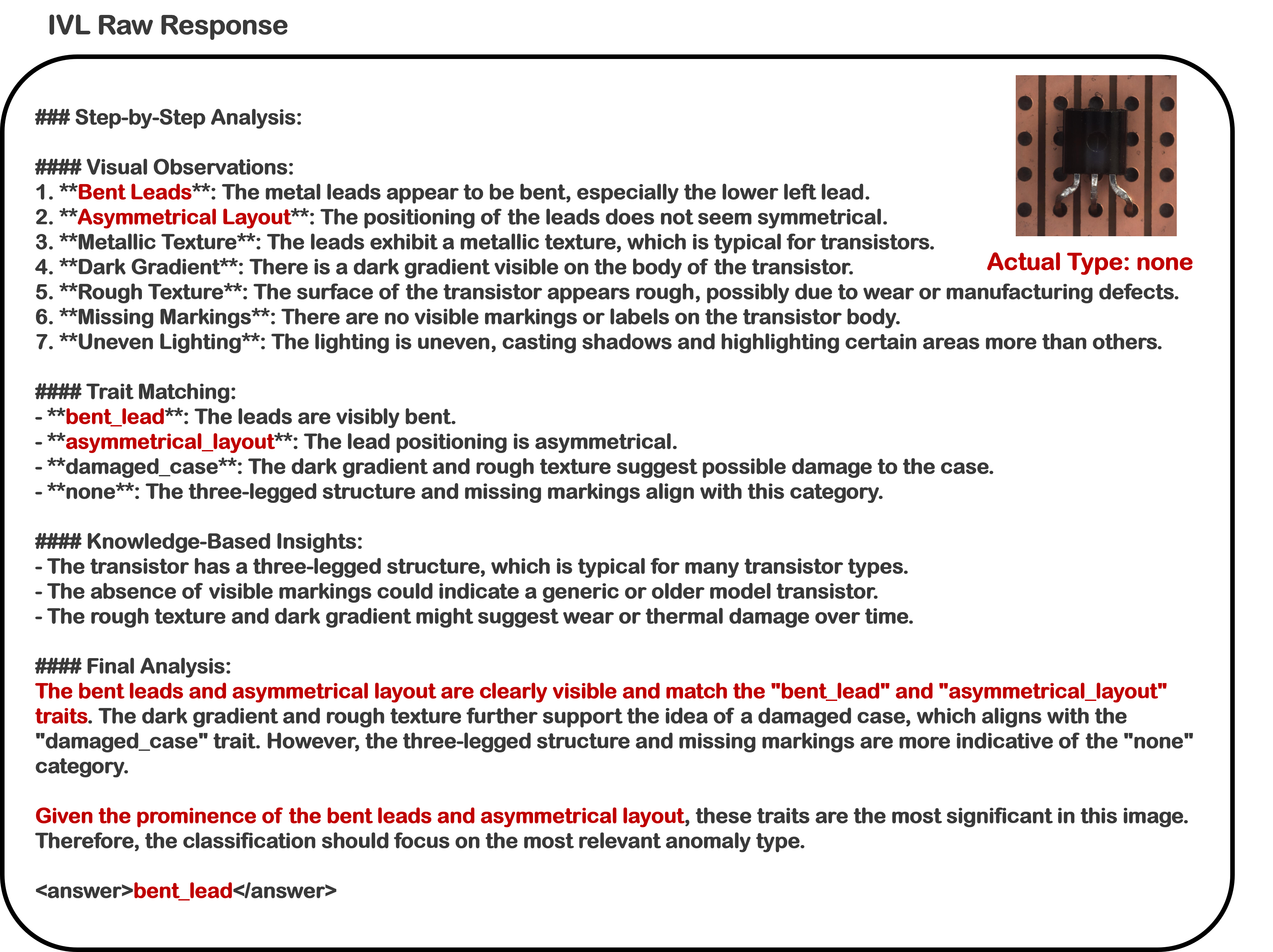}
    \caption{
    Reasoning breakdown on a normal MVTec AD sample (\textit{none}). IVL flags the object as \textit{bent\_lead} based on \textbf{Visual Observation 1} ("The metal leads appear to be bent"). While the leads are indeed curved, they lack the unnatural distortion of the actual defect (refer to \cref{fig:transistor_all_types} for a comparison). The failure stems from the lack of a comparative mechanism: IVL detects the "bent" primitive but, without comparing against a reference image, cannot discern that this curvature is within standard tolerance.}
    \label{fig:transistor_failure_mode}
\end{figure}

This pattern is driven by comparative ambiguity in the dataset. The true \textit{bent\_lead} defect corresponds to abnormally distorted metal leads, but other defect types and even normal samples may also contain naturally curved leads, as shown in \cref{fig:transistor_all_types}. Because IVL analyzes each image independently rather than against a normal reference, it often treats curved lead geometry as sufficient evidence for \textit{bent\_lead}. As a result, multiple visually distinct transistor states are collapsed into one dominant prediction. A representative failure case is shown in \Cref{fig:transistor_failure_mode}, where a normal sample is incorrectly flagged. Future work could address this by incorporating comparative reasoning modules that explicitly measure geometric deviation between the query and support samples.

\subsection{FractalDB: Applicability Boundary}
\label{appendix:failure_mode:fractaldb}

FractalDB exposes a structural limit of trait-based reasoning that is distinct from the failure modes above. Its class labels are induced by Iterated Function System (IFS) parameter families rather than human-verbalizable semantic categories, and different IFS families can yield visually similar patterns, so plausible natural-language descriptions of the images are often non-discriminative with respect to class identity. FractalDB is therefore distant-OOD but not trait-separable. It is distant-OOD by low NMI (0.318 $< \tau_{\mathrm{NMI}}$) while its MK-MMD (0.483) sits just below $\tau_{\mathrm{MMD}}$, placing it in the same low-divergence low-alignment quadrant as Pok\'emon yet with the opposite IVL outcome. This co-location shows that the diagnostic identifies the distant-OOD regime without predicting trait-separability, which motivates the separate trait-separability condition in Sec.~3.3 of the main paper.

Quantitatively, on a 20-class subset under Qwen2.5-VL (1-shot), all methods stay near chance. Zero-shot reaches $0.3\%$, SFT $5.5\%$, and IVL $4.9\%$. IVL provides no advantage here, as expected when classes are not separable by nameable visual traits. This contrasts with the trait-separable distant-OOD datasets, where IVL improves over SFT, and delineates the applicability boundary of the framework rather than a fixable defect.

\begin{figure}[h!]
    \centering
    \includegraphics[width=0.60\linewidth]{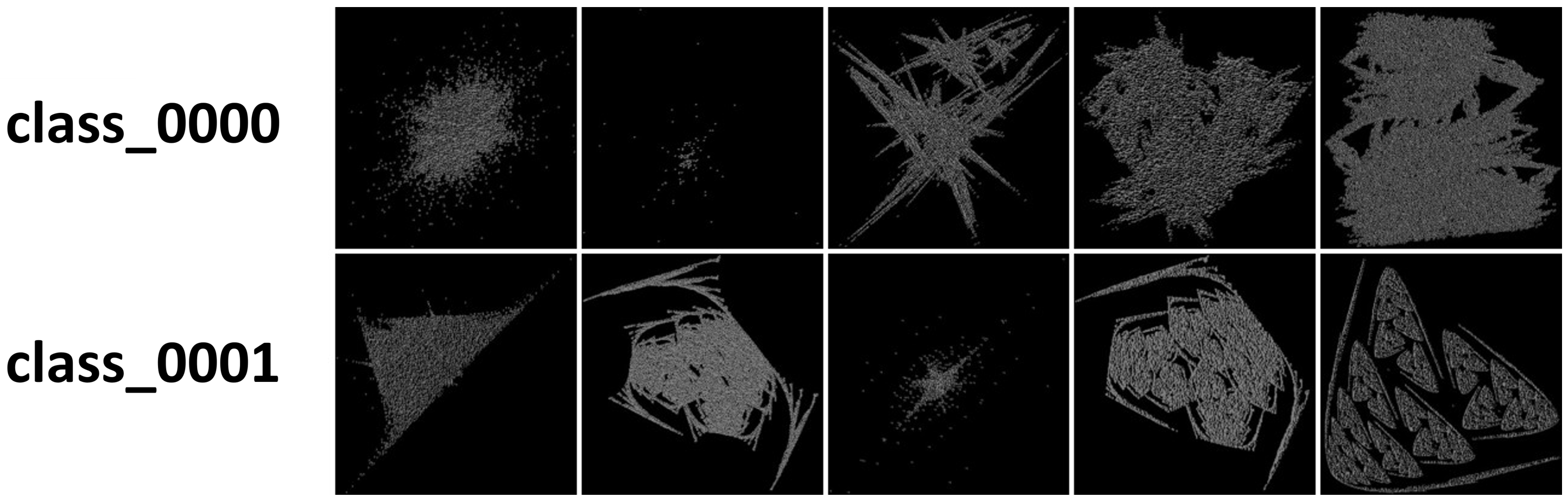}
    \caption{Example images from two FractalDB classes. The visual patterns are generated by different IFS parameter families but are difficult for humans to distinguish, limiting trait-based reasoning.}
    \label{fig:fractaldb_examples}
\end{figure}

\section{Computational Analysis}
\label{sec:computational_analysis}

We provide a comprehensive efficiency comparison of IVL against SFT and ICL baselines. All experiments were conducted on NVIDIA H100 GPUs using the Pok\'emon dataset (18 classes, 1-shot) with the Qwen2.5-VL-7B backbone. Costs are decomposed into \textbf{offline} (one-time adaptation) and \textbf{online} (per-image inference) stages.

\subsection{Efficiency Comparison Across Adaptation Methods}
\begin{table}[htbp]
    \centering
    \caption{Unified efficiency comparison across adaptation methods. Offline cost is a one-time expense amortized over all subsequent queries; online cost is per test image. IVL achieves the lowest total GPU compute for adaptation while requiring only a single consumer-grade GPU.}
    \label{tab:compute_unified}
    \small
    \begin{tabular}{@{} l c c c c c @{}}
        \toprule
        & \textbf{IVL} & \textbf{SFT} & \textbf{V-RFT} & \textbf{ICL-1} & \textbf{ICL-8} \\
        \midrule
        \multicolumn{6}{@{}l}{\textit{Offline (one-time adaptation)}} \\[2pt]
        Wall-clock (s)         & 603   & 309   & 412   & ---   & ---   \\
        \# GPUs                & 1     & 6     & 6     & ---   & ---   \\
        GPU-seconds            & 603   & 1{,}854 & 2{,}472 & 0   & 0     \\
        Peak VRAM (GB)         & 32    & 199   & 199   & ---   & ---   \\
        \midrule
        \multicolumn{6}{@{}l}{\textit{Online (per-image inference)}} \\[2pt]
        Latency (s)            & 7.06  & 4.33  & 4.33  & 3.14  & 5.24  \\
        \quad\textit{of which: retrieval pipeline} & \textit{0.76} & --- & --- & --- & --- \\
        \quad\textit{of which: VLM generation}     & \textit{6.26} & \textit{4.30} & \textit{4.30} & \textit{3.10} & \textit{5.20} \\
        VRAM (GB)              & 32    & 16    & 16    & 16    & 16    \\
        \midrule
        \multicolumn{6}{@{}l}{\textit{Accuracy on distant-OOD (from Table~1)}} \\[2pt]
        Pok\'emon (\%)         & \textbf{50.30} & 45.52 & 48.77 & 22.00 & 34.24 \\
        MVTec Mean (\%)        & \textbf{46.1}  & 41.3  & 38.1  & 26.0  & ---   \\
        \bottomrule
    \end{tabular}
\end{table}

\paragraph{Offline: single-GPU accessibility.}
IVL's trait building pipeline can run entirely on a single GPU with 32~GB VRAM, consuming 603 GPU-seconds. By contrast, SFT requires a minimum of 3 H100 GPUs (peak 199~GB VRAM) 
and 1{,}854 GPU-seconds; our experiments used 6 GPUs to reduce 
wall-clock time. This gap widens further for Visual-RFT (2{,}472 GPU-seconds). IVL can therefore run on consumer-grade hardware (\eg, a single NVIDIA RTX 5090), whereas gradient-based methods demand multi-node infrastructure. Importantly, the offline cost is amortized: once the trait database is built, it is reused for all subsequent queries without modification.

\paragraph{Online: moderate overhead dominated by VLM generation.}
IVL's per-image inference latency is 7.06~s, approximately $1.6\times$ that of SFT (4.33~s). However, as the breakdown in \Cref{tab:compute_unified} reveals, the trait retrieval pipeline (ROI cropping, two-stage filtering) accounts for only 0.76~s of this overhead. The remaining 1.96~s arises from the VLM processing a longer input context that includes retrieved trait descriptions and localized visual crops, which is the same enriched prompt that produces IVL's accuracy advantage. Compared to 8-shot ICL (5.24~s), which achieves far lower accuracy, IVL's additional 1.8~s latency is a modest cost for the substantial accuracy gains on distant-OOD benchmarks.

\paragraph{Cost--accuracy tradeoff.}
The bottom rows of \Cref{tab:compute_unified} contextualize efficiency against accuracy. On the Pok\'emon benchmark, IVL outperforms SFT by $+4.8$ pp while using $67\%$ fewer GPU-seconds for adaptation. On MVTec AD (mean over 15 categories), IVL exceeds SFT by $+4.8$ pp and Visual-RFT by $+8.0$ pp. For practitioners targeting specialized domains where gradient-based methods yield marginal or even negative gains (cf.\ Proposition~2), IVL offers a compelling tradeoff: lower hardware requirements, no gradient infrastructure, and higher accuracy precisely where it matters most.

\subsection{Scalability with Number of Classes}
\label{sec:computational_analysis:scalability}

A potential concern is whether IVL's inference cost scales unfavourably with the number of classes $C$, since the trait database $\mathcal{T}$ grows with $C$. It does not, because IVL classifies through a single VLM pass over a fixed-size filtered trait set rather than one verification pass per class.

\begin{table}[h!]
\centering
\caption{Scalability of IVL with number of classes $C$ on FGVC Aircraft (Qwen2.5-VL-7B, 1-shot). ``Offline s/class'' is the per-class adaptation wall-clock time (stages S1--S4). ``Online s/image'' is total per-image inference time (S5 + retrieval). ``Inference s/image'' is the VLM generation time within S5 only. Ratio columns are relative to the $C{=}20$ baseline.}
\label{tab:scalability}
\small
\setlength{\tabcolsep}{4pt}
\resizebox{\linewidth}{!}{%
\begin{tabular}{rrrrrrrr}
\toprule
$C$ & \textbf{\#Test} & \textbf{Offline s/class} & \textbf{Ratio} & \textbf{Online s/image} & \textbf{Ratio} & \textbf{Inference s/image} & \textbf{Ratio} \\
\midrule
 20  & 100  & 68.60 & 1.000 & 16.85 & 1.000 & 9.056 & 1.000 \\
 40  & 200  & 63.65 & 0.928 & 15.80 & 0.938 & 8.902 & 0.983 \\
 60  & 300  & 61.23 & 0.893 & 15.45 & 0.917 & 8.711 & 0.962 \\
 80  & 400  & 58.84 & 0.858 & 15.34 & 0.910 & 8.404 & 0.928 \\
100  & 500  & 61.60 & 0.898 & 15.17 & 0.900 & 9.052 & 1.000 \\
\bottomrule
\end{tabular}
}
\end{table}

Two patterns emerge from \Cref{tab:scalability}. First, online latency is nearly constant as $C$ grows from 20 to 100, with the ratio staying within $[0.90, 1.00]$ and showing no systematic upward trend. This follows from the inference design. Retrieval is linear in $|\mathcal{T}|$ but contributes only a small fraction of the total, and the two-stage filtering caps the traits entering the final VLM pass at $k_2$ independent of $|\mathcal{T}|$, so the dominant generation cost does not grow with $C$. Second, per-class offline cost is stable or slightly decreasing with $C$ (ratio $0.86$--$0.93$), indicating that HDBSCAN amortizes certain fixed-cost operations over more support images as the database grows. The higher absolute online cost here than in \Cref{tab:compute_unified} reflects Aircraft's larger trait list and higher-resolution inputs, not a dependence on $C$. For settings with very large $|\mathcal{T}|$, approximate nearest-neighbour indices such as FAISS would be a natural extension.


\section{Non-Generative Baselines}
\label{appendix:non_generative_baselines}

For completeness, we additionally evaluate non-generative vision-language models on the distant-OOD benchmarks. \Cref{tab:non_generative_baselines} reports results for CLIP ViT-B/32 with adapter-based methods (Tip-Adapter~\cite{zhang2021tip}, CoOp~\cite{coop}) and CuPL~\cite{pratt2023does}, as well as the source-free domain adaptation baseline DIFO~\cite{tang2024source} operating on ResNet features.
\vspace{-1.0em}
\begin{table}[htp!]
    \centering
    \caption{Non-generative vision-language baselines on distant-OOD benchmarks (1-shot for Tip-Adpater and CoOp, 0-shot for other methods).}
    \vspace{-1.0em}
    \label{tab:non_generative_baselines}
    \setlength{\aboverulesep}{0pt}
    \setlength{\belowrulesep}{0pt}
    \renewcommand{\arraystretch}{1.15}
    \resizebox{\textwidth}{!}{%
    \begin{tabular}{@{}l @{\hspace{0.8em}} ccc @{\hspace{0.8em}} ccccccccccccccc @{\hspace{0.8em}} c@{}}
        \toprule
        & \multicolumn{3}{c}{\textbf{Standalone}} &
          \multicolumn{15}{c}{\textbf{MVTec AD (per-category)}} & \\
        \cmidrule(lr){2-4} \cmidrule(lr){5-19}
        \rowcolor{gray!20}
        \textbf{Method} &
        \rotatebox{70}{\textbf{Pok\'{e}mon}} &
        \rotatebox{70}{\textbf{Ret.\ OCT}} &
        \rotatebox{70}{\textbf{WM811k}} &
        \rotatebox{70}{Bottle} &
        \rotatebox{70}{Cable} &
        \rotatebox{70}{Capsule} &
        \rotatebox{70}{Carpet} &
        \rotatebox{70}{Grid} &
        \rotatebox{70}{Hazelnut} &
        \rotatebox{70}{Leather} &
        \rotatebox{70}{Metal Nut} &
        \rotatebox{70}{Pill} &
        \rotatebox{70}{Screw} &
        \rotatebox{70}{Tile} &
        \rotatebox{70}{T.brush} &
        \rotatebox{70}{Trans.} &
        \rotatebox{70}{Wood} &
        \rotatebox{70}{Zipper} &
        \rotatebox{70}{\textbf{MVT.\ Avg}} \\
        \midrule
        \multicolumn{20}{l}{\cellcolor{gray!5}\textit{CLIP ViT-B/32}} \\
        CLIP Vanilla &
            42.54 & 12.29 & 13.49 &
            49.4 & 9.9 & 16.7 & 16.2 & 11.1 & 15.2 & 30.5 & 21.8 & 12.6 & 13.6 & 31.5 & 72.5 & 12.6 & 13.7 & 11.2 &
            22.6 \\
        CuPL \cite{pratt2023does} &
            36.11 & 12.68 & 9.64 &
            34.2 & 6.4 & 20.6 & 16.2 & 16.7 & 35.2 & 22.9 & 34.6 & 5.0 & 14.9 & 58.6 & 72.5 & 9.5 & 35.6 & 11.2 &
            26.3 \\
        Tip-Adapter \cite{zhang2021tip} &
            36.11 & 32.93 & 36.75 &
            46.8 & 34.0 & 26.4 & 43.2 & 44.4 & 48.6 & 36.4 & 41.8 & 27.7 & 25.3 & 45.9 & 70.0 & 12.6 & 57.5 & 12.6 &
            38.1 \\
        CoOp \cite{coop} &
            27.78 & 32.3 & 23.1 &
            48.1 & 14.2 & 26.2 & 34.2 & 13.9 & 40.9 & 59.3 & 19.1 & 12.0 & 16.2 & 81.1 & 65.0 & 11.6 & 65.8 & 9.8 &
            34.5 \\
        \midrule
        \multicolumn{20}{l}{\cellcolor{gray!5}\textit{DIFO-C-B32 \cite{tang2024source} (ResNet)}} \\
        DIFO &
            5.56 & 12.07 & 14.58 &
            25.3 & 14.9 & 19.0 & 17.1 & 23.6 & 14.3 & 15.2 & 4.5 & 14.5 & 17.5 & 36.9 & 62.5 & 48.4 & 28.8 & 10.5 &
            23.6 \\
        \bottomrule
    \end{tabular}%
    }\vspace{-2.0em}
\end{table}

\section{Reproducibility}

\label{appendix:reproducibility}
\subsection{IVL Pipeline Hyperparameters}
\label{appendix:reproducibility:hyperparams}

\begin{table}[!htp]
\centering
\caption{IVL pipeline default hyperparameters.}
\vspace{-1.0em}
\label{tab:hyperparams}
\small
\begin{tabular}{l r l}
\toprule
\textbf{Hyperparameter} & \textbf{Value} & \textbf{Description} \\
\midrule
\multicolumn{3}{l}{\textit{Trait Filtering ($k_1$, $k_2$)}} \\
Coarse ratio & 0.8 & Fraction of traits retained at global stage \\
$k_1$ (coarse min) & 50 & Minimum number of globally retained traits \\
$k_1$ (coarse max) & 1000 & Maximum number of globally retained traits \\
$k_2$ (fine filter) & 250 & Number of traits retained after local refinement \\
\midrule
\multicolumn{3}{l}{\textit{HDBSCAN Clustering}} \\
\texttt{min\_cluster\_size} & 2 & Minimum points to form a cluster \\
\texttt{min\_samples} & 1 & Core-point neighbourhood size \\
Metric & Euclidean & Applied to L2-normalised embeddings \\
& & (mathematically equivalent to cosine distance) \\
\midrule
\multicolumn{3}{l}{\textit{Region Extraction}} \\
NMS IoU threshold & 0.4 & Non-maximum suppression for bounding boxes \\
Top regions & 3 & Number of cropped regions fed to the VLM \\
\midrule
\multicolumn{3}{l}{\textit{Trait Extraction}} \\
Runs per image ($K$) & 5 & Independent extraction runs (self-consistency) \\
Traits per run & 6--8 per mode & Target range specified in the extraction prompt \\
\bottomrule
\end{tabular}
\end{table}

\Cref{tab:hyperparams} reports all non-trivial hyperparameters used in the IVL pipeline. Default values were fixed before any experiments and were not tuned per dataset.

The effective $k_1$ value for a given dataset is computed as
$k_1 = \mathrm{clip}\!\bigl(\mathrm{round}(|\mathcal{T}| \times 0.8),\; 50,\; 1000\bigr)$,
where $|\mathcal{T}|$ is the total number of traits in the database after canonicalization.
This adaptive rule retains $80\%$ of the trait database while enforcing minimum and maximum bounds,
ensuring that very small databases are not over-filtered and very large ones do not saturate the
attention computation.

\subsection{IVL Prompts}
\begin{figure}[!t]
    \centering
    \includegraphics[width=\linewidth]{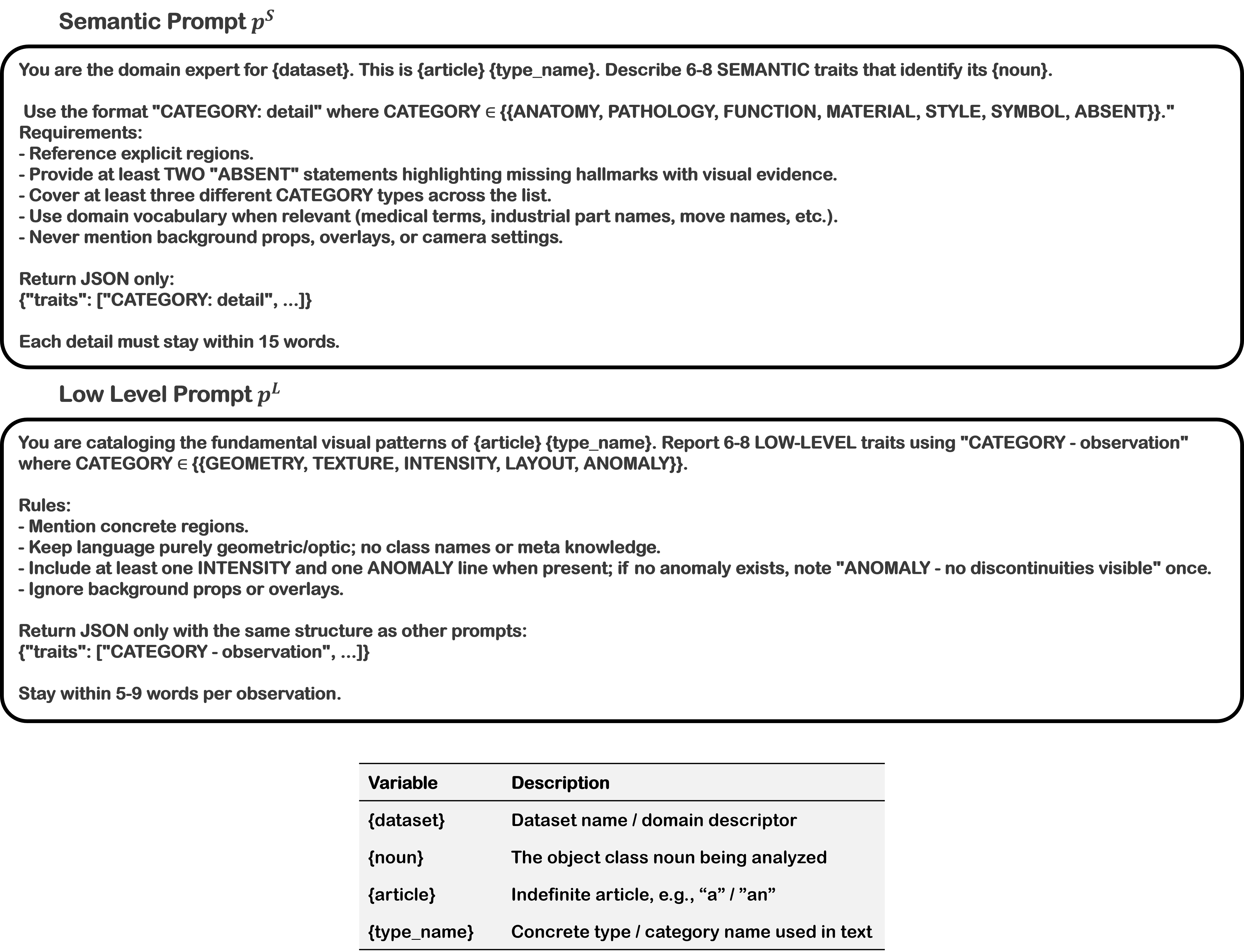}
    \vspace{-2.0em}
    \caption{Extraction Prompt. }
    \label{fig:enhanced_hybrid_categorized_prompt}
\end{figure}

\begin{figure}[!t]
    \centering
    \includegraphics[width=\linewidth]{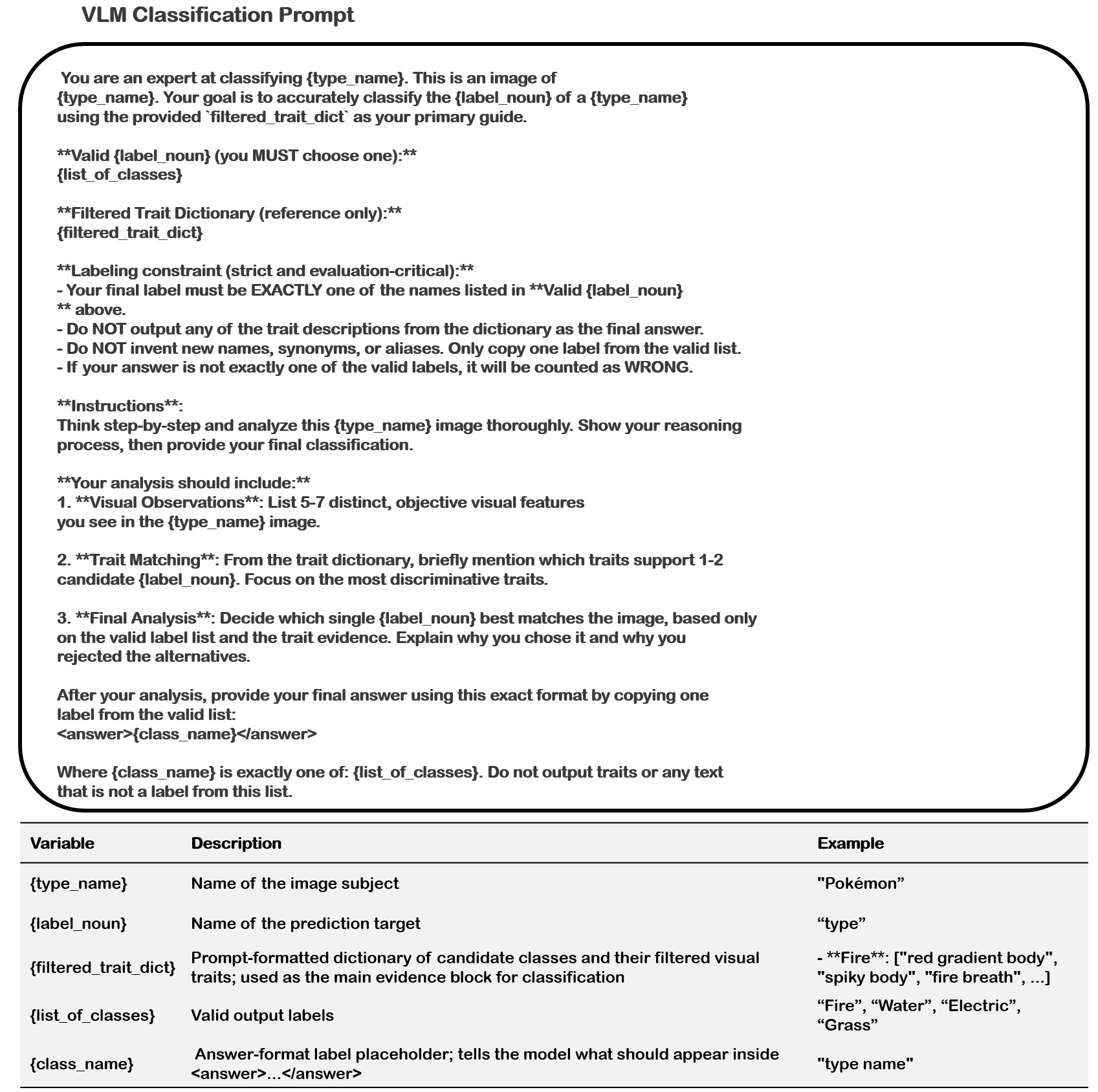}
    \vspace{-2.0em}
    \caption{VLM Classification Prompt.}
    \label{fig:vlm_classification_prompt}
\end{figure}

{As shown in \cref{fig:enhanced_hybrid_categorized_prompt}, we employ a unified prompt template that dynamically adapts to each dataset via format strings. This dual-branch approach extracts both semantic traits and low-level visual primitives, which are subsequently combined for the next stage.

\Cref{fig:vlm_classification_prompt} shows the classification prompt used at inference time. The prompt
receives three inputs: (1)~the original test image $x_{\mathrm{test}}$, (2)~the refined trait
dictionary $\mathcal{T}^{(2)}$ as structured text, and (3)~the three localized image crops
extracted by the region extraction step. The structured output format (Visual Observations,
Trait Matching, Final Analysis) is what enables the automated trait-class alignment parsing
described in Supp.~\S\ref{sec:supp:trait_alignment}.

IVL's accuracy advantage stems from the content of the trait dictionary (observations grounded in support images) rather than from prompt structure alone. Two results confirm this. First, the Menon~\etal\ and CuPL baselines (Table~1 of the main paper) supply VLM-generated descriptors at inference using comparable structured prompts, yet underperform IVL by more than 13~pp on MVTec~AD mean, confirming that prompt richness without support-image-grounded traits is insufficient. Second, the ``w/o Grounding'' ablation (Supp.~\cref{tab:ablation:visual_grounding}) shows that removing localized evidence while retaining the full trait dictionary and prompt structure costs $-4.2$~pp, confirming that the structure of evidence drives the accuracy gain.}

\begin{figure}[!t]
    \centering
    \includegraphics[width=\linewidth]{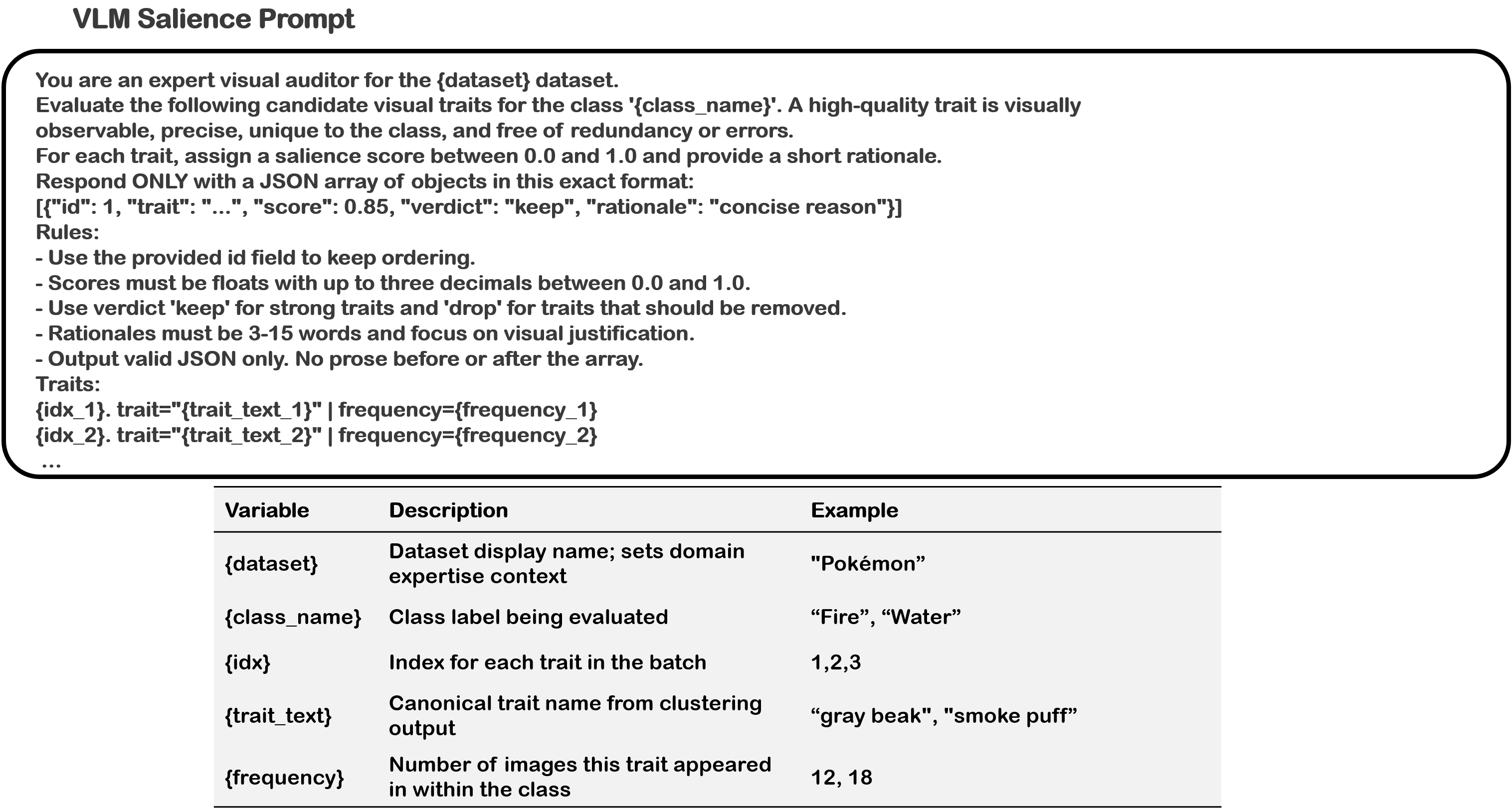}
    \vspace{-2.0em}
    \caption{VLM Salience Prompt.}
    \vspace{-1.5em}
    \label{fig:vlm_salience_prompt}
\end{figure}

\subsection{Baseline}
\label{appendix:reproducibility:baseline}

We standardized computation and methodology across baselines by using the same backbone and unified training/evaluation protocols. Unless otherwise stated, we kept default hyperparameters from each framework. For Qwen-VL, we use Qwen/Qwen2.5-VL-7B-Instruct from Hugging Face, for LLaVA, we use llava-hf/llava-v1.6-mistral-7b-hf from Hugging Face. SFT and SFT+LoRA are implemented via \textbf{LLaMA-Factory}; RFT uses the \textbf{Visual-RFT} pipeline~\cite{visualrft}. For the method proposed in Menon \etal\cite{menon2022visual}, we replaced both the GPT-based descriptor generation stage and the CLIP-based visual grounding and scoring stage with the same backbone used in each experimental condition: Qwen2.5-VL-7B when evaluating under the Qwen backbone, and LLaVA-1.6-Mistral-7B when evaluating under the LLaVA backbone. The original method uses GPT-3/4 for descriptor generation and CLIP for visual grounding. We substitute these with the same backbone used across all other methods in the comparison so that observed performance differences reflect methodological design rather than backbone capability. For the method CuPL~\cite{pratt2023does}, which originally uses GPT-3 to generate class-specific image prompts and CLIP for image-text similarity scoring, we similarly replace these components with the unified backbone used in our experiments. Specifically, GPT-3 for prompt generation is replaced with Qwen2.5-VL-7B-Instruct. In addition, the CLIP image encoder and text encoder used for computing image-text similarity are replaced with the corresponding visual and textual representations produced by Qwen. For the method DIFO~\cite{tang2024source}, we follow the original implementation and experimental
configuration without modification.

This unified setup ensures that performance differences reflect methodological advantages rather than implementation artifacts or backbone variance. Keeping the computational configuration consistent across all methods enables direct efficiency comparisons and highlights the benefits of the proposed trait-based approach relative to parameter-adaptation baselines.

\subsection{In-Context Learning Evaluation}
\label{appendix:reproducibility:icl}
\begin{figure}[!t]
    \centering
    \includegraphics[width=0.85\linewidth]{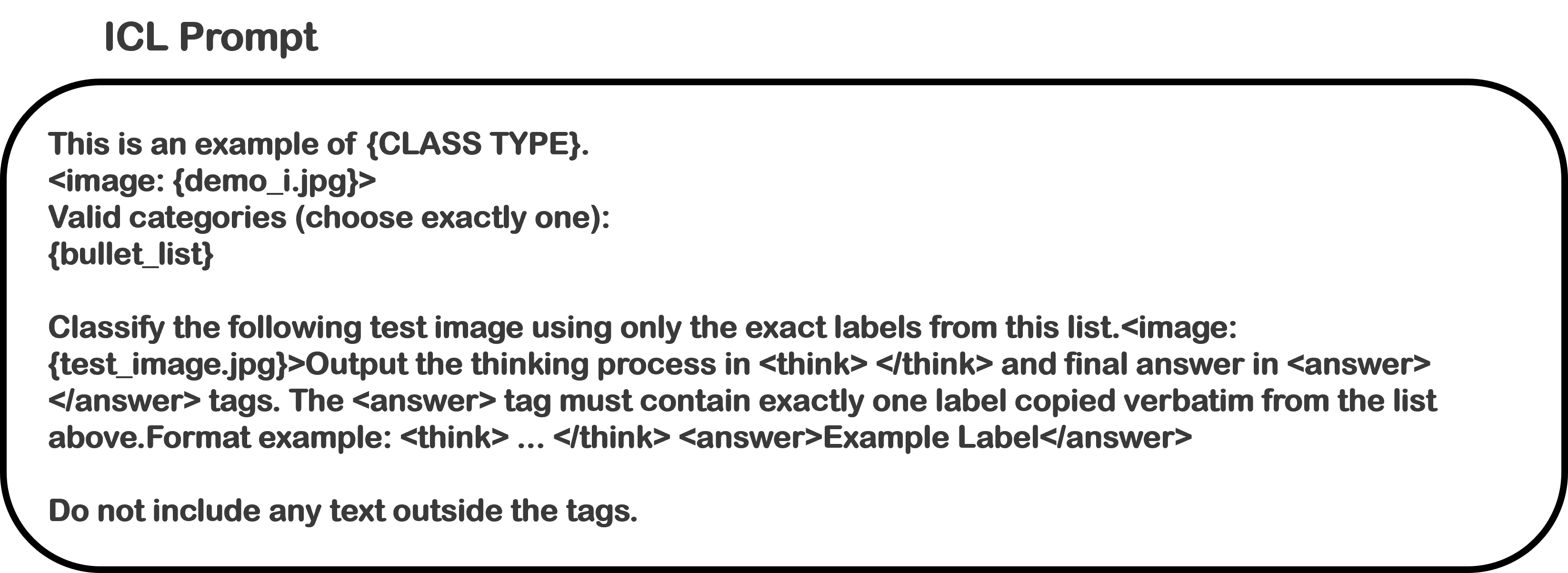}
    \caption{ICL Prompt. For each dataset, we use a simple format string to make the model aware of which dataset it is inferring on.}
    \label{fig:icl_prompt}
\end{figure}

To benchmark training-free performance, we include few-shot in-context learning (ICL) as a baseline that, like our approach, requires no parameter updates. For each dataset and shot setting $n \in \{1,8\}$, we construct a fixed class-balanced support set $\mathcal{D}={(x_i,y_i)}_{i=1}^{nC}$ before evaluation using the same deterministic sampling protocol as all other few-shot methods. For every query, ICL prepends the support examples from $\mathcal{D}$ as image--instruction--answer demonstrations in a fixed canonical order (alphabetical by class label, then by image filename within each class), following the prompt format shown in \cref{fig:icl_prompt}. The query block contains only the query image and instruction. No test image is used as a demonstration.

\section{Qualitative Results}
\label{appendix:qualitative}
\begin{figure}[!htp]
    \centering
    \includegraphics[width=\linewidth]{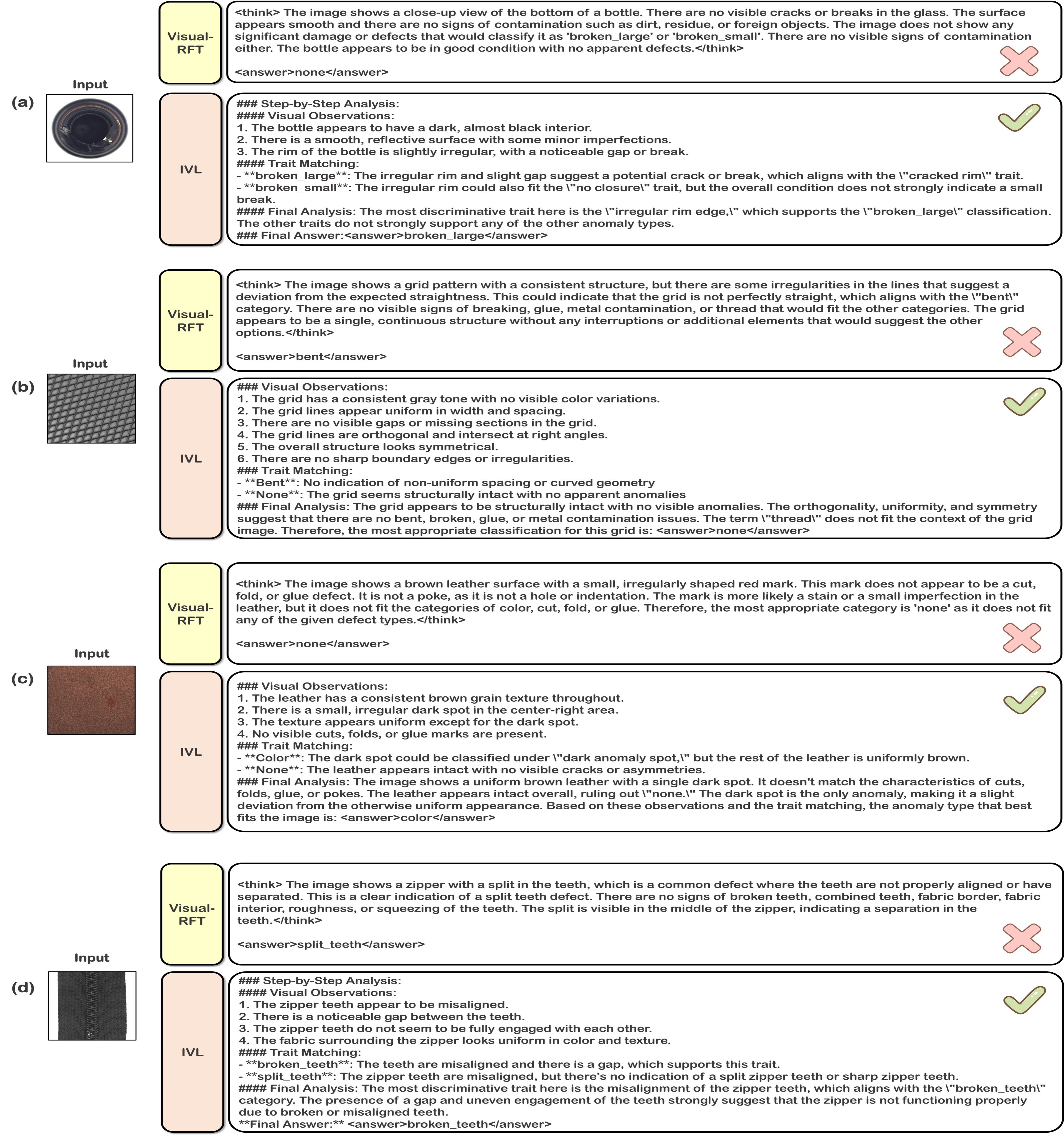}
    \caption{More qualitative results on the MVTec AD dataset.}
    \label{fig:mvtec_qualitative_results_1}
\end{figure}

\begin{figure}[!htp]
    \centering
    \includegraphics[width=\linewidth]{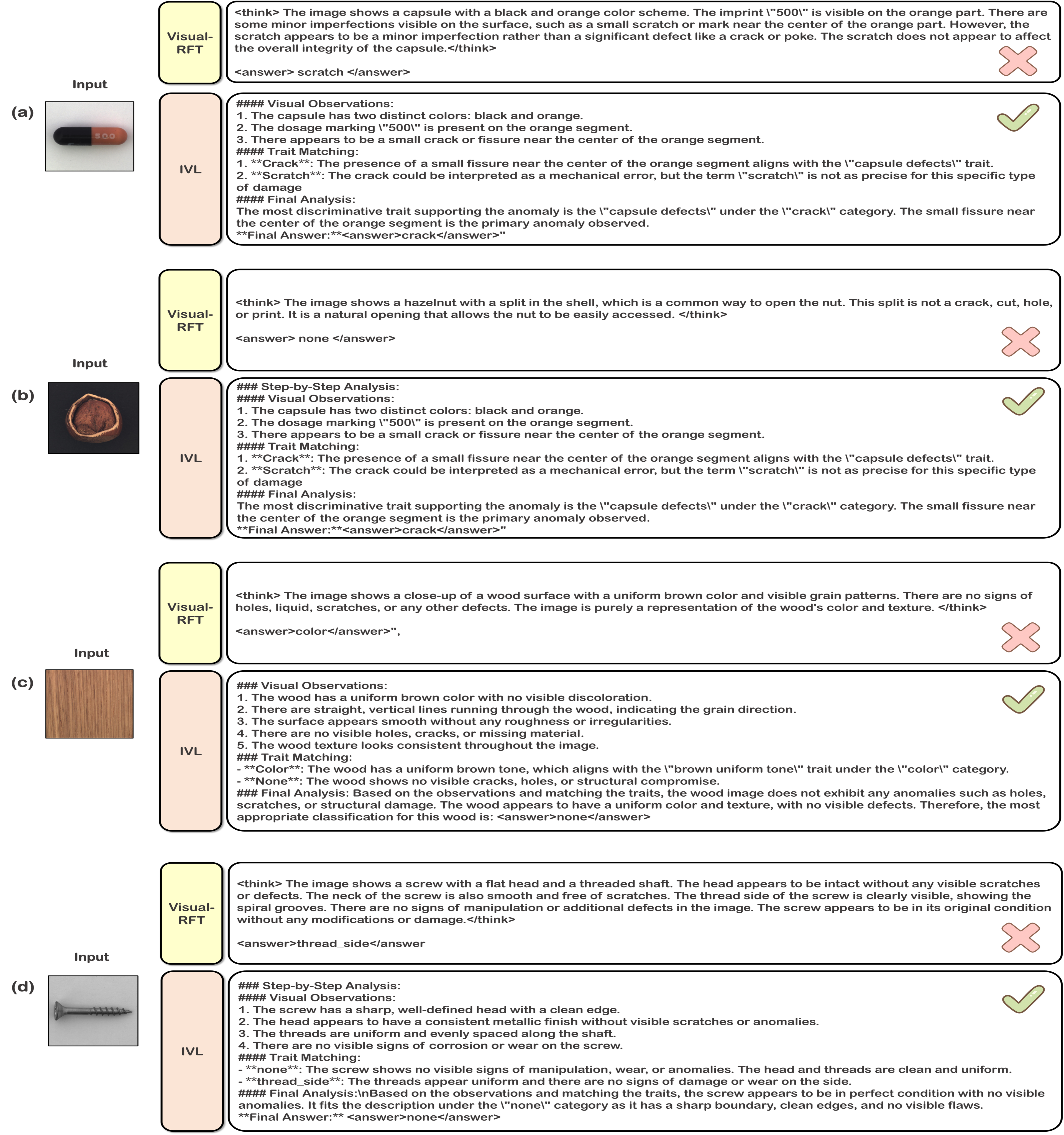}
    \caption{More qualitative results on the MVTec AD dataset.}
    \label{fig:mvtec_qualitative_results_2}
\end{figure}

\begin{figure}[!htp]
    \centering
    \includegraphics[width=\linewidth]{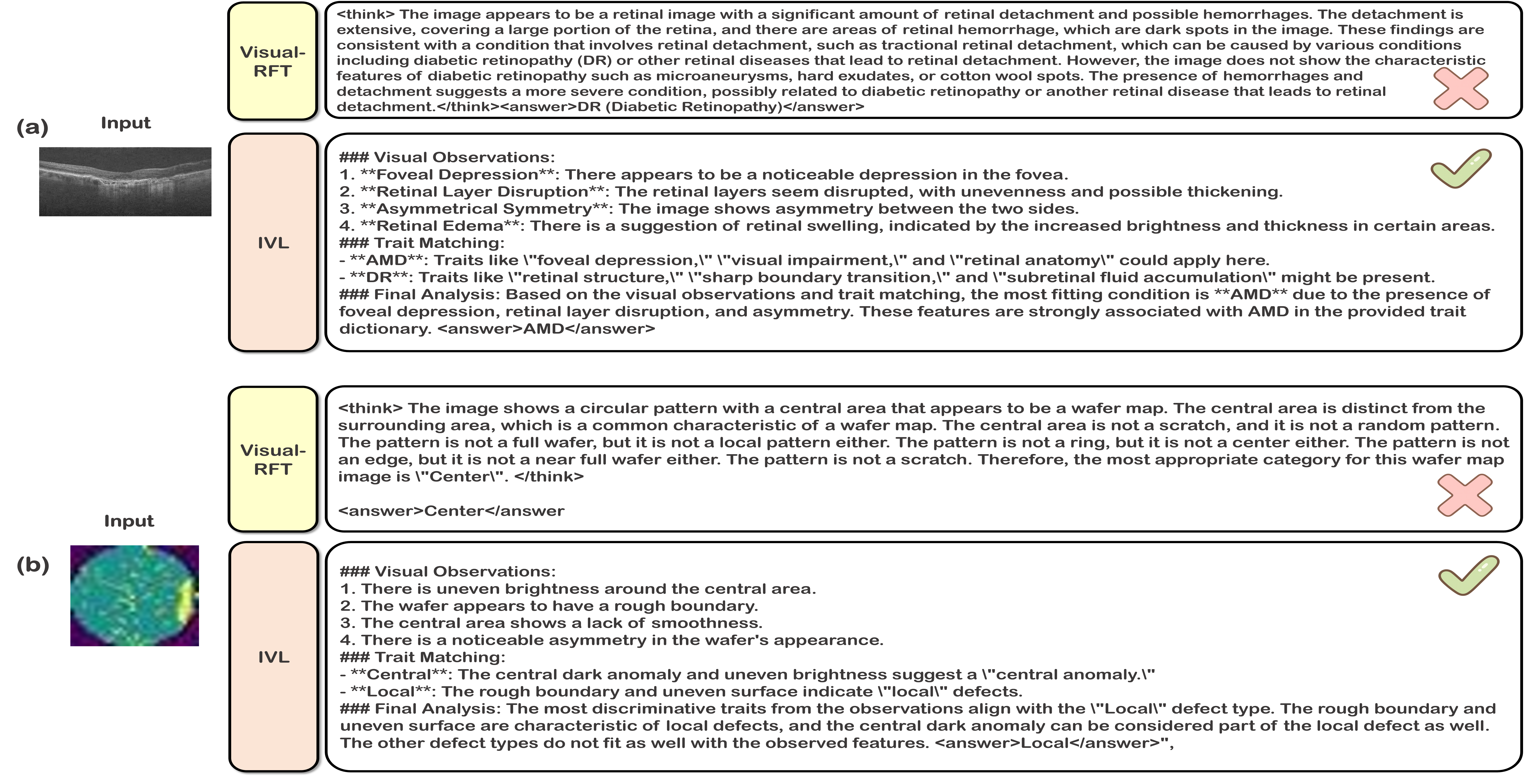}
    \caption{More qualitative results on the Retinal OCT, WM-811K dataset.}
    \label{fig:mvtec_qualitative_results_3}
\end{figure}

We provide additional qualitative comparisons between IVL and Visual-RFT~\cite{visualrft} in \cref{fig:mvtec_qualitative_results_1}, \cref{fig:mvtec_qualitative_results_2}, and \cref{fig:mvtec_qualitative_results_3}. These examples illustrate the differences in reasoning patterns between the two approaches when performing classification.

Visual-RFT typically produces its final answer before articulating intermediate reasoning. In several examples, its subsequent explanation is inconsistent with the visual evidence and leads to incorrect classifications. For example, in \cref{fig:mvtec_qualitative_results_2}(b), Visual-RFT interprets the split in the hazelnut shell as a natural opening used to access the nut rather than a structural defect. It therefore predicts \textit{none} and overlooks the crack-like fissure that constitutes the actual anomaly. Although the explanation is fluent and seemingly plausible, it results in a missed detection.

In contrast, IVL follows a structured inductive reasoning pipeline. It first identifies explicit visual observations and then compares them against a predefined trait dictionary to evaluate candidate classes. In the same example, IVL identifies the crack-like fissure as evidence for the \texttt{crack} trait and predicts the corresponding anomaly class. This procedure links the final prediction to cited visual evidence.

These qualitative examples illustrate that IVL produces more interpretable reasoning traces whose cited evidence is consistent with the final prediction. By grounding decisions in observable traits, IVL provides a transparent basis for classification when the visual evidence is subtle.

\end{document}